\documentclass[nohyperref]{article}

\usepackage{microtype}
\usepackage{graphicx}
\usepackage{booktabs} 

\usepackage{epsfig}
\usepackage{graphicx}
\usepackage{amsmath}
\usepackage{amssymb}
\usepackage{tabularx}
\usepackage{multirow}
\usepackage{caption}
\usepackage{hhline}
\usepackage[export]{adjustbox}
\usepackage{subfig}
\usepackage{tabu}
\usepackage{boldline}
\usepackage{verbatimbox}
\usepackage{pifont}
\usepackage{tablefootnote}
\usepackage{footnote}
\usepackage{physics}
\usepackage{multicol}
\usepackage{enumitem}

\usepackage{hyperref}

\usepackage[accepted]{icml2022}

\usepackage{amsmath}
\usepackage{amssymb}
\usepackage{mathtools}
\usepackage{amsthm}

\usepackage[capitalize,noabbrev]{cleveref}

\theoremstyle{plain}
\newtheorem{theorem}{Theorem}[section]

\theoremstyle{definition}

\theoremstyle{remark}

\usepackage[textsize=tiny]{todonotes}

\icmltitlerunning{REvolveR: Continuous Evolutionary Models for Robot-to-robot Policy Transfer}

\begin{document}

\twocolumn[
\icmltitle{ REvolveR: Continuous Evolutionary Models for Robot-to-robot Policy Transfer }




\begin{icmlauthorlist}
\icmlauthor{Xingyu Liu}{cmu}
\icmlauthor{Deepak Pathak}{cmu}
\icmlauthor{Kris M. Kitani}{cmu}
\end{icmlauthorlist}

\icmlaffiliation{cmu}{The Robotics Institute, Carnegie Mellon University,
Pittsburgh, PA 15213, USA}

\icmlcorrespondingauthor{Xingyu Liu}{xingyul3@cs.cmu.edu}

\icmlkeywords{Policy Transfer, Transfer Learning, Imitation Learning}

\vskip 0.3in
]


\printAffiliationsAndNotice{}  

\begin{abstract}
A popular paradigm in robotic learning is to train a policy from scratch for every new robot. 
This is not only inefficient but also often impractical for complex robots. 
In this work, we consider the problem of transferring a policy across two different robots with significantly different parameters such as kinematics and morphology. 
Existing approaches that train a new policy by matching the action or state transition distribution, including imitation learning methods, fail due to optimal action and/or state distribution being mismatched in different robots. 
In this paper, we propose a novel method named \emph{REvolveR} of using continuous evolutionary models for robotic policy transfer implemented in a physics simulator. 
We interpolate between the source robot and the target robot by finding a continuous evolutionary change of robot parameters. 
An expert policy on the source robot is transferred through training on a sequence of intermediate robots that gradually evolve into the target robot. 
Experiments on a physics simulator show that the proposed continuous evolutionary model can effectively transfer the policy across robots and achieve superior sample efficiency on new robots.
The proposed method is especially advantageous in sparse reward settings where exploration can be significantly reduced.
Code is released at \url{https://github.com/xingyul/revolver}.
\end{abstract}

\section{Introduction}
\label{sec:intro}
A popular paradigm in learning robotic skills is to leverage reinforcement learning (RL) algorithms to train a policy for every new robot in every new environment from scratch.
This is not only inefficient in terms of sample efficiency but also often impractical for complex robots due to an extremely large exploration space.
How can one transfer a well-trained policy on one robot to another robot?

\begin{figure}[t] 
\centering
\newcommand\teaserwidth{0.98}
\subfloat{
    \includegraphics[width=\teaserwidth\linewidth, valign=t]{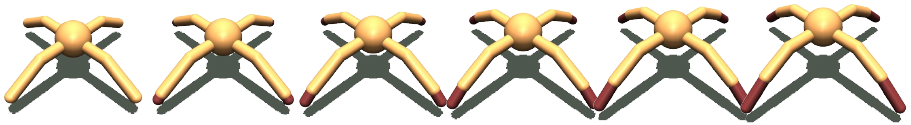}
}
\\
\subfloat{
    \includegraphics[width=\teaserwidth\linewidth, valign=t]{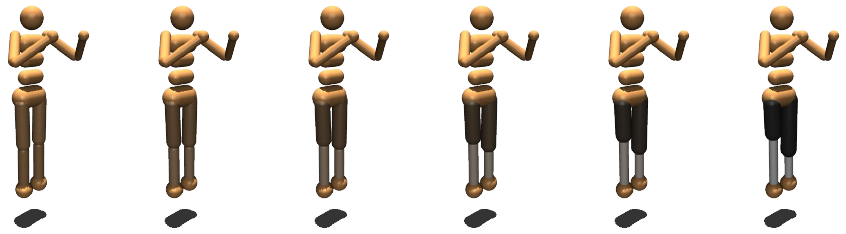}
}
\\
\subfloat{
    \includegraphics[width=\teaserwidth\linewidth, valign=t]{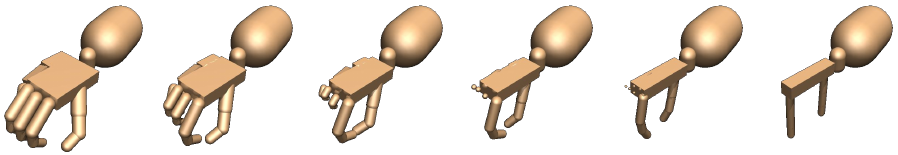}
}
\caption{ \textbf{Continuous robot evolution model}
allows policy to be transferred from one robot to another robot.
Upper row: an Ant robot continuously grow additional legs from the tip of its feet.
Middle row: a Humanoid robot continuously changes the length and mass of its legs.
Lower row: a dexterous gripper continuously shrink three of its fingers to evolve to a two-finger gripper.
We show the robots at evolution parameters of 0.0, 0.2, 0.4, 0.6, 0.8 and 1.0 respectively from left to right in each row.
}
\label{fig:teaser}
\end{figure}

Past endeavors have explored two main directions for transferring policy between robots.
Statistic matching Imitation learning (IL) methods train a new policy on the target robot with the aim of matching the behavior of the policy on a source robot.
Methods that optimize to match the distribution of actions \cite{bc}, state rollouts \cite{sail,soil}, or reward function \cite{irl,gail} have been successful on robotic learning tasks on robot with \textbf{similar} dynamics.
However, these methods are unable to deal with cases with \textbf{very large difference} in robot parameters and dynamics, since when mapped to the same state and action space, the robots could have very different optimal distributions of states or actions.
An alternative to imitation learning is to learn the robot hardware dynamics together with the policy by encoding the robot hardware specifics with neural networks
\cite{chen2018hardware,huang2020one}.
However, to train such hardware-aware policies, it usually requires training diverse tasks on a huge number of robots in advance, which could be computationally prohibitive.

In this paper, we propose a new paradigm for policy transfer between robots.
Our framework, named \emph{REvolveR}, is based on defining a continuous evolution of robots, where both the robot morphology and kinematics are continually adjusted to allow transforming one robot (source) to another robot (target), as illustrated in Figure \ref{fig:teaser}.

Specifically, the continuous evolutionary model interpolates two different robots by producing an infinite number of intermediate robots whose parameters are represented in continuous space.
These intermediate robots act as the ``bridge'' for transferring the policy from the source robot to the target robot. 
We are able to evaluate any robot along this continuum using physics simulation.
Then the policy is progressively trained on a sequence of intermediate robots whose robot parameters gradually evolve towards the target robot.
Since the change of the evolved robot parameters and hardware dynamics is small enough, it is typically easy for the policy to adapt to the new robots.
By the joint gradual evolution of robot hardware dynamics and the policy, we decompose the difficult robot-to-robot policy transfer problem into a sequence of policy fine-tuning problems that are much easier to solve.

Additionally, we propose several approaches that improve sample efficiency and stabilize training during the robot-to-robot policy transfer.
To stabilize training, we propose a local randomized evolution strategy where in each training epoch, we randomly sample a set of robots over a small continuous range of robot agents. Over time, the set of robots gradually transform into the target robot.
This allows the policy to adapt to a diverse set of robot transition dynamics within a local range. 
To improve sample efficiency, we propose an evolution reward shaping technique where we enforce larger weights on the reward received from more ``evolved'' robots to encourage the policy to adapt towards the target robots.
We present theoretical results to show that this strategy improves the adaptation.

We develop the continuous robot evolution models on a diverse set of robots and demonstrate the effectiveness of the proposed policy transfer approach with three different RL algorithms.
We showcase 
our REvolveR
on three MuJoCo Gym environments \cite{mujoco:gym} with dense reward.
Our method achieves significantly higher performance than direct policy transfer and imitation learning baselines.
We also experiment on Hand Manipulation Suite tasks \cite{dapg} in sparse rewards setting.
While methods for learning from human demonstration completely fails, our method can still transfer the policy in the challenging sparse reward setting. 

We expect the new problem of robot-to-robot policy transfer as well as the proposed REvolveR framework to be the new paradigm for inter-robot transfer learning and inspire research in related domains.

\section{Related Work}

\paragraph{Morphological Evolution}
Ideas centered around evolutionary mechanisms to develop complex robot morphologies dates back to the work from Von Neumman~\cite{von1966theory}. 
The series of seminal work from Karl Sims showed how genetic algorithms can be leveraged to develop both complex morphologies as well as their controllers through an evolutionary optimization process~\cite{sims1994evolving1,sims1994evolving2}. 
Morphological changes at evolutionary scales have also been related to development during the life of the organism and how are these related to each other~\cite{clune2012ontogeny,kriegman2018morphological}. 
Our work instead assumes that the source and target robots are given and 
figures out how to 
evolves latter from the former to transfer the controller policy.

\paragraph{Learning Controllers for Diverse Robot Morphology}
It is often difficult to design controllers for complex robots. Learning controllers via a curriculum of robots with gradually growing complexity provides a path towards controlling high-dimensional robot morphologies. This concept has been used by a recent line of work that grows control and morphology simultaneously. For instance, \citet{pathak2019learning} learns to control and develop different morphologies simultaneously to build agents that can generalize to new scenarios using dynamic graph neural networks. Vanilla GNNs~\cite{scarselli2009graph} have been used to control diverse robot morphologies in NerveNet~\cite{wang2018nervenet} to control different robots obtained by growing the limbs within topology~\cite{wang2019neural,hejna2021task} or across topology~\cite{gupta2021embodied}. Learning-driven evolution could be used to improve the design as well of the agent~\cite{cheney2014unshackling,ha2017joint,ha2018reinforcement,schaff2018jointly,xinlei:evolution}. Similarly, one could also evolve the environment itself too~\cite{wang2019paired}. Another rich approach to improve the design is to evolve the robot with a predefined grammar of physical components~\cite{zhao2020robogrammar}. In contrast to these works, we do not co-develop the controller with morphology but transfer the policy from a source robot to target robot by simulating an evolutionary process. Our approach can be applied to any given robot without being limited to the robots that appear as a biproduct of co-evolution.

Closer to ours is the line of work that tries to build controllers that can work across large kind of robots. \citet{huang2020one} leverages modularity using graph neural networks across limbs of robots to train agent-agnostic policies, which have later been replaced by transformer architectures~\cite{kurin2020my}. 
Another simple way is to condition on the hardware one-hot vector if topology remains the same~\cite{chen2018hardware}. Hierarchical controllers have also been shown to be effective while transferring across morphologies~\cite{hejna2020hierarchically}. 
Our work differ from these prior works in the sense that we assume that we are already given a good controller for some morphology and we use that to generate a controller for some new robot rather than training from scratch.

\paragraph{Modular Robotics}
Another closesly related area in robotics is that of building modular components which can be used to build diverse robot morphologies. These modular systems can either be self-configurable~\cite{stoy2010self, murata2007self} or docked manually to build complex robotic shapes~\cite{yim2000polybot,wright2007design,romanishin2013m,gilpin2008miche,daudelin2018integrated}. Recent work in this direction uses model-based learning to build and control these modular robots~\cite{whitman2021learning,whitman2020modular}.

Our work converts discrete optimization to continuous optimization. Similar ideas can be found in differentiable neural architecture search \cite{zoph2016neural,darts} where neural architecture is equivalent to our robot architecture. Furthermore, our work can be viewed as a domain transfer between two MDP domains.

\section{Preliminary and Problem Statement}

\paragraph{MDP Preliminary}
We consider  an infinite-horizon Markov Decision Process (MDP) defined by $M = (\mathcal{S}, \mathcal{A}, \mathcal{T}, R, \gamma)$, where  $\mathcal{S}$ is the set of states, $\mathcal{A}$ is the set of actions, $\mathcal{T}: \mathcal{S} \times \mathcal{A} \times \mathcal{S} \rightarrow [0, 1] $ is transition dynamics with $\mathcal{T}(s, a, s^\prime)$ the probability of transitioning from state $s$ to $s^\prime$ when action $a \in \mathcal{A}$ is taken, $R: \mathcal{S} \times \mathcal{A} \rightarrow \mathbb{R}$ is the reward associated with taking action $a$ at state $s$, and $\gamma$ is the discount factor.
The set of all MDPs is $\mathcal{M}$.
We assume both the state space $\mathcal{S}$ and the action space $\mathcal{A}$ are continuous.

A policy $\pi$ is a function that maps states to a probability distribution over
actions where $\pi(a \mid s)$ is the probability of taking action $a$ at state $s$.
Given a MDP $M$ with transition $\mathcal{T}$ and policy $\pi$, let $V^{\pi, M}$ be the value function on the model $M$ and policy $\pi$, defined as:
\begin{equation}
    V^{\pi, M} (s) = \mathop{\mathbb{E}}_{ 
    \substack{a_t \sim \pi(\cdot \mid s_t)\\ s_{t+1} \sim M(\cdot \mid s_t, a_t)}}
    [\sum_{t=0}^\infty \gamma^t R(s_t, a_t) \mid s_0 = s]
\end{equation}
The optimal policy $\pi_{M}^*$ is the policy that maximize the value function on MDP $M$, defined as $\pi_{M}^*(s) = \mathop{\arg \max}_{\pi} V^{\pi, M} (s)$.
The objective of MDP optimization is to find the optimal policy under a given MDP.

\paragraph{Problem Statement}
We consider the problem of transferring a \emph{source} policy trained for one robot to a new \emph{target policy} that must work on a different robot. 
To limit the scope of this problem, we make the assumption that the two robots share the same state space $\mathcal{S}$, action space $\mathcal{A}$, reward function $\mathcal{R}$ and discount factor $\gamma$. 
The main difference between the source policy and target policy is that they are optimal for different transition dynamics.

Formally, we consider two robots represented by two MDP $M_{\text{S}}$ (\emph{source}) and $M_{\text{T}}$ (\emph{target}) respectively.
We assume the state and action space of $M_{\text{S}}$ and $M_{\text{T}}$ are shared. Given a well-trained expert policy $\pi_{M_{\text{S}}}$ on a source robot $M_{\text{S}}$, the goal is to find the optimal policy $\pi_{M_{\text{T}}}^*$ on a target robot $M_{\text{T}}$.
Though an ordinary reinforcement learning algorithm could be used to find $\pi_{M_{\text{T}}}^*$, we would like to investigate using the information in $\pi_{M_{\text{S}}}^*$ to improve the sample efficiency as well as the final performance of $\pi_{M_{\text{T}}}$.

\section{Method}

In general, transferring the policy of one robot (source) to a different robot (target) can be challenging, especially when there is a large mismatch in the dynamics of the two robots (\emph{e.g.}, different number of joints or limbs, extreme difference in limb length). However, when the difference between the dynamics of two robots is sufficiently small, we also hypothesize that it may be easier to directly transfer the policy of the source robot to the target robot. If this hypothesis is true, it stands to reason that by defining a sequence of micro-evolutionary changes of the source robots into the new dynamics of the target robot, we should be able to transfer the policy of the source robot to the target robot through incremental policy updates over that sequence. 

Motivated by this hypothesis, our strategy is to define an evolutionary sequence of dynamics models that connects the source dynamics to the target dynamics. 
Then we will incrementally optimize the source policy by interacting with each model in the sequence until the policy is able to act (near) optimally under the target dynamics.
With multiple steps of robot change and training, the robot could eventually evolve to the target robot and transfer the policy.
However, the maximum amount of changes that can preserve sufficient task completion rate and reward is unknown and could be arbitrarily small.
An overlarge change to the robot could bring it to a ``trap'' where it never receive enough reward again and completely fail in transferring the policy.

Our solution is to develop a continuous evolution model from the source to the target robot.
The continuous model allows arbitrarily small changes towards the target robot to be made and hence transfer the policy with a smoothly developed curriculum.
The overall idea is in Algorithm \ref{alg:revolver}.

\subsection{Continuous Robot Model Evolution}

Given the source robot $M_{\text{S}}$ and target robot $M_{\text{T}}$, we define a continuous function $E: [0,1]\rightarrow \mathcal{M}$ such that $E(0) = M_{\text{S}}$ and $E(1) = M_{\text{T}}$.
The function $E$ returns an interpolation between two MDPs.
Since we assume the same state, action and reward for the source and target robots, the function $E$ essentially produces a newly interpolated transition dynamics model.
For any evolution parameter $\alpha \in (0,1)$, $E(\alpha)$ is an intermediate robot between $M_{\text{S}}$ and $M_{\text{T}}$.
In general, interpolating two different robots requires both the morphology matching and kinematics interpolation.

\begin{figure}[t] 
\centering
\newcommand\morphwidth{.9}
\subfloat{
    \includegraphics[width=\morphwidth\linewidth, valign=t]{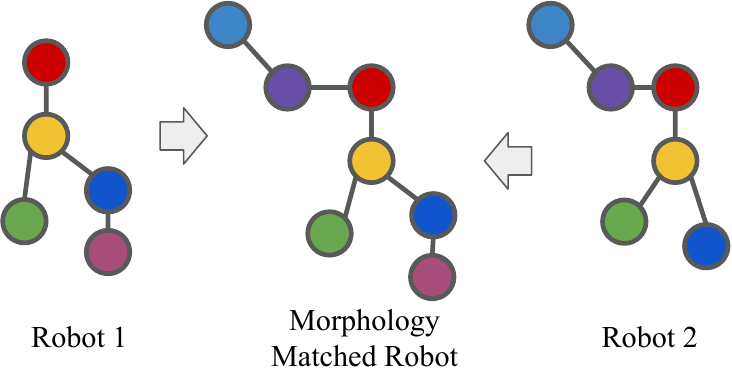}
}
\caption{ \textbf{Morphology matching of two robots.}
Though the two robots may be different in morphology (i.e. kinematic tree topology), by properly choosing a root node (e.g. the yellow node), we can always add new nodes and edges to the kinematic tree of the robot(s) to match their morphology.
}
\label{fig:morph}
\end{figure}

\paragraph{Morphology Matching}
The first step of robot interpolation is two match the morphology of the two robots.
The body and joint connection of a robot can be described by a kinematic tree.
This step essentially finds the topological matching of the kinematic trees of the two robots.
By determining the root nodes of both kinematic trees and if necessary, creating the missing nodes and/or edges, we can always find an one-to-one correspondence of node and edges between the two robots.
The procedure is illustrated in Figure \ref{fig:morph}.
In practice, however, to minimize the gap between source and target robots, we choose root nodes so that the adding of new nodes is minimal.
For example, a two-finger robot gripper could be mapped to a five-finger dexterous hand by attaching three zero sized fingers and joints to the palm node.
Creating new nodes and edges usually changes the state space $\mathcal{S}$ and action space $\mathcal{A}$ with zero insertions in the state and action vectors so that the original MDP transition dynamics $\mathcal{T}$ is not changed.

\paragraph{Kinematic Interpolation}
Given the correspondence in bodies and joints, the source and target robots may still have mismatch in other kinematics parameters that affects the physical dynamics, such as size and inertial of the bodies, and motor and damping of the joints etc.
Suppose the morphology of source and target robots is matched and the kinematic parameters of the source and target robots are mapped to the same space.
Function $E(\alpha)$ can define an interpolated robot by interpolation between all pairs of kinematics parameters.
Formally, suppose the kinematic parameters of source and target robots are $\theta_\text{S}$ and $\theta_\text{T}$ in the same space.
The parameter of the interpolated robot $M_\alpha=E(\alpha)$ is
\begin{equation}
\theta(\alpha)= (1-f(\alpha)) \theta_\text{S} + f(\alpha) \theta_\text{T}
\end{equation}
where $f:[0,1]\rightarrow[0,1]$ is a continuous function and $f(0)=0$, $f(1)=1$ to ensure $\theta(0)=\theta_\text{S}$ and $\theta(1)=\theta_\text{T}$ so that $M_0=M_\text{S}$ and $M_1=M_\text{T}$. 
In this paper, we choose to use a simple linear interpolation of $f(\alpha)=\alpha$, though a more sophisticated interpolation strategy can also be adopted.

The above morphology matching and kinematic interpolation steps can be easily implemented by editing the robots' URDF or MJCF files in physics simulation engines such as MuJoCo \cite{mujoco:gym} and pyBullet \cite{pybullet}. 
Note that evolution parameter $\alpha$ not only represents the evolution of robot hardware specifics described by a real scalar, but also can describe the continuous change of more complex hardware specifics, such as the progress of continuous mesh deformation if the shapes of the two corresponding robot bodies are different.

\subsection{Policy Transfer on Continuously Evolving Robots}\label{sec:policy:evolve}

Suppose a well-trained policy $\pi_{E(0)}$ for source robot $E(0)$ is given. 
Instead of directly transferring $\pi_{E(0)}$ to robot $E(1)$, we decompose the problem into $K$ phases of policy optimization.
In $k$-th phase, the policy is trained on robot $E(\alpha_k)$ with evolution parameter $\alpha_k = \sum_{i=1}^k l_i$, where $l_i$ is a small positive real number representing the progression of evolution parameter in $i$-th phase.
The optimization objective in the $(k+1)$-th phase is
\begin{equation} \label{eq:det:opt}
\begin{split}
    & \pi_{E(\alpha_{k+1})} = 
    \mathop{\arg \max}\limits_{\pi}
    \mathop{\mathbb{E}}_{ 
    \substack{a_t \sim \pi(\cdot \mid s_t)\\ s_{t+1} \sim M_{\alpha_{k+1}}(\cdot \mid s_t, a_t)}}
    \sum_{t} \gamma^t r_t 
\end{split}
\end{equation}
where $M_{\alpha_{k+1}} = E(\alpha_{k+1})$ is the next evolved robot and $\pi$ is initialized to be $\pi_{E(\alpha_{k})}$ at the start of the optimization.
By definition of the problem, we have $\alpha_K = \sum_{i=1}^K l_i=1$.
The $K$ and $l_k$ values are set such that $l_k$ is small enough so that during the policy optimization in Equation \eqref{eq:det:opt}, there exists sufficient amount of positive reward in the rollouts to allow policy training.
Ideally, $l_k$ values are maximized so that the number of optimization phases $K$ and the total number of RL iterations can minimized.

However, it is not possible to foresee the maximum allowed $l_k$ before training the policy on robot $E(\sum_{i=1}^k l_i)$.
In fact, there is a dilemma of trade-off between the value of $l_k$ and the total number of RL iterations:
with large $l_k$ and aggressive progression of $\alpha_k$, the policy may not receive enough positive reward from the new robot to be trained and adapted, 
and with small $l_k$ and conservative progression of $\alpha_k$, the policy may waste RL iterations on tiny robot changes.

\begin{algorithm}[t]
\caption{Continuous Robot Evolution Policy Transfer}
\label{alg:revolver}
\begin{algorithmic}[1]
\STATE{\textbf{Notation Summary:}}
\STATE{$\alpha \in [0,1]$: robot evolution parameter}
\STATE{$E: [0,1]\rightarrow \mathcal{M}$: continuous robot evolution model}
\STATE{$\pi_{E(0)}$: expert policy on the source robot $E(0)$}
\STATE{$\mathcal{R}$: replay buffer buffer}
\STATE{$\xi \in \mathbb{R}^+$: range of sampling of evolution parameters}
\STATE{$l_k \in \mathbb{R}^+$: progression of $\alpha$ in phase $k$, where $l_k<\xi$}
\STATE{$h \in \mathbb{R}^+$: evolution reward shaping factor}
\\\hrulefill
\STATE{\textcolor{darkgray}{// initialize evolution parameter, policy and replay buffer}}
\STATE{$\alpha \leftarrow 0$, $\pi \leftarrow \pi_{E(0)}$, $\mathcal{R} \leftarrow \emptyset$}
\WHILE{$\alpha < 1$}
    \FOR{epoch \textbf{in} $0,1,\ldots,N_e$}
        \STATE{\textcolor{darkgray}{// sample an intermediate robot}}
        \STATE{ $\beta \sim \text{Uniform}(\alpha, \min \{ \alpha+\xi, 1 \} )$ }
        \STATE{$M_\beta \leftarrow E(\beta)$}
        \STATE{$s_0^{\beta} \sim s_0$ \textcolor{darkgray}{// initial state distribution}}
        \FOR{$t=0,1,\ldots,N$}
            \STATE{ \textcolor{darkgray}{// execute current policy on the sampled robot and store the transition tuple to replay buffer}}
            \STATE{$a_t^{\beta} \sim \pi(\cdot \mid s_t^{\beta})$}
            \STATE{($s_{t+1}^{\beta}, r_t^{\beta}) \sim M_\beta(\cdot \mid s_t^{\beta}, a_t^{\beta})$}
        \STATE{\textcolor{darkgray}{// local reward shaping} }
        \STATE{${r_t^\prime}^{\beta}\leftarrow r_t^{\beta} \cdot \exp(h \cdot \beta)$ }
            \STATE{$\mathcal{R} \leftarrow \mathcal{R} \cup \{(s_t^{\beta}, a_t^{\beta}, s_{t+1}^{\beta}, {r_t^\prime}^{\beta})\}$}
            \STATE{sample $\{(s, a, s^\prime, r^\prime)\} \sim \mathcal{R}$}
            \STATE{train $\pi$ with $\{(s, a, s^\prime, r^\prime)\}$ using RL}
        \ENDFOR
    \ENDFOR
    \STATE{ \textcolor{darkgray}{// progress evolution parameter}}
    \STATE{$\alpha \leftarrow \alpha + l_k$}
    \STATE{ \textcolor{darkgray}{// clean up replay buffer}}
    \STATE{$\mathcal{R} \leftarrow \{ (s_t^{\beta}, a_t^{\beta}, s_{t+1}^{\beta}, {r_t^\prime}^{\beta}) \in \mathcal{R}, \forall \beta \in [\alpha, \alpha+\xi]\}$}
\ENDWHILE
\STATE{\textbf{return} $\pi$}
\end{algorithmic}
\end{algorithm}

\paragraph{Local Randomized Evolution Progression }
We propose a randomized approach to address the above problem.
At phase $k+1$, instead of repetitively choosing a deterministic progressed $\alpha_{k+1}$, we uniformly sample progressed evolution parameter $\beta$ from a local neighborhood $[\alpha_k, \alpha_k + \xi]$ where $\xi \in \mathbb{R}^+$ and $\xi > l_k$, and train policy on the rollouts of robot $M_\beta = E(\beta)$.
The optimization objective in Equation \eqref{eq:det:opt} is updated to be
\begin{equation} \label{eq:random:opt}
\begin{split}
    & \pi_{E(\alpha_{k+1})} = 
    \mathop{\arg \max}\limits_{\pi}
    \mathop{\mathbb{E}}_{ 
    \substack{
    \beta \sim U(\alpha_k, \alpha_k + \xi) \\
    M_\beta = E(\beta)
    }}
    \mathop{\mathbb{E}}_{ 
    \substack{a_t \sim \pi(\cdot \mid s_t) \\ 
    s_{t+1} \sim M_\beta(\cdot \mid s_t, a_t)}}
    \sum_{t} \gamma^t r_t 
\end{split}
\end{equation}
where $U(p, q)$ denotes the uniform distribution over $[p,q] \subset \mathbb{R}$.
The above randomized progression strategy allows the policy to be trained on sampled robots with small evolution to maintain sufficient sample efficiency and ensure adaptation, while also giving the policy a chance to risk on the robots with large evolution to improve the efficiency of policy transfer.
Note that we choose the neighborhood size $\xi$ to be larger than the progression step size $l_k$, which enables the policy to experience and explore more evolved robots with probability in advance.
As shown in Section \ref{sec:exp}, the local randomized progression strategy improves the stability of training and achieves significantly higher performance.

We point out that the similar idea of domain randomization, i.e. fine-tuning neural networks 
on randomized domains, can also be seen in the literatures on Sim2Real domain transfer such as \cite{domain:randomization} and \cite{cad2rl}.
Our robot-to-robot evolution approach can be also viewed as a series of mini robot domain randomization where the domain window $[\alpha_k, \alpha_k + \xi]$ is gradually shifting from the source robot $E(0)$ towards the target robot $E(1)$.

\subsection{Evolution Reward Shaping}\label{sec:reward:shaping}

During local randomized evolution, to better adapt the policy towards the goal of the target robot, it is helpful to encourage the policy to give more weight to robots with larger $\alpha$. 
To implement this, we devise a strategy to reshape the reward by making it a function of the evolution parameter $\alpha$. 
Specifically, we scale the reward $r_t$ received from rollouts on robot $\alpha$ to be
\begin{equation} \label{eq:local:reward:shaping}
    r_t^\prime =
    r_t \cdot \exp(h \cdot \alpha)
\end{equation}
where the evolution reward shaping factor $h\in \mathbb{R}_{\ge 0}$ controls the weight applied to the reward.
The optimization objective in Equation \eqref{eq:random:opt} is then updated to be
\begin{equation} \label{eq:local:reward:opt}
\begin{split}
    \pi_{E(\alpha_{k+1})} = 
    \mathop{\arg \max}\limits_{\pi}
    \mathop{\mathbb{E}}_{ 
    \substack{
    \beta \sim U(\alpha_k, \alpha_k+\xi) \\
    M_\beta = E(\beta)
    }}
    \mathop{\mathbb{E}}_{ 
    \substack{a_t \sim \pi(\cdot \mid s_t) \\ 
    s_{t+1} \sim M_\beta(\cdot \mid s_t, a_t)}}
    \sum_{t} \gamma^t r_t^\prime 
\end{split}
\end{equation}
How should one use the $h$ to control the weight on more evolved robots in practice? 
Under reasonable assumptions, we show the relation between the evolution reward shaping factor $h$ and the resulted change of optimization objective with the following theorem.

\begin{theorem}\label{thm:revolver}

Suppose the policy that optimizes the objective in Equation \eqref{eq:local:reward:opt} with evolution reward shaping factor of $h$ is the optimal policy $\pi_{M_\varphi}^*$ on robot $M_\varphi=E(\varphi), \varphi \in [\alpha_k, \alpha_k + \xi]$, i.e. 
\begin{equation}\label{eq:thm}
\begin{split}
    & 
    \mathop{\arg\max}_{\pi}
    \mathop{\mathbb{E}}_{ 
    \substack{
    \beta \sim U(\alpha_k, \alpha_k + \xi) \\
    M_\beta = E(\beta)
    }}
    \mathop{\mathbb{E}}_{ 
    \substack{a_t \sim \pi(\cdot \mid s_t) \\ 
    s_{t+1} \sim M_\beta(\cdot \mid s_t, a_t)}}
    \sum_{t} \gamma^t r_t \exp(h \cdot \beta)
    \\
    & = \pi_{M_\varphi}^* = \mathop{\arg\max}_{\pi}
    \mathop{\mathbb{E}}_{ 
    \substack{a_t \sim \pi(\cdot \mid s_t) \\ 
    s_{t+1} \sim M_\varphi(\cdot \mid s_t, a_t)}}
    \sum_{t} \gamma^t r_t
\end{split}
\end{equation}
Then when $\xi\rightarrow 0$, $\varphi = \alpha_k + \frac{1}{2}\xi + \frac{1}{4} h\xi^2 + o(\xi^2)$.  
\end{theorem}

The proof of Theorem \ref{thm:revolver} is in the Section \ref{sec:thm:proof}.
Theorem \ref{thm:revolver} shows that a positive evolution reward shaping factor shifts the objective of policy optimization towards the direction of the target robot compared to setting $h=0$.
This allows the policy to give more weight to the experiences gained with ``more evolved'' robots, without sacrificing sample efficiency due to changes in the sampling distribution.

\subsection{Other Implementation Details}\label{sec:other:implementation}

\paragraph{Adaptive Training Scheduling}
From the description so far, the training scheduling of robot evolution parameters $\alpha_k$ has been fixed.
Moreover, the number of epochs trained in each evolution phase (i.e. $N_e$ in Algorithm \ref{alg:revolver}) is also fixed.
In practice, however, we can dynamically schedule the training by changing both hyperparameters on the fly, especially when the difficulty of each transfer phase is different.
Determining the progression step size $l_k$ or the next $\alpha_k$ is usually hard since the training results are usually not predictable.
A more practical strategy is to fix all $l_k$ while setting the initial value of $N_e$ to be small in each phase.
When the training of policy struggles during the current phase, e.g. the reward or success rate drops significantly compared to previous phases, we iteratively increase $N_e$ until the policy performance is sufficient to move on to the next phase.
We adopted this strategy in the experiments in Section \ref{sec:dapg:exp}.

\begin{figure*}[t]
\centering
\small
\newcommand\antwidth{0.14}
\newcommand\humanoidwidth{0.14}
\subfloat[\texttt{Ant-length-mass} environment.]{
\begin{tabular}{c}
\addvbuffer[0pt 0pt]{\includegraphics[width=\antwidth\textwidth, valign=m]{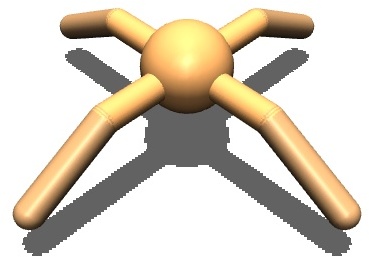}} $\rightarrow$ \addvbuffer[0pt 0pt]{\includegraphics[width=\antwidth\textwidth,valign=m]{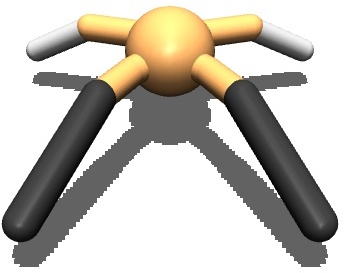}}
\end{tabular} 
\label{fig:ant:length:mass}
} 
\hspace{5ex}
\subfloat[\texttt{Ant-length-mass} policy transfer experiment results. ]{
\begin{tabular}{l|c|c}
\hline
& TD3 & SAC \\ 
\hline
Expert on Source & 6826.52 & 6985.90 \\
\specialrule{.12em}{.05em}{.05em}
From Scratch & 4644.09 $\pm$ 502.05 & 5908.51 $\pm$ 339.81 \\ \hline
Direct Transfer & 4903.77 $\pm$ 801.12 & 6194.55 $\pm$ 165.82 \\ \hline
SOIL & 4891.67 $\pm$ 819.18 & 6061.32 $\pm$ 1102.58 \\ \hline
\textbf{Ours} & \textbf{5903.28 $\pm$ 416.07} & \textbf{6473.21 $\pm$ 207.98} \\ \hline
\end{tabular} 
\label{tab:ant:length:mass}
} 
\\
\subfloat[\texttt{Humanoid-length-mass} environment.]{
\begin{tabular}{c}
\addvbuffer[0pt 0pt]{\includegraphics[height=\humanoidwidth\textwidth, valign=m]{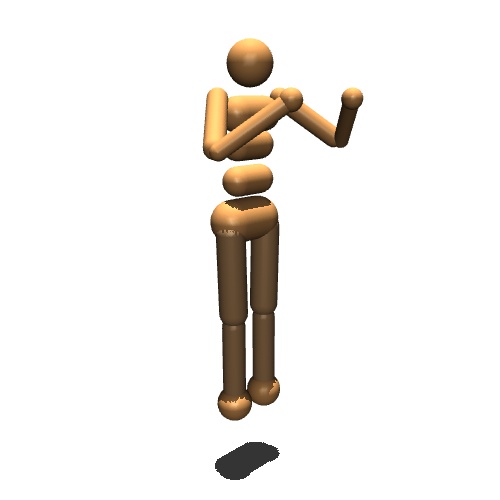}} $\rightarrow$ \addvbuffer[3pt 0pt]{\includegraphics[height=\humanoidwidth\textwidth,valign=m]{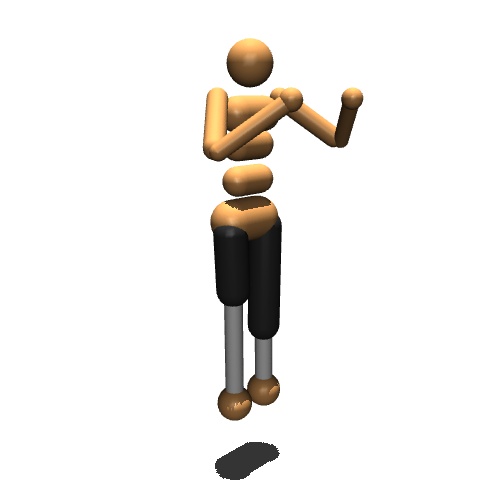}} 
\end{tabular} 
\label{fig:humanoid:length:mass}
} 
\hspace{5ex}
\subfloat[\texttt{Humanoid-length-mass} policy transfer experiment results. ]{
\begin{tabular}{l|c|c}
\hline
 & TD3 & SAC \\ 
\hline
Expert on Source & 6663.25 & 8271.70 \\
\specialrule{.1em}{.05em}{.05em}
From Scratch & 5824.07 $\pm$ 233.46 & 6468.60 $\pm$ 157.26 \\ \hline
Direct Transfer & 6256.15 $\pm$ 253.63 & 7639.61 $\pm$ 278.02 \\ \hline
SOIL & 6414.25 $\pm$ 505.79 & 6970.15 $\pm$ 659.32 \\ \hline
\textbf{Ours} & \textbf{7386.44 $\pm$ 151.24} & \textbf{7986.97 $\pm$ 129.21} \\ \hline
\end{tabular}
\label{tab:humanoid:length:mass}
} 
\\
\subfloat[\texttt{Ant-leg-emerge} environment.]{
\begin{tabular}{c}
\addvbuffer[0pt 0pt]{\includegraphics[width=\antwidth\textwidth, valign=m]{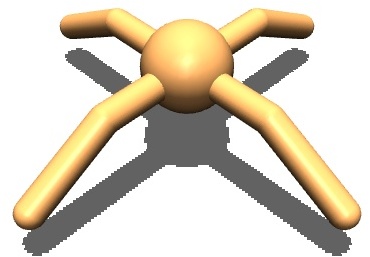}} $\rightarrow$ \addvbuffer[3pt 0pt]{\includegraphics[width=\antwidth\textwidth,valign=m]{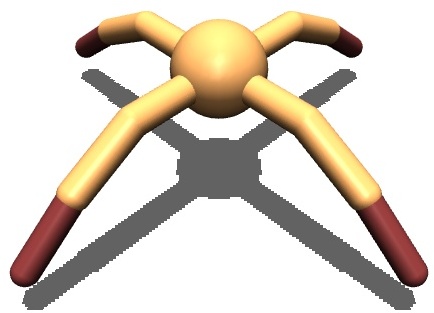}} \\
\end{tabular}
\label{fig:ant:leg:emerge}
} 
\hspace{5ex}
\subfloat[\texttt{Ant-leg-emerge} policy transfer experiment results.]{
\begin{tabular}{l|c|c}
\hline
& TD3 & SAC \\
\hline
Expert on Source & 6591.832 & 6431.32 \\
\specialrule{.1em}{.05em}{.05em}
From Scratch & 2551.30 $\pm$ 316.95 & 3845.69 $\pm$ 243.92 \\ \hline
Direct Transfer & 3579.40 $\pm$ 118.06 & 6031.27 $\pm$ 320.81  \\ \hline
SOIL & 1703.62 $\pm$ 351.83 & 4533.09 $\pm$ 578.32 \\ \hline
\textbf{Ours} &  \textbf{4688.68 $\pm$ 270.51} & \textbf{6612.78 $\pm$ 264.50} \\ \hline
\end{tabular}
\label{tab:ant:leg:emerge}
}
\caption{
\textbf{Experiments on source-to-target policy transfer on the MuJoCo Gym environments.}
All methods are trained for three million iterations with five different seeds. 
Mean and standard deviation of the reward of an epoch are reported. Our approach outperforms the baselines across different policy optimization schemes and across environments.
}
\label{fig:gym:combine}
\end{figure*}

\paragraph{Replay Buffer Cleaning}
As the policy optimization moves on to the next phase, the past transition sampled from less evolved robots in previous phases are outdated and should no longer be used. To implement this, we remove transition tuples that are older than the current interpolation range from the replay buffer upon entering the next phase.

\section{Experiments}\label{sec:exp}

The design of our REvolveR framework is motivated by the hypothesis that compared to directly transferring the policy from source to target robot, transferring the policy through a sequence of micro-evolutionary changes of robot dynamics is an easier task and achieves better sample efficiency and performance.
To show this, we apply our REvolveR to two sets of robotic control tasks:
MuJoCo Gym environments \cite{mujoco:gym}, and
Hand Manipulation Suite \cite{dapg}.
We compare the performance to a variety of baselines including training policy from scratch, direct policy transfer, and imitation learning methods.

\subsection{MuJoCo Gym Environments } \label{sec:gym}

\paragraph{Environments and Rewards}
We adopt the default Ant-v2 and Humanoid-v2 robots from MuJoCo Gym \cite{mujoco:gym} as our source robots. 
We construct three environments where the target robots are created by continuously modifying some properties of the source robots.
In \texttt{Ant-length-mass} and \texttt{Humanoid-length-mass} environments, the mass and lengths of all leg bodies are changed;
In \texttt{Ant-leg-emerge} environment, new legs and joints grow from the tip of the toe.
The robot evolution is illustrated Figures \ref{fig:teaser} and \ref{fig:gym:combine}\subref{fig:ant:length:mass}\subref{fig:humanoid:length:mass}\subref{fig:ant:leg:emerge}.

\paragraph{Reward Function}

In all three environments, the robot agents get rewarded by the distance they moved forward.
Specially, in \texttt{Humanoid-length-mass} environment, the robot agents are heavily penalized for falling down.
The reward function is the same for source, target and all intermediate robots.

\paragraph{RL Algorithms} 
We use two state-of-the-art actor-critic reinforcement learning algorithms, TD3 \cite{td3} and SAC \cite{sac}, in our experiments.
We first train both RL algorithms on the source robots until convergence and use the well-trained policy as the expert policy to be transferred to target robot.
During transfer, both the actor and critic are updated.
Note that the expert policy performance will be different for the two source robots because \texttt{Ant-leg-emerge} robots have an additional leg and therefore different state space.

\paragraph{Baselines} 
We compare our method with the following baselines for learning a policy on the target robot.
\begin{itemize}[leftmargin=3ex,topsep=0ex]
\setlength{\itemsep}{0pt}
\setlength{\parskip}{0pt}
\setlength{\parsep}{-100pt}
    \item \textit{From Scratch}: we train policy on the target robot from scratch with the RL algorithm.
    \item \textit{Direct Transfer}: we initialize the target robot policy with the expert policy on source robot and fine-tune the policy directly on the target robot.
    \item  \textit{State-only Imitation Learning (SOIL) \cite{soil}}:
    SOIL is an imitation learning method.
    It trains an inverse dynamics model to match the distribution of the next state between the student and teacher agents. 
    Then it augments the policy gradient with a term that aims to maximize the probability of the predicted actions from inverse dynamics.
    In our experiments, the source robot is the teacher and the target robot is the student.
\end{itemize}

The above baseline methods are trained on the target robots for three million RL iterations on the \texttt{Ant-length-mass} and \texttt{Humanoid-length-mass} tasks and one million iterations on the \texttt{Ant-leg-emerge} task.
We train our REvolveR for the same 
RL iterations as the baselines in total, not only on the target robot but on all intermediate robots during evolution.
All methods are trained with five different random seeds.
The experiment results are illustrated in Tables \ref{fig:gym:combine}\subref{tab:ant:length:mass}\subref{tab:humanoid:length:mass}\subref{tab:ant:leg:emerge} \footnote{We did not use the standard bold-line/shaded-area curves to illustrate the policy performance, because unlike ordinary RL algorithms that are trained on a single robot and have performance vs. time results, our REvolveR is only able to deliver \textbf{valid target robot performance} at the end of the policy transfer. }.

\paragraph{Results and Analysis}
Our REvolveR framework outperforms all related baselines by a notable margin in terms of episode reward, especially on \texttt{Ant-leg-emerge} and \texttt{Humanoid-length-mass}.

On \texttt{Ant-leg-emerge} environment, an interesting finding is that the performance of SOIL is even worse than directly transferring the policy.
An explanation is that the emerging legs of the source robot have length and mass close to zero and show random behaviors with expert policy.
Though the random behavior does not affect the source robot, it causes it to struggle at the start of training when directly transferred to target robot.  

On \texttt{Humanoid-length-mass} environment, the expert policy is able to control the source humanoid robot to both stand and jog for higher rewards.
However, due to dynamics mismatch, source expert policy cannot support target humanoid robot to stand.
When directly trained on the target robot, even with source expert policy provided, all the baseline methods learned to discard jogging to learn standing first  due to heavy penalty on falling down.
As a comparison, when transferring the policy through continuously evolving intermediate robots with our REvolveR, both standing and jogging skills can be kept and smoothly transferred, which highlights the advantage of our method.

\begin{table}[t]
\centering
\small
\newcommand\dapgwidth{0.15}

\subfloat[\texttt{Hammer} task experiment results. ]{
\begin{tabular}{l|c|c}
\hline
& Dense Reward & Sparse Reward \\ \hline
From Scratch & $>$100K & $\infty$ \\ \hline
Direct Finetune & $>$100K & $\infty$ \\ \hline
DAPG & 17.1K &  $\infty$  \\ \hline
\textbf{Ours} & - & \textbf{11.9K} \\ \hline
\end{tabular}
}
 
\vspace{-1ex}
\\
\subfloat[\texttt{Relocate} task experiment results. ]{
\begin{tabular}{l|c|c}
\hline
& Dense Reward & Sparse Reward \\
\hline
From Scratch & $>$100K & $\infty$ \\ \hline
Direct Finetune & 43.5K & $\infty$ \\ \hline
DAPG & 23.3K &  $\infty$  \\ \hline
\textbf{Ours} & - & \textbf{18.1K} \\ \hline
\end{tabular}
}
\vspace{-1ex}
\\
\subfloat[\texttt{Door} task experiment results.]{
\begin{tabular}{l|c|c}
\hline
& Dense Reward & Sparse Reward \\ \hline
From Scratch & - & $\infty$ \\ \hline
Direct Finetune & 7.6K & $\infty$ \\ \hline
DAPG & 5.4K &  $\infty$  \\ \hline
\textbf{Ours} & - & \textbf{2.6K} \\ \hline
\end{tabular}
}

\vspace{-1ex}
\caption{\textbf{Experiments on the Hand Manipulation Suite}.
The evaluation metrics is the number of epochs needed to reach 90\% task success rate. Our method with sparse reward outperforms all the baselines even with dense reward.
}
\label{tab:dapg:result}
\end{table}
\begin{figure}[ht]
\centering
\newcommand\dapgwidth{0.14}
\newcommand\dapgoffset{-3.2}
\captionsetup[subfigure]{oneside,margin={0cm,0cm}}
\subfloat[\texttt{Hammer}. ]{
\begin{tabular}{c}
\includegraphics[width=\dapgwidth\textwidth]{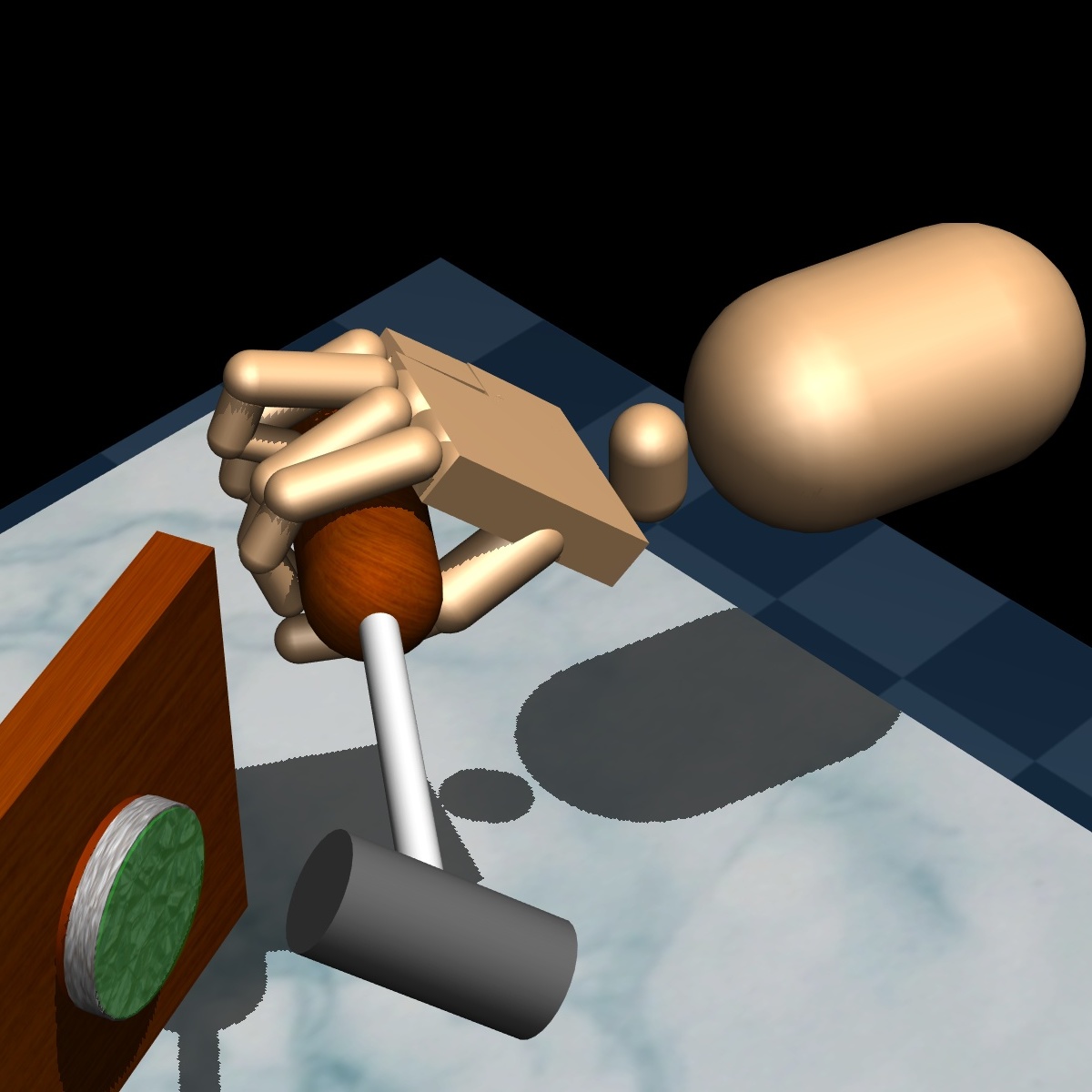} \\
$\downarrow$ \\
\includegraphics[width=\dapgwidth\textwidth]{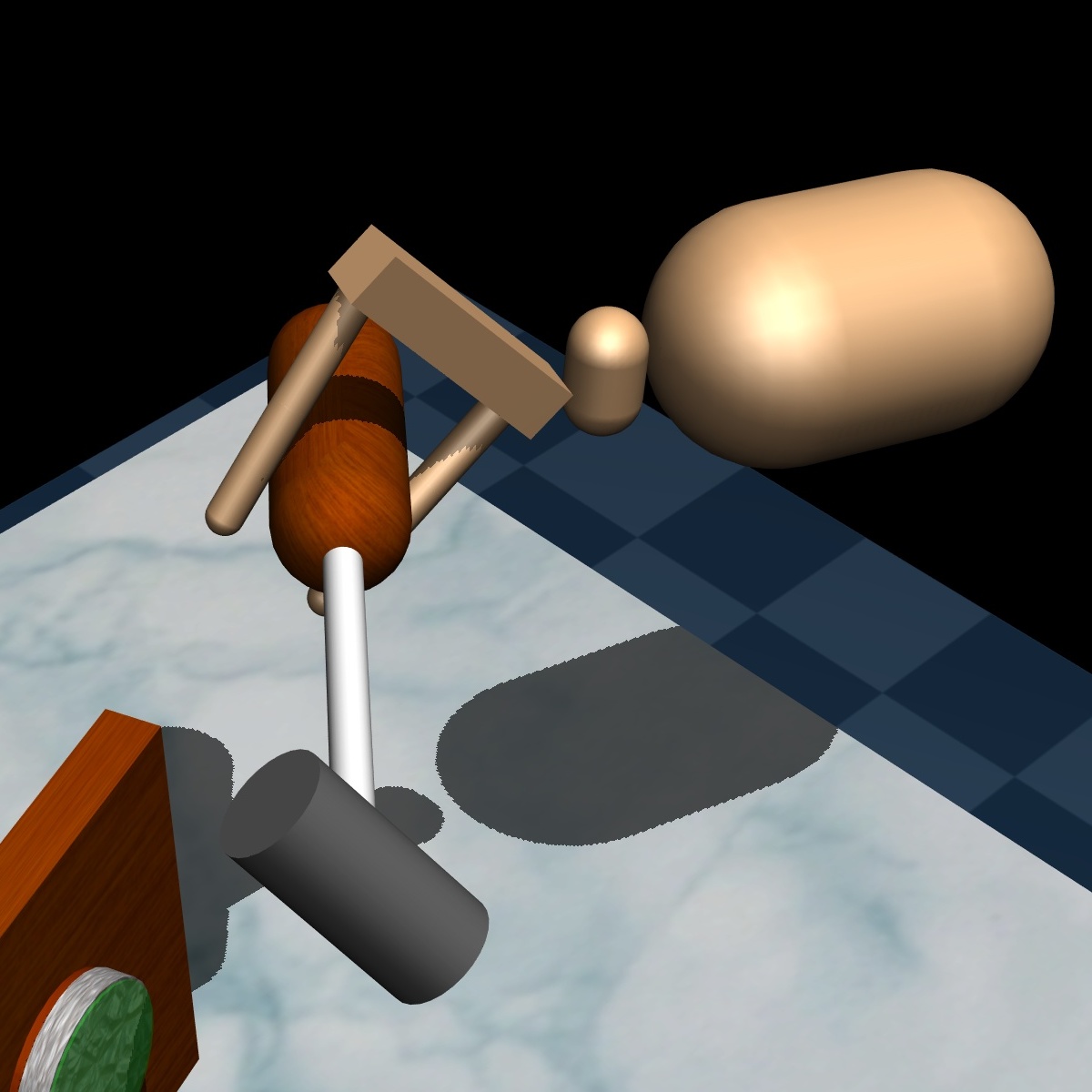}
\end{tabular}
}
\hspace{\dapgoffset ex}
\subfloat[\texttt{Relocate}. ]{
\begin{tabular}{c}
\includegraphics[width=\dapgwidth\textwidth]{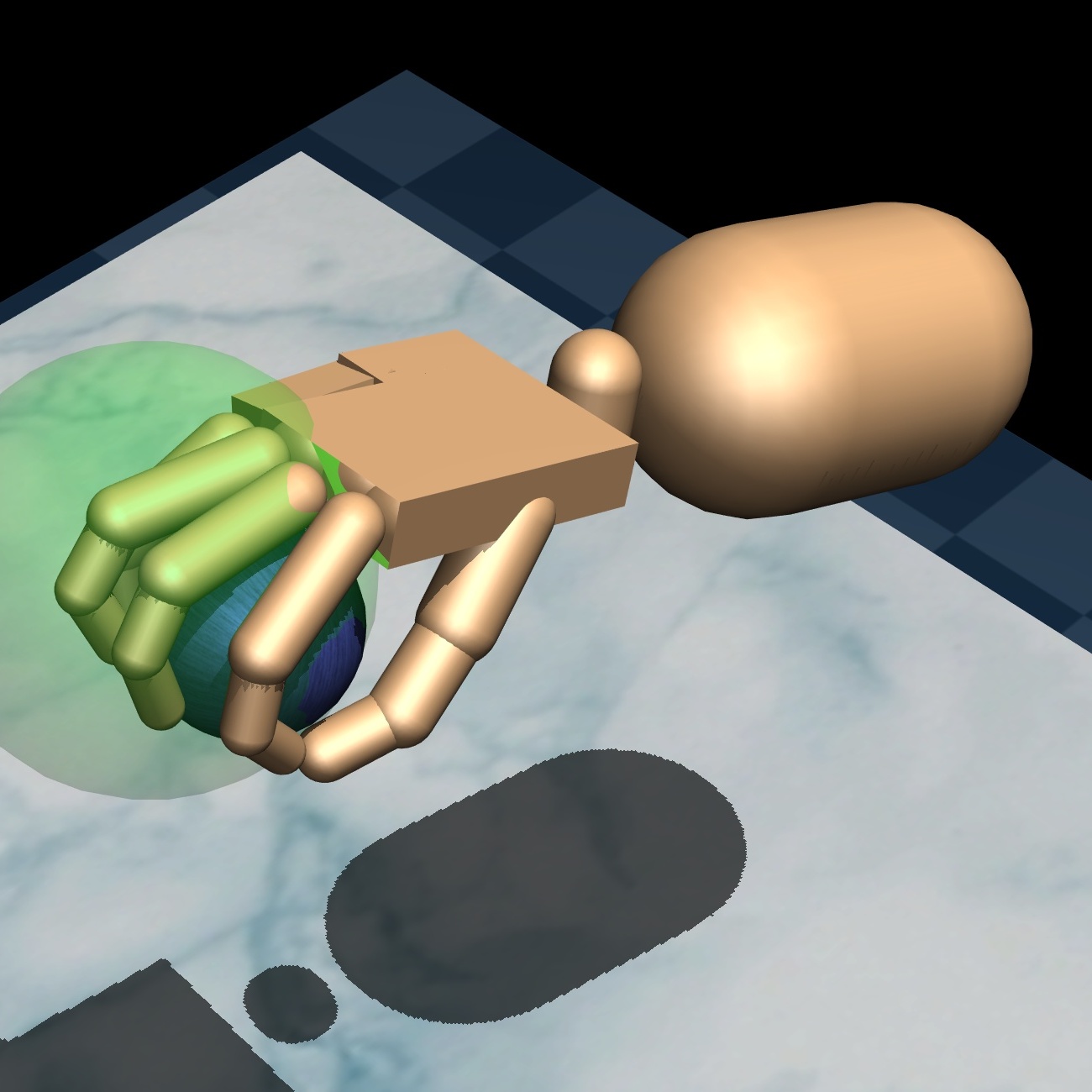} \\
$\downarrow$ \\
\includegraphics[width=\dapgwidth\textwidth]{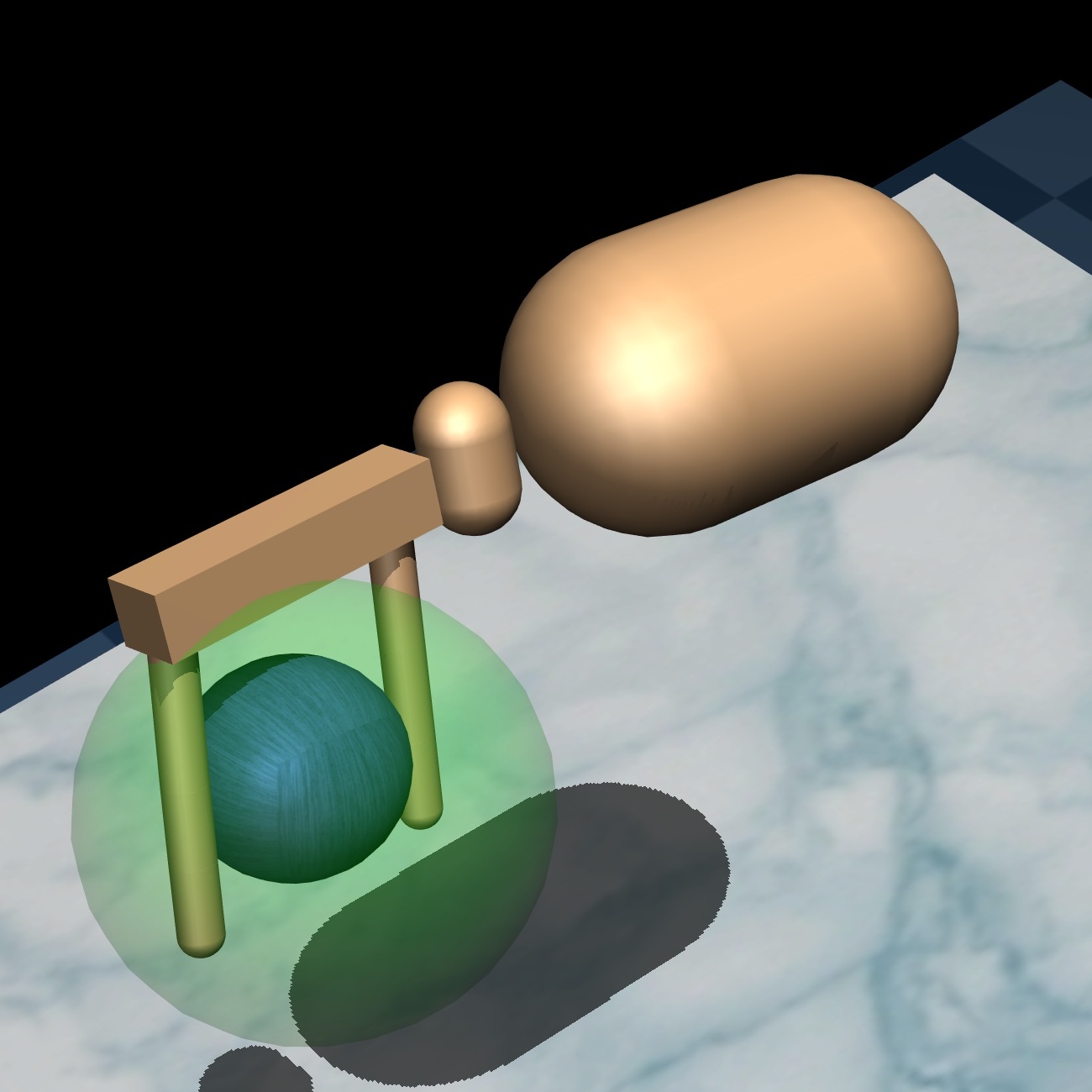}
\end{tabular}
}
\hspace{\dapgoffset ex}
\subfloat[\texttt{Door}. ]{
\begin{tabular}{c}
\includegraphics[width=\dapgwidth\textwidth]{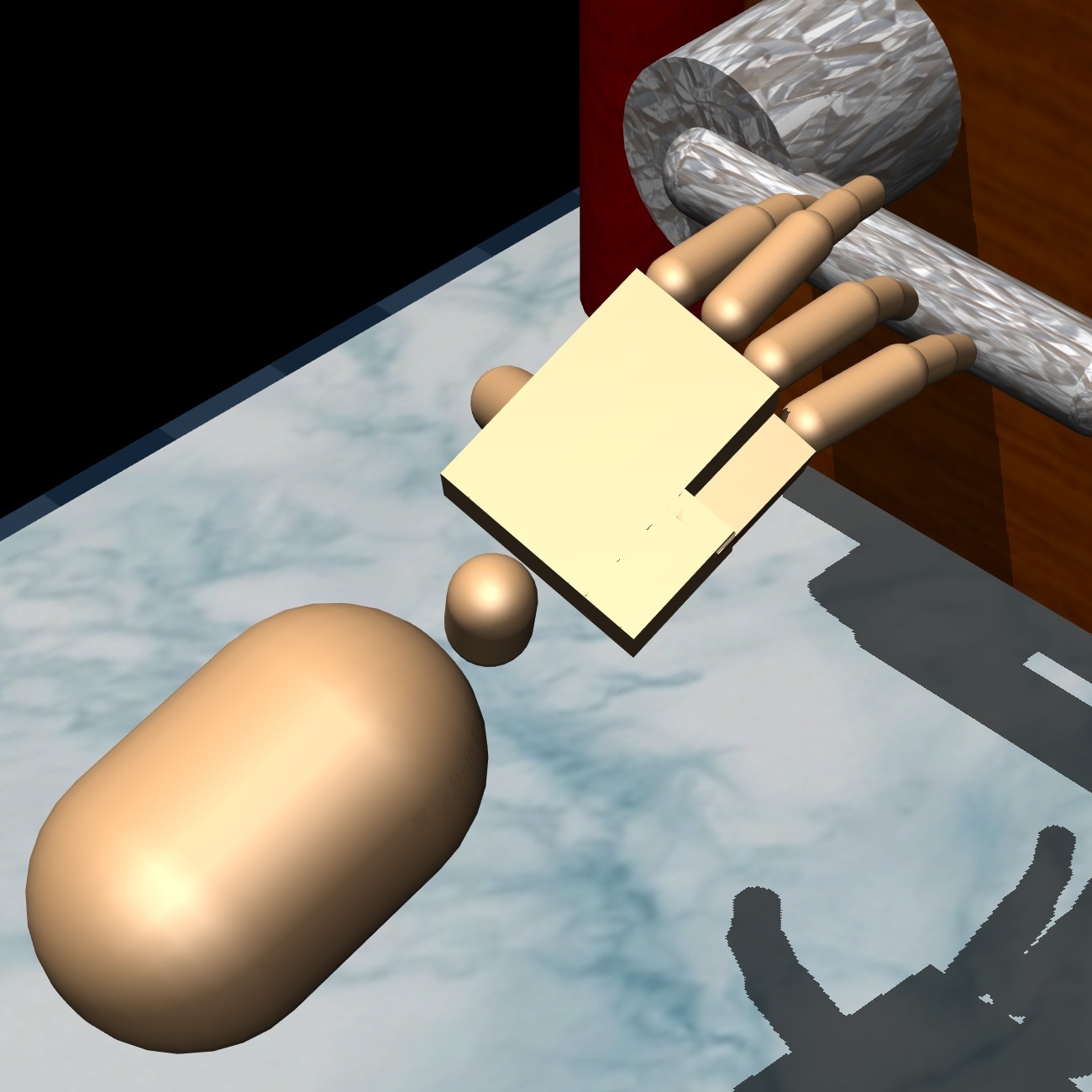} \\
$\downarrow$ \\
\includegraphics[width=\dapgwidth\textwidth]{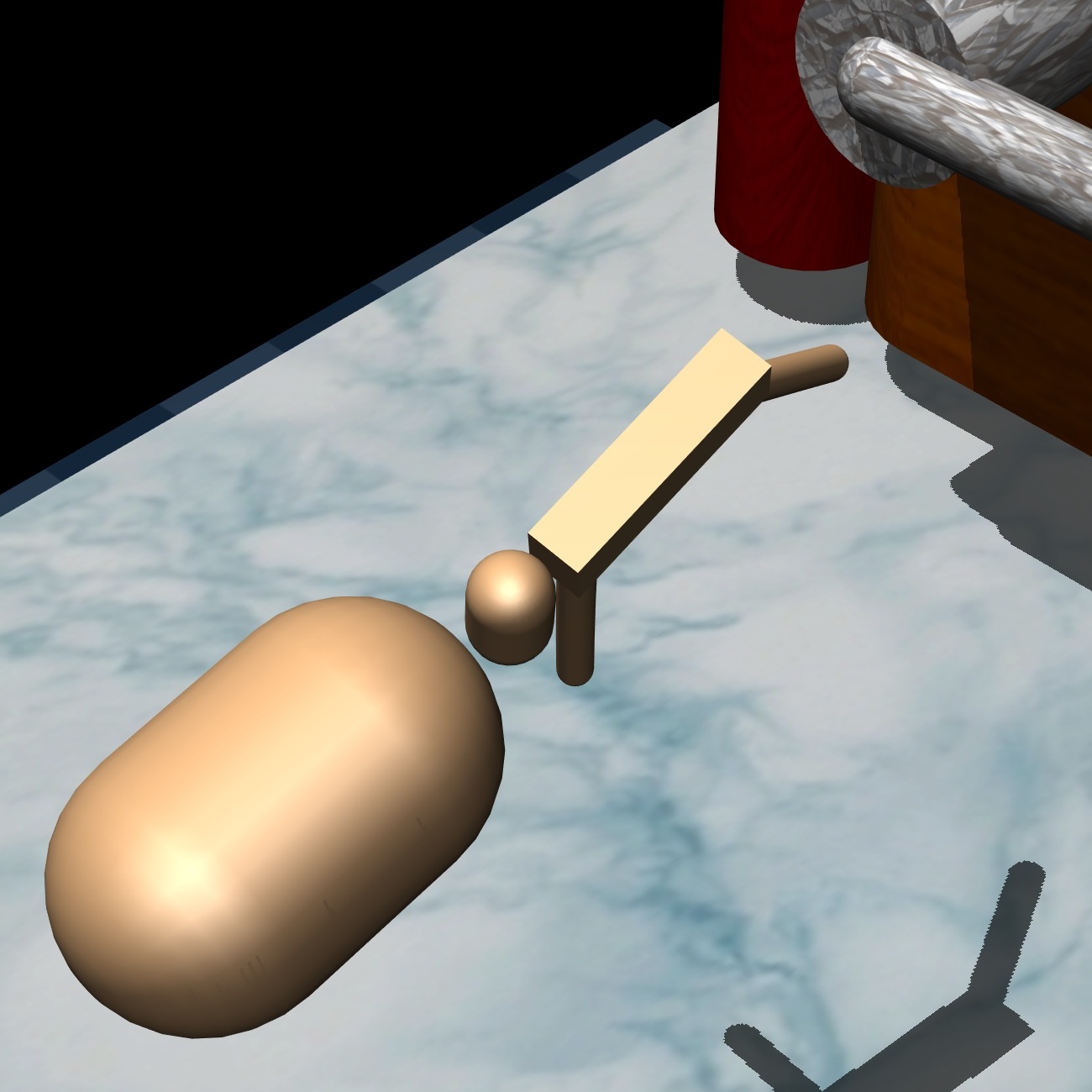}
\end{tabular}
}
\vspace{-1ex}
\caption{\textbf{Hand Manipulation Suite tasks.}
(a) \texttt{Hammer} environment: the goal is to pick up the hammer and smash the nail into the board;
(b) \texttt{Relocate} environment: the goal is to pick up the blue ball and take it to the desired site shown by the green semitransparent sphere;
(c) \texttt{Door} environment: the goal is to switch the latch and fully open the door.
}
\label{fig:dapg}
\end{figure}

\subsection{Hand Manipulation Suite}\label{sec:dapg:exp}

\paragraph{Robot Evolution}

We adopt the five-finger dexterous hand provided in the ADROIT platform \cite{adroit} as our source robot and follow \citet{dapg} for the environment settings.
The robot evolve to the target robot of a two-finger gripper by gradually shrinking three fingers except the thumb and index finger.
The robot evolution is illustrated Figures \ref{fig:teaser} and \ref{fig:dapg}.

\paragraph{Task and Reward Function}

We use the three tasks from the proposed suite in \cite{dapg}: 
\texttt{Hammer}, \texttt{Relocate} and \texttt{Door}.
In \texttt{Hammer}, the task is to pick up the hammer and smash the nail into the board;
in \texttt{Relocate}, the task is to pick up the ball and take it to the target position;
in \texttt{Door}, the task is to turn the door handle and fully open the door.
In a sparse reward setting, only task completion is rewarded.
In a dense reward setting, a distance reward is provided at every step.

The baselines compared against include \textit{From Scratch} and \textit{Direct Transfer} from Section \ref{sec:gym}.
We also compare against DAPG \cite{dapg} which is a variant of NPG \cite{npg} with demonstration-augmented policy gradient for learning from human demonstrations.
The expert policies for the five-finger source robot are imported from \cite{dapg} and used in all methods as needed.

For our REvolveR, we use NPG  \cite{npg} as our RL algorithm.
We use the adaptive training scheduling strategy proposed in Section \ref{sec:other:implementation} to improve training efficiency.
Therefore, the total number of RL iterations cannot be set beforehand to fairly compare the performance under the same number of iterations. 
So we instead compare in terms of the number of RL optimization steps needed to reach 90\% success rate on the tasks.
The results are illustrated in Table \ref{tab:dapg:result}. 

\newcommand\chck{{\color{black}\ding{51}}}
\newcommand\chx{{\color{black}\ding{55}}}
\begin{table}[t]
\small
\centering
\begin{tabular}{c|c|c}
\hline
\multirow{2}{*}{\begin{tabular}[c]{@{}c@{}} Randomized \\  Evolution \end{tabular}} & \multirow{2}{*}{\begin{tabular}[c]{@{}c@{}} Reward \\ Shaping Factor \end{tabular}} & \multirow{2}{*}{Reward} \\
&&\\ \hline
\chx & $h=0.0$ & 5363.89 $\pm$ 419.90 \\ \hline
\chck & $h=0.0$ & 5844.46 $\pm$ 33.63 \\ \hline
\chck & $h=1.0$ &  6279.35 $\pm$ 290.33 \\ \hline
\end{tabular}
\caption{
\textbf{Ablation studies} on local randomized evolution and evolution reward shaping.
The RL algorithm used in the experiments is SAC and the task evaluated is \texttt{Ant-length-mass}. 
}
\label{tab:ablation}
\end{table}

\paragraph{Results and Analysis}

In sparse reward case, all baselines never receive positive reward for improving itself and are ineffective in solving the tasks.
This is because due to dynamics mismatch, source expert policy cannot find a single successful trajectory on the target robot, e.g. cannot pickup the ball or hammer.
At the same time, the exploration in the high-dimension is too hard.
As a comparison, when transferring the policy through continuously evolving intermediate robots with REvolveR, the intermediate robots maintain sufficient success rate to ensure sample efficiency and successfully transfer the policy.

We show the detailed results of comparison in Table \ref{tab:dapg:result}. As the results show, when trained with sparse reward, our REvolveR even outperforms baselines trained with dense reward in terms of total number of iterations.
This again highlights the advantage of our method in terms of improving sample efficiency and performance in policy transfer.

\subsection{Ablation Studies}

We perform ablation experiments on \texttt{Ant-length-mass} environment. 
The RL algorithm used in the experiments is SAC \cite{sac}. We ablate the following two components of our method:

\paragraph{Deterministic vs. Local Randomized Evolution }
We study the effect of using local randomized evolution progression strategy proposed in Equation \eqref{eq:random:opt} and compare against deterministic evolution in Equation \eqref{eq:det:opt}.
As illustrated in Table \ref{tab:ablation}, local randomized evolution progression not only improves performance of transfer but also improves the robustness as shown by the decrease of standard deviation of the episode reward.

\paragraph{Evolution Reward Shaping }
We study the effect of evolution reward shaping as proposed in Equation \eqref{eq:local:reward:shaping} and compare against Equation \eqref{eq:random:opt} without reward shaping.
As illustrated in Table \ref{tab:ablation}, evolution reward shaping can effectively improve performance as shown by the improvement of the mean of the episode reward.

\section{Conclusion}
In this paper, we  propose a novel method named REvolveR for robotic policy transfer between two different robots so that one does not have to train a policy from scratch for every new robot.
Our method is based on continuous evolutionary models implemented in a physics simulator. 
An expert policy on the source robot can be transferred through training on a sequence of intermediate robots that evolve into the target robot.
We conduct experiments on several tasks on MuJoCo simulation engine and show that the proposed method can effectively transfer the policy across robots and achieve superior sample efficiency on new robots 
and is especially advantageous in sparse reward settings.

\section*{Acknowledgement}
This work is in part funded by JST AIP Acceleration, Grant Number JPMJCR20U1, Japan.

\bibliography{bib}

\begin{thebibliography}{48}
\providecommand{\natexlab}[1]{#1}
\providecommand{\url}[1]{\texttt{#1}}
\expandafter\ifx\csname urlstyle\endcsname\relax
  \providecommand{\doi}[1]{doi: #1}\else
  \providecommand{\doi}{doi: \begingroup \urlstyle{rm}\Url}\fi

\bibitem[Brockman et~al.(2016)Brockman, Cheung, Pettersson, Schneider,
  Schulman, Tang, and Zaremba]{mujoco:gym}
Brockman, G., Cheung, V., Pettersson, L., Schneider, J., Schulman, J., Tang,
  J., and Zaremba, W.
\newblock Openai gym, 2016.

\bibitem[Chen et~al.(2018)Chen, Murali, and Gupta]{chen2018hardware}
Chen, T., Murali, A., and Gupta, A.
\newblock Hardware conditioned policies for multi-robot transfer learning.
\newblock In \emph{Advances in Neural Information Processing Systems}, pp.\
  9355--9366, 2018.

\bibitem[Cheney et~al.(2014)Cheney, MacCurdy, Clune, and
  Lipson]{cheney2014unshackling}
Cheney, N., MacCurdy, R., Clune, J., and Lipson, H.
\newblock Unshackling evolution: evolving soft robots with multiple materials
  and a powerful generative encoding.
\newblock \emph{ACM SIGEVOlution}, 2014.

\bibitem[Clune et~al.(2012)Clune, Pennock, Ofria, and
  Lenski]{clune2012ontogeny}
Clune, J., Pennock, R.~T., Ofria, C., and Lenski, R.~E.
\newblock Ontogeny tends to recapitulate phylogeny in digital organisms.
\newblock \emph{The American Naturalist}, 180\penalty0 (3):\penalty0 E54--E63,
  2012.

\bibitem[Coumans \& Bai(2016)Coumans and Bai]{pybullet}
Coumans, E. and Bai, Y.
\newblock Pybullet, a python module for physics simulation for games, robotics
  and machine learning, 2016.

\bibitem[Daudelin et~al.(2018)Daudelin, Jing, Tosun, Yim, Kress-Gazit, and
  Campbell]{daudelin2018integrated}
Daudelin, J., Jing, G., Tosun, T., Yim, M., Kress-Gazit, H., and Campbell, M.
\newblock An integrated system for perception-driven autonomy with modular
  robots.
\newblock \emph{Science Robotics}, 2018.

\bibitem[Fujimoto et~al.(2018)Fujimoto, Hoof, and Meger]{td3}
Fujimoto, S., Hoof, H., and Meger, D.
\newblock Addressing function approximation error in actor-critic methods.
\newblock In \emph{International Conference on Machine Learning}, pp.\
  1587--1596. PMLR, 2018.

\bibitem[Gilpin et~al.(2008)Gilpin, Kotay, Rus, and Vasilescu]{gilpin2008miche}
Gilpin, K., Kotay, K., Rus, D., and Vasilescu, I.
\newblock Miche: Modular shape formation by self-disassembly.
\newblock \emph{IJRR}, 2008.

\bibitem[Gupta et~al.(2021)Gupta, Savarese, Ganguli, and
  Fei-Fei]{gupta2021embodied}
Gupta, A., Savarese, S., Ganguli, S., and Fei-Fei, L.
\newblock Embodied intelligence via learning and evolution.
\newblock \emph{arXiv preprint arXiv:2102.02202}, 2021.

\bibitem[Ha(2018)]{ha2018reinforcement}
Ha, D.
\newblock Reinforcement learning for improving agent design.
\newblock \emph{arXiv preprint arXiv:1810.03779}, 2018.

\bibitem[Ha et~al.(2017)Ha, Coros, Alspach, Kim, and Yamane]{ha2017joint}
Ha, S., Coros, S., Alspach, A., Kim, J., and Yamane, K.
\newblock Joint optimization of robot design and motion parameters using the
  implicit function theorem.
\newblock In \emph{Robotics: Science and Systems}, 2017.

\bibitem[Haarnoja et~al.(2018)Haarnoja, Zhou, Abbeel, and Levine]{sac}
Haarnoja, T., Zhou, A., Abbeel, P., and Levine, S.
\newblock Soft actor-critic: Off-policy maximum entropy deep reinforcement
  learning with a stochastic actor.
\newblock In \emph{International conference on machine learning}, pp.\
  1861--1870. PMLR, 2018.

\bibitem[Hejna et~al.(2020)Hejna, Pinto, and Abbeel]{hejna2020hierarchically}
Hejna, D., Pinto, L., and Abbeel, P.
\newblock Hierarchically decoupled imitation for morphological transfer.
\newblock In \emph{International Conference on Machine Learning}, pp.\
  4159--4171. PMLR, 2020.

\bibitem[Hejna~III et~al.(2021)Hejna~III, Abbeel, and Pinto]{hejna2021task}
Hejna~III, D.~J., Abbeel, P., and Pinto, L.
\newblock Task-agnostic morphology evolution.
\newblock \emph{arXiv preprint arXiv:2102.13100}, 2021.

\bibitem[Ho \& Ermon(2016)Ho and Ermon]{gail}
Ho, J. and Ermon, S.
\newblock Generative adversarial imitation learning.
\newblock \emph{Advances in neural information processing systems},
  29:\penalty0 4565--4573, 2016.

\bibitem[Huang et~al.(2020)Huang, Mordatch, and Pathak]{huang2020one}
Huang, W., Mordatch, I., and Pathak, D.
\newblock One policy to control them all: Shared modular policies for
  agent-agnostic control.
\newblock In \emph{International Conference on Machine Learning}, pp.\
  4455--4464. PMLR, 2020.

\bibitem[Kriegman et~al.(2018)Kriegman, Cheney, and
  Bongard]{kriegman2018morphological}
Kriegman, S., Cheney, N., and Bongard, J.
\newblock How morphological development can guide evolution.
\newblock \emph{Scientific reports}, 8\penalty0 (1):\penalty0 1--10, 2018.

\bibitem[Kumar et~al.(2013)Kumar, Xu, and Todorov]{adroit}
Kumar, V., Xu, Z., and Todorov, E.
\newblock Fast, strong and compliant pneumatic actuation for dexterous
  tendon-driven hands.
\newblock In \emph{2013 IEEE international conference on robotics and
  automation}, pp.\  1512--1519. IEEE, 2013.

\bibitem[Kurin et~al.(2020)Kurin, Igl, Rockt{\"a}schel, Boehmer, and
  Whiteson]{kurin2020my}
Kurin, V., Igl, M., Rockt{\"a}schel, T., Boehmer, W., and Whiteson, S.
\newblock My body is a cage: the role of morphology in graph-based incompatible
  control.
\newblock \emph{arXiv preprint arXiv:2010.01856}, 2020.

\bibitem[Liu et~al.(2019)Liu, Ling, Mu, and Su]{sail}
Liu, F., Ling, Z., Mu, T., and Su, H.
\newblock State alignment-based imitation learning.
\newblock \emph{arXiv preprint arXiv:1911.10947}, 2019.

\bibitem[Liu et~al.(2018)Liu, Simonyan, and Yang]{darts}
Liu, H., Simonyan, K., and Yang, Y.
\newblock Darts: Differentiable architecture search.
\newblock \emph{arXiv preprint arXiv:1806.09055}, 2018.

\bibitem[Luo et~al.(2018)Luo, Xu, Li, Tian, Darrell, and Ma]{slbo}
Luo, Y., Xu, H., Li, Y., Tian, Y., Darrell, T., and Ma, T.
\newblock Algorithmic framework for model-based deep reinforcement learning
  with theoretical guarantees.
\newblock In \emph{International Conference on Learning Representations}, 2018.

\bibitem[Murata \& Kurokawa(2007)Murata and Kurokawa]{murata2007self}
Murata, S. and Kurokawa, H.
\newblock Self-reconfigurable robots.
\newblock \emph{IEEE Robotics \& Automation Magazine}, 2007.

\bibitem[Ng et~al.(2000)Ng, Russell, et~al.]{irl}
Ng, A.~Y., Russell, S.~J., et~al.
\newblock Algorithms for inverse reinforcement learning.
\newblock In \emph{Icml}, volume~1, pp.\ ~2, 2000.

\bibitem[Pan et~al.(2021)Pan, Garg, Anandkumar, and Zhu]{xinlei:evolution}
Pan, X., Garg, A., Anandkumar, A., and Zhu, Y.
\newblock Emergent hand morphology and control from optimizing robust grasps of
  diverse objects.
\newblock In \emph{2021 IEEE International Conference on Robotics and
  Automation (ICRA)}, pp.\  7540--7547. IEEE, 2021.

\bibitem[Pathak et~al.(2019)Pathak, Lu, Darrell, Isola, and
  Efros]{pathak2019learning}
Pathak, D., Lu, C., Darrell, T., Isola, P., and Efros, A.~A.
\newblock Learning to control self-assembling morphologies: a study of
  generalization via modularity.
\newblock \emph{NeurIPS}, 2019.

\bibitem[Radosavovic et~al.(2020)Radosavovic, Wang, Pinto, and Malik]{soil}
Radosavovic, I., Wang, X., Pinto, L., and Malik, J.
\newblock State-only imitation learning for dexterous manipulation.
\newblock \emph{arXiv preprint arXiv:2004.04650}, 2020.

\bibitem[Rajeswaran et~al.(2017)Rajeswaran, Lowrey, Todorov, and Kakade]{npg}
Rajeswaran, A., Lowrey, K., Todorov, E., and Kakade, S.
\newblock Towards generalization and simplicity in continuous control.
\newblock \emph{arXiv preprint arXiv:1703.02660}, 2017.

\bibitem[Rajeswaran et~al.(2018)Rajeswaran, Kumar, Gupta, Vezzani, Schulman,
  Todorov, and Levine]{dapg}
Rajeswaran, A., Kumar, V., Gupta, A., Vezzani, G., Schulman, J., Todorov, E.,
  and Levine, S.
\newblock Learning complex dexterous manipulation with deep reinforcement
  learning and demonstrations.
\newblock In \emph{Proceedings of Robotics: Science and Systems (RSS)},
  Pittsburgh, Pennsylvania, June 2018.
\newblock \doi{10.15607/RSS.2018.XIV.049}.

\bibitem[Romanishin et~al.(2013)Romanishin, Gilpin, and Rus]{romanishin2013m}
Romanishin, J.~W., Gilpin, K., and Rus, D.
\newblock M-blocks: Momentum-driven, magnetic modular robots.
\newblock In \emph{IROS}, 2013.

\bibitem[Ross et~al.(2011)Ross, Gordon, and Bagnell]{bc}
Ross, S., Gordon, G., and Bagnell, D.
\newblock A reduction of imitation learning and structured prediction to
  no-regret online learning.
\newblock In \emph{Proceedings of the fourteenth international conference on
  artificial intelligence and statistics}, pp.\  627--635. JMLR Workshop and
  Conference Proceedings, 2011.

\bibitem[Sadeghi \& Levine(2016)Sadeghi and Levine]{cad2rl}
Sadeghi, F. and Levine, S.
\newblock Cad2rl: Real single-image flight without a single real image.
\newblock \emph{arXiv preprint arXiv:1611.04201}, 2016.

\bibitem[Scarselli et~al.(2009)Scarselli, Gori, Tsoi, Hagenbuchner, and
  Monfardini]{scarselli2009graph}
Scarselli, F., Gori, M., Tsoi, A.~C., Hagenbuchner, M., and Monfardini, G.
\newblock The graph neural network model.
\newblock \emph{IEEE Transactions on Neural Network}, 2009.

\bibitem[Schaff et~al.(2018)Schaff, Yunis, Chakrabarti, and
  Walter]{schaff2018jointly}
Schaff, C., Yunis, D., Chakrabarti, A., and Walter, M.~R.
\newblock Jointly learning to construct and control agents using deep
  reinforcement learning.
\newblock \emph{arXiv preprint arXiv:1801.01432}, 2018.

\bibitem[Sims(1994{\natexlab{a}})]{sims1994evolving1}
Sims, K.
\newblock Evolving virtual creatures.
\newblock In \emph{Computer graphics and interactive techniques},
  1994{\natexlab{a}}.

\bibitem[Sims(1994{\natexlab{b}})]{sims1994evolving2}
Sims, K.
\newblock Evolving 3d morphology and behavior by competition.
\newblock \emph{Artificial life}, 1994{\natexlab{b}}.

\bibitem[Stoy et~al.(2010)Stoy, Brandt, Christensen, and Brandt]{stoy2010self}
Stoy, K., Brandt, D., Christensen, D.~J., and Brandt, D.
\newblock \emph{Self-reconfigurable robots: an introduction}.
\newblock Mit Press Cambridge, 2010.

\bibitem[Tobin et~al.(2017)Tobin, Fong, Ray, Schneider, Zaremba, and
  Abbeel]{domain:randomization}
Tobin, J., Fong, R., Ray, A., Schneider, J., Zaremba, W., and Abbeel, P.
\newblock Domain randomization for transferring deep neural networks from
  simulation to the real world.
\newblock In \emph{2017 IEEE/RSJ international conference on intelligent robots
  and systems (IROS)}, pp.\  23--30. IEEE, 2017.

\bibitem[Von~Neumann et~al.(1966)Von~Neumann, Burks, et~al.]{von1966theory}
Von~Neumann, J., Burks, A.~W., et~al.
\newblock Theory of self-reproducing automata.
\newblock \emph{IEEE Transactions on Neural Networks}, 1966.

\bibitem[Wang et~al.(2019{\natexlab{a}})Wang, Lehman, Clune, and
  Stanley]{wang2019paired}
Wang, R., Lehman, J., Clune, J., and Stanley, K.~O.
\newblock Paired open-ended trailblazer (poet): Endlessly generating
  increasingly complex and diverse learning environments and their solutions.
\newblock \emph{arXiv preprint arXiv:1901.01753}, 2019{\natexlab{a}}.

\bibitem[Wang et~al.(2018)Wang, Liao, Ba, and Fidler]{wang2018nervenet}
Wang, T., Liao, R., Ba, J., and Fidler, S.
\newblock Nervenet: Learning structured policy with graph neural networks.
\newblock \emph{ICLR}, 2018.

\bibitem[Wang et~al.(2019{\natexlab{b}})Wang, Zhou, Fidler, and
  Ba]{wang2019neural}
Wang, T., Zhou, Y., Fidler, S., and Ba, J.
\newblock Neural graph evolution: Towards efficient automatic robot design.
\newblock \emph{arXiv preprint arXiv:1906.05370}, 2019{\natexlab{b}}.

\bibitem[Whitman et~al.(2020)Whitman, Bhirangi, Travers, and
  Choset]{whitman2020modular}
Whitman, J., Bhirangi, R., Travers, M., and Choset, H.
\newblock Modular robot design synthesis with deep reinforcement learning.
\newblock In \emph{Proceedings of the AAAI Conference on Artificial
  Intelligence}, volume~34, pp.\  10418--10425, 2020.

\bibitem[Whitman et~al.(2021)Whitman, Travers, and Choset]{whitman2021learning}
Whitman, J., Travers, M., and Choset, H.
\newblock Learning modular robot control policies.
\newblock \emph{arXiv preprint arXiv:2105.10049}, 2021.

\bibitem[Wright et~al.(2007)Wright, Johnson, Peck, McCord, Naaktgeboren,
  Gianfortoni, Gonzalez-Rivero, Hatton, and Choset]{wright2007design}
Wright, C., Johnson, A., Peck, A., McCord, Z., Naaktgeboren, A., Gianfortoni,
  P., Gonzalez-Rivero, M., Hatton, R., and Choset, H.
\newblock Design of a modular snake robot.
\newblock In \emph{IROS}, 2007.

\bibitem[Yim et~al.(2000)Yim, Duff, and Roufas]{yim2000polybot}
Yim, M., Duff, D.~G., and Roufas, K.~D.
\newblock Polybot: a modular reconfigurable robot.
\newblock In \emph{ICRA}, 2000.

\bibitem[Zhao et~al.(2020)Zhao, Xu, Konakovi{\'c}-Lukovi{\'c}, Hughes,
  Spielberg, Rus, and Matusik]{zhao2020robogrammar}
Zhao, A., Xu, J., Konakovi{\'c}-Lukovi{\'c}, M., Hughes, J., Spielberg, A.,
  Rus, D., and Matusik, W.
\newblock Robogrammar: graph grammar for terrain-optimized robot design.
\newblock \emph{ACM Transactions on Graphics (TOG)}, 39\penalty0 (6):\penalty0
  1--16, 2020.

\bibitem[Zoph \& Le(2016)Zoph and Le]{zoph2016neural}
Zoph, B. and Le, Q.~V.
\newblock Neural architecture search with reinforcement learning.
\newblock \emph{arXiv preprint arXiv:1611.01578}, 2016.

\end{thebibliography}
\bibliographystyle{icml2022}

\newpage
\appendix

\onecolumn

\section{Proof of Theorem \ref{thm:revolver}} \label{sec:thm:proof}

Since $E(\cdot)$ is a continuous function of robot models, we assume the transition dynamics of the robot $E(\alpha)$ is differentiable and locally $L_1$-Lipschitz w.r.t. $\alpha$ in the sense that
\begin{equation}\label{eq:alpha:lipschitz:assumption}
\begin{split}
     \exists \varepsilon > 0,
    ||E(\alpha)(s,a) - E(\alpha^\prime)(s,a)|| \le L_1 |\alpha - \alpha^\prime|, 
    \forall s\in\mathcal{S}, \forall a\in\mathcal{A}, \forall |\alpha^\prime - \alpha | < \varepsilon    
\end{split}
\end{equation}
Moreover, we follow \cite{slbo} and assume the value functions of the robot models are $L_2$-Lipschitz w.r.t to some norm $||\cdot||$ in state space in the sense that
\begin{equation}\label{eq:state:lipschitz:assumption}
    |V^{\pi, M}(s) - V^{\pi, M}(s^\prime)| \le L_2 ||s-s^\prime||, \forall s, s^\prime \in \mathcal{S}
\end{equation}

By the assumption in Equation \eqref{eq:state:lipschitz:assumption} that the value functions of the robots are $L_2$-Lipschitz, as proven in Lemma 4.1 of \cite{slbo}, $\forall \phi > 0$, $M=E(\alpha), M^\prime=E(\alpha+\phi)$, we have 
\begin{equation}\label{eq:proof2}
\begin{split}
\exists \varepsilon > 0, \text{s.t. }
\mathop{\mathbb{E}}_{s} [ |V^{\pi,M}(s) - V^{\pi,M^\prime}(s)| ] 
& \le \frac{\gamma}{1-\gamma} L_2 \mathop{\mathbb{E}}_{(s,a)\sim \pi} || M(s,a) - M^\prime(s,a) || \\
& = \frac{\gamma}{1-\gamma} L_2 \mathop{\mathbb{E}}_{(s,a)\sim \pi} || E(\alpha)(s,a) - E(\alpha^\prime)(s,a) || \\
& \le \frac{\gamma}{1-\gamma} L_2 L_1 |\alpha - \alpha^\prime|, \forall |\alpha^\prime - \alpha | < \varepsilon 
\end{split}
\end{equation}
Since the value functions are differentiable by assumption in Equation \eqref{eq:alpha:lipschitz:assumption}, suppose $D = | \frac{\partial{V^{\pi,E(\alpha)}}}{\alpha} | $ is the absolute value of the derivative of $V^{\pi,E(\alpha)}$ w.r.t. to $\alpha$. 
According to Equation \eqref{eq:proof2}, $D$ is bounded.
We make a (strong) assumption that for all $\varphi\in [\alpha, \alpha+\xi]$, policy $\pi^*_{E(\varphi)}$ only achieves the best expected reward on robot $E(\varphi)$, so that
\begin{equation}
    \forall \varphi, \beta \in [\alpha, \alpha+\xi], 
    V^{\pi^*_{E(\varphi)},E(\varphi)} -
    V^{\pi^*_{E(\varphi)},E(\beta)} = D \cdot |\varphi - \beta| + o(|\beta-\varphi|^2)
    = D \cdot |\varphi - \beta| + o(\xi^2)
\end{equation}
From the definition, the value function of a policy $\pi$ on uniformly sampled robots $E(\beta)$ from $\beta\sim U(\alpha, \alpha+\xi)$ is
\begin{equation}
\begin{split}\label{eq:proof1}
    & \mathop{\mathbb{E}}_{ 
    \substack{
    \beta \sim U(\alpha, \alpha+\xi)
    }}
    \mathop{\mathbb{E}}_{ 
    \substack{(s_t, a_t) \sim \pi, E(\beta)
    }}
    \sum_{t} \gamma^t r_t \cdot\exp(h\beta) \\
    = & \mathop{\mathbb{E}}_{ 
    \substack{
    \beta \sim U(\alpha, \alpha+\xi)
    }}
    \exp(h\beta)
    \mathop{\mathbb{E}}_{ 
    \substack{(s_t, a_t) \sim \pi, E(\beta)
    }}
    \sum_{t} \gamma^t r_t  \\ 
    = & \mathop{\mathbb{E}}_{ 
    \substack{
    \beta \sim U(\alpha, \alpha+\xi)
    }}
    \exp(h\beta)
    V^{\pi,E(\beta)}  \\
    = & \frac{1}{\xi} \int_{\beta=\alpha}^{\alpha+\xi} \exp(h\beta)
    V^{\pi,E(\beta)} \dd{\beta}
\end{split}
\end{equation}
Supposed the $\pi$ that optimizes Equation \eqref{eq:proof1} is the policy that directly optimizes on one robot $E(\varphi), \varphi \in [\alpha, \alpha+\xi]$, i.e. $\pi^*_{E(\varphi)}$ optimizes Equation \eqref{eq:proof1}, then we treat $\pi^*_{E(\varphi)}$ and $V^{\pi^*_{E(\varphi)},E(\varphi)}$ as constants and the following variation should be zero
\begin{equation}
\begin{split}
    \delta \int_{\beta=\alpha}^{\alpha+\xi} \exp(h\beta)
    (V^{\pi^*_{E(\varphi)},E(\beta)} - V^{\pi^*_{E(\varphi)},E(\varphi)})  \dd{\beta} = 0
\end{split}
\end{equation}
which means
\begin{equation}
\begin{split}
    0 = & \int_{\beta=\alpha}^{\alpha+\xi} \frac{\partial}{\partial \beta} [ \exp(h\beta)
    (V^{\pi^*_{E(\varphi)},E(\beta)} - V^{\pi^*_{E(\varphi)},E(\varphi)}) ]  \dd{\beta}  \\ 
    \approx & \int_{\beta=\alpha}^{\varphi} \frac{\partial}{\partial \beta} [\exp(h\beta) D |\varphi-\beta| ] \dd{\beta} - \int_{\beta=\varphi}^{\alpha+\xi} \frac{\partial}{\partial \beta} [\exp(h\beta) D |\varphi-\beta| ] \dd{\beta} + o(\xi^2) \\
    = & D \int_{\beta=\alpha}^{\varphi} \frac{\partial}{\partial \beta} [\exp(h\beta) (\varphi - \beta) ] \dd{\beta} - D \int_{\beta=\varphi}^{\alpha+\xi} \frac{\partial}{\partial \beta} [\exp(h\beta) (\beta - \varphi) ] \dd{\beta} + o(\xi^2) \\
    = & \frac{e^{h\alpha}}{h} D [(h\varphi-1)(2e^{h(\varphi-\alpha)} - 1 - e^{h\xi}) + \\ & h(\alpha+\xi)e^{h\xi} - e^{h\xi} - h\varphi e^{h(\varphi-\alpha)} + e^{h(\varphi-\alpha)} - h\varphi e^{h(\varphi-\alpha)} + e^{h(\varphi-\alpha)} + h\alpha - 1 ] + o(\xi^2) \\
    = & \frac{e^{h\alpha}}{2h} D [(\alpha h^{2} \xi^{2} + 2 \alpha h \xi + 4 \alpha  + 2 h \xi^{2}  + 2 \xi) - (h^{2} \xi^{2} + 2 h \xi +4)\varphi ] +  o(\xi^2) 
\end{split}
\end{equation}
The last equation assumes $0 \le h(\varphi - \alpha) \le h\xi \ll 1$ and used second-order Taylor approximation of $e^y = 1 + y + \frac{1}{2}y^2 + o(y^2), y\in \mathbb{R}$. 
Then we have
\begin{equation}
\begin{split}
    \varphi & = \alpha + \frac{1}{2} \xi + \frac{\xi^2 h (2-\xi h)}{8 + 4\xi h + 2\xi^2 h^2} + o(\xi^2) \\
    & = \alpha + \frac{1}{2}\xi + \frac{1}{4} h\xi^2 + o(\xi^2)
\end{split}    
\end{equation}
Therefore, if
\begin{equation}
\begin{split}
    & \lim_{\xi \rightarrow 0}
    \left[
    \mathop{\arg\max}_{\pi}
    \mathop{\mathbb{E}}_{ 
    \substack{
    \beta \sim U(\alpha, \alpha + \xi) \\
    M_\beta = E(\beta)
    }}
    \mathop{\mathbb{E}}_{ 
    \substack{a_t \sim \pi(\cdot \mid s_t) \\ 
    s_{t+1} \sim M_\beta(\cdot \mid s_t, a_t)}}
    \sum_{t} \gamma^t r_t \exp(h \beta)
    \right] \\
    &
    = \pi_{M_\varphi}^* = \mathop{\arg\max}_{\pi}
    \mathop{\mathbb{E}}_{ 
    \substack{a_t \sim \pi(\cdot \mid s_t) \\ 
    s_{t+1} \sim M_\varphi(\cdot \mid s_t, a_t)}}
    \sum_{t} \gamma^t r_t,
\end{split}
\end{equation}
then when $\xi\rightarrow 0$, $\varphi = \alpha + \frac{1}{2}\xi + \frac{1}{4} h\xi^2 + o(\xi^2)$.
\qed

\section{Experiment Details}

\paragraph{ Caching Intermediate Robots  }
In practice, frequently calling the function for generating simulation environments could be costly.
To avoid repetitively generating a new environment at the start of every epoch and to speed up the training process, we pre-generate and cache a large number (e.g. 1,000) of environments.
At the start of each epoch, we randomly sample within the desired interpolation range and fetch a simulation environment from the cache.
When the number of cached environments is large, the above sampling behavior is a good approximation to sampling from continuous robot evolution.

\paragraph{Experiment Hyperparameter Setting}
We illustrate the hyperparameters of the neural networks used in Gym and Hand Manipulation Suite experiments, including layer size, batch size and learning rate, in Table \ref{tab:supp:net:param}.

\begin{table}[H]
\centering
\small
\begin{tabular}{l|c|c}
\hline
& MuJoCo Gym Environments & Hand Manipulation Suite \\
\hline
Actor & $[s,256,256,a]$ & $[s,32,32,a]$ \\ 
\hline
Critic & $[s+a,256,256,1]$ & $[s+a,32,32,1]$ \\ 
\hline
Batch size & 256 & 16 \\ 
\hline
Learning Rate & $3\times 10^{-4}$ & $1\times 10^{-4}$ \\
\hline
\end{tabular}
\vspace{-1ex}
\caption{Size of the Neural network used in the experiments. 
$s$ and $a$ represents the dimension of state space and action space respectively.
}
\label{tab:supp:net:param}
\end{table}

\section{Qualitative Results}

\begin{figure}[ht]
\centering
\newcommand\suppwidth{0.15}
\begin{tabular}{cccccc}
\includegraphics[width=\suppwidth\textwidth,valign=m]{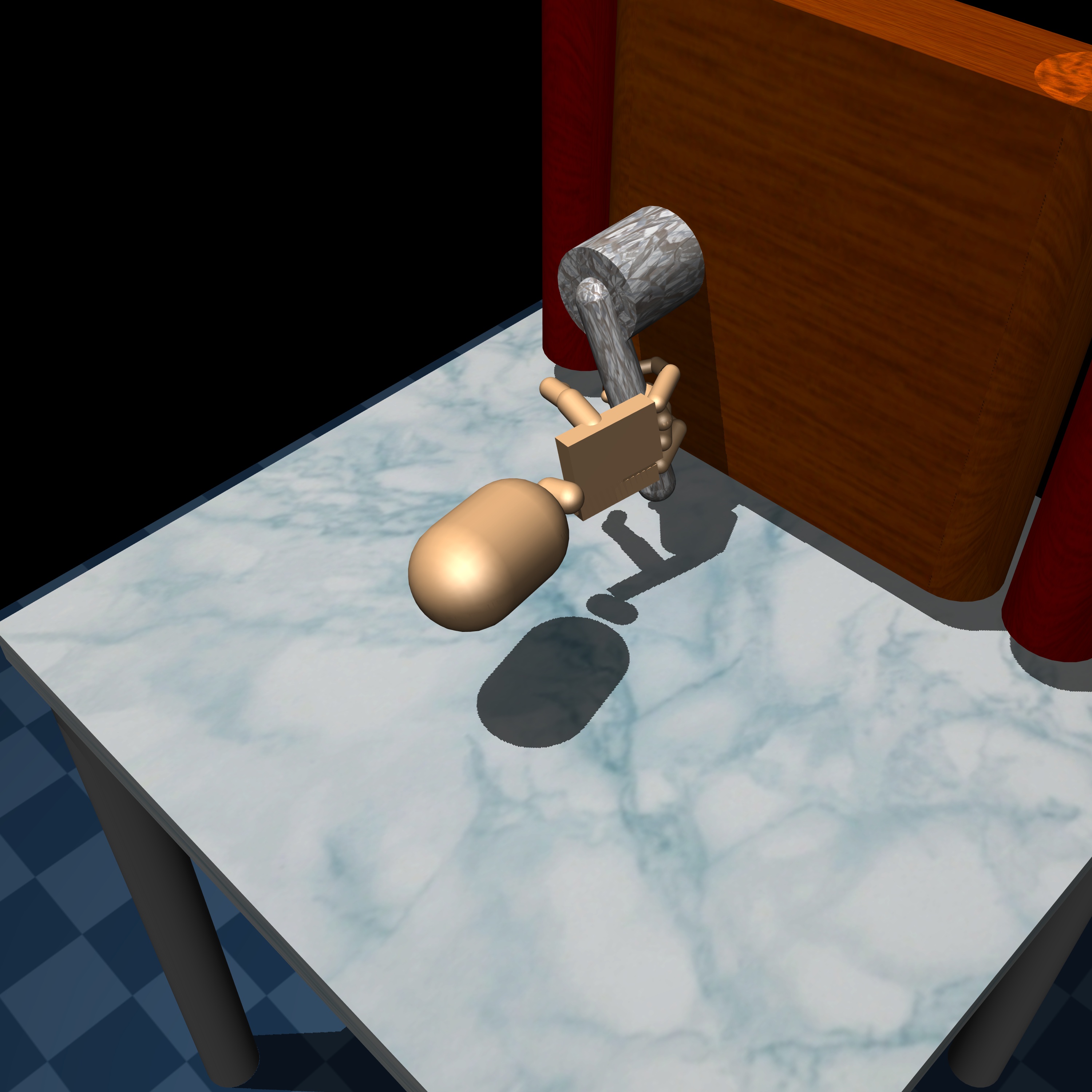}
\includegraphics[width=\suppwidth\textwidth,valign=m]{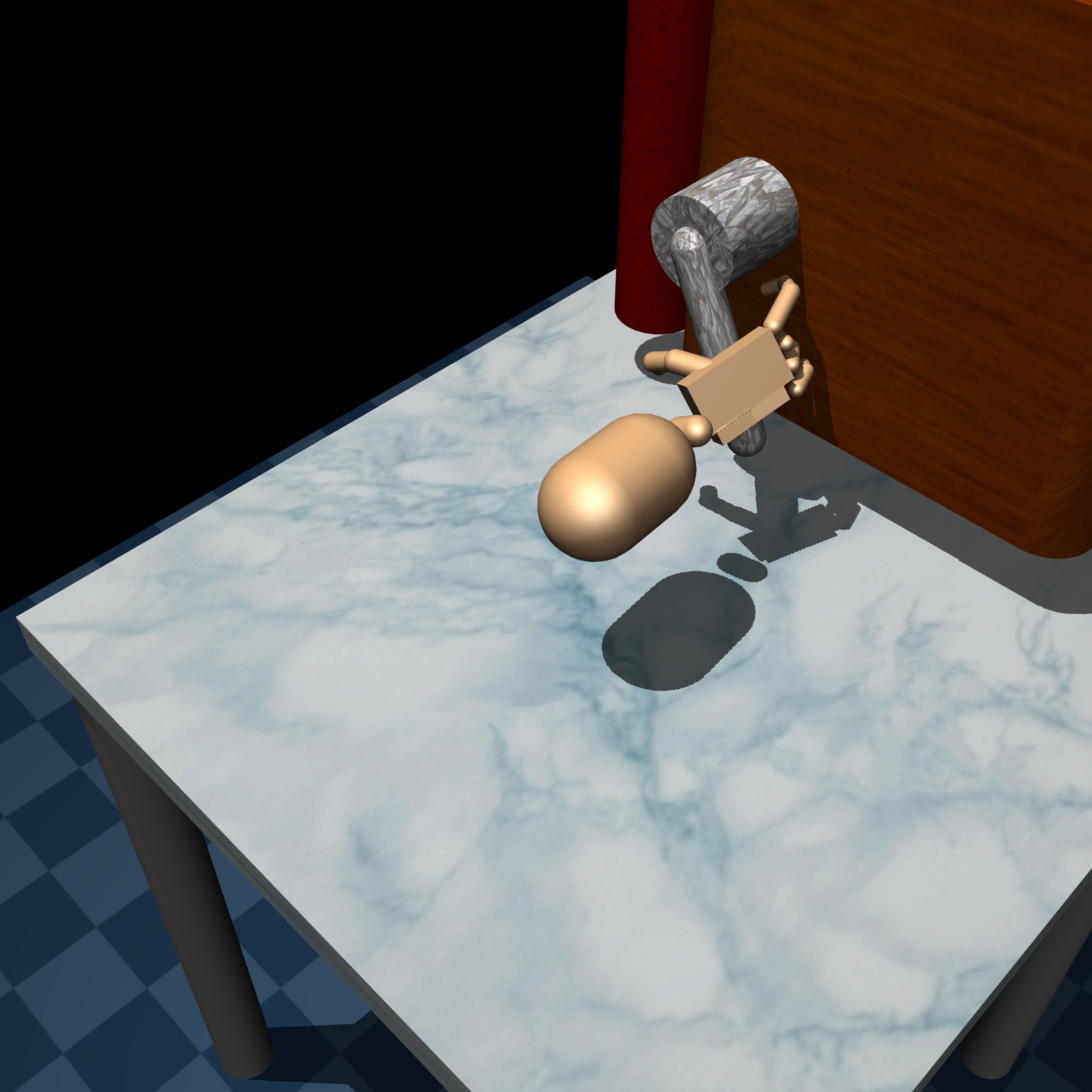}
\includegraphics[width=\suppwidth\textwidth,valign=m]{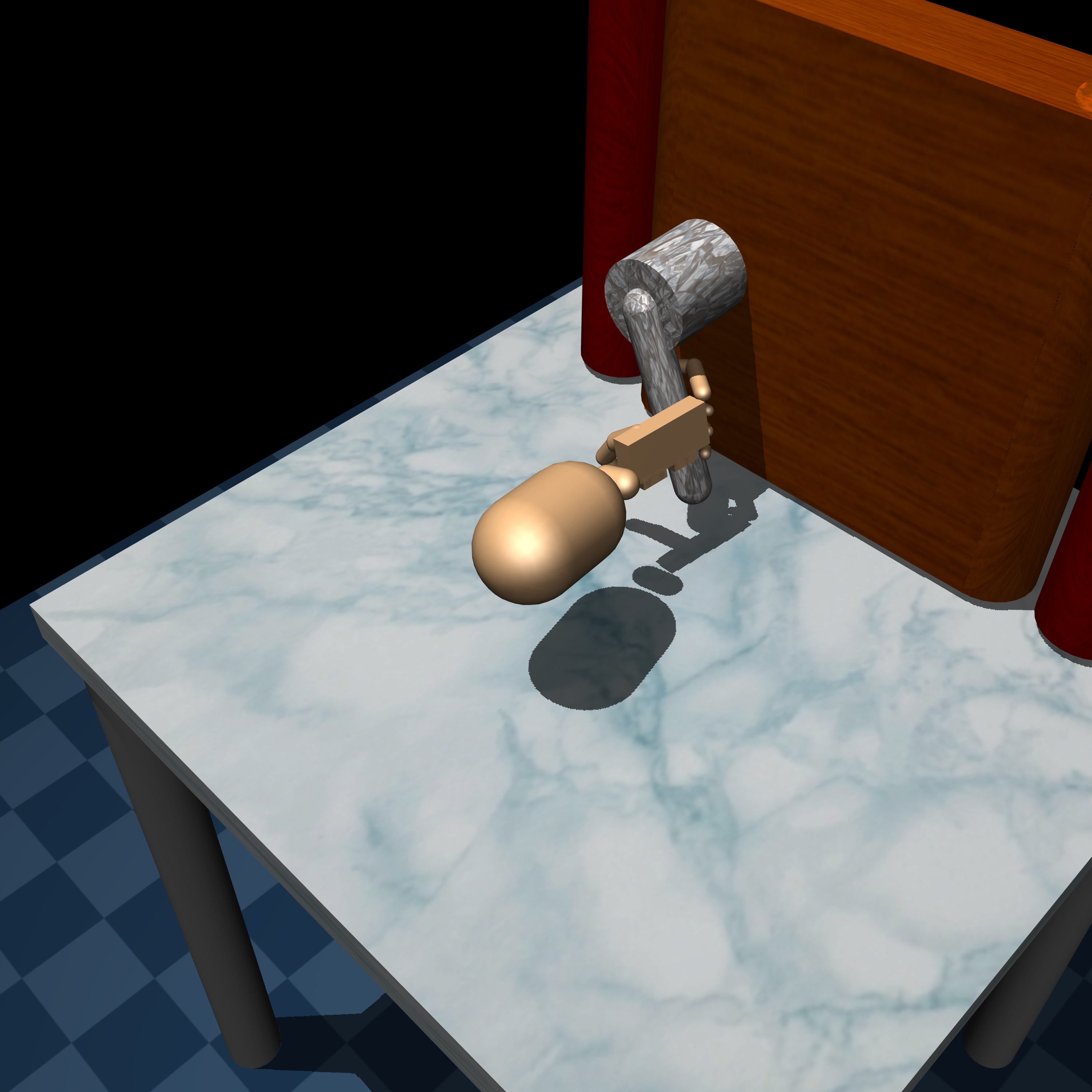}
\includegraphics[width=\suppwidth\textwidth,valign=m]{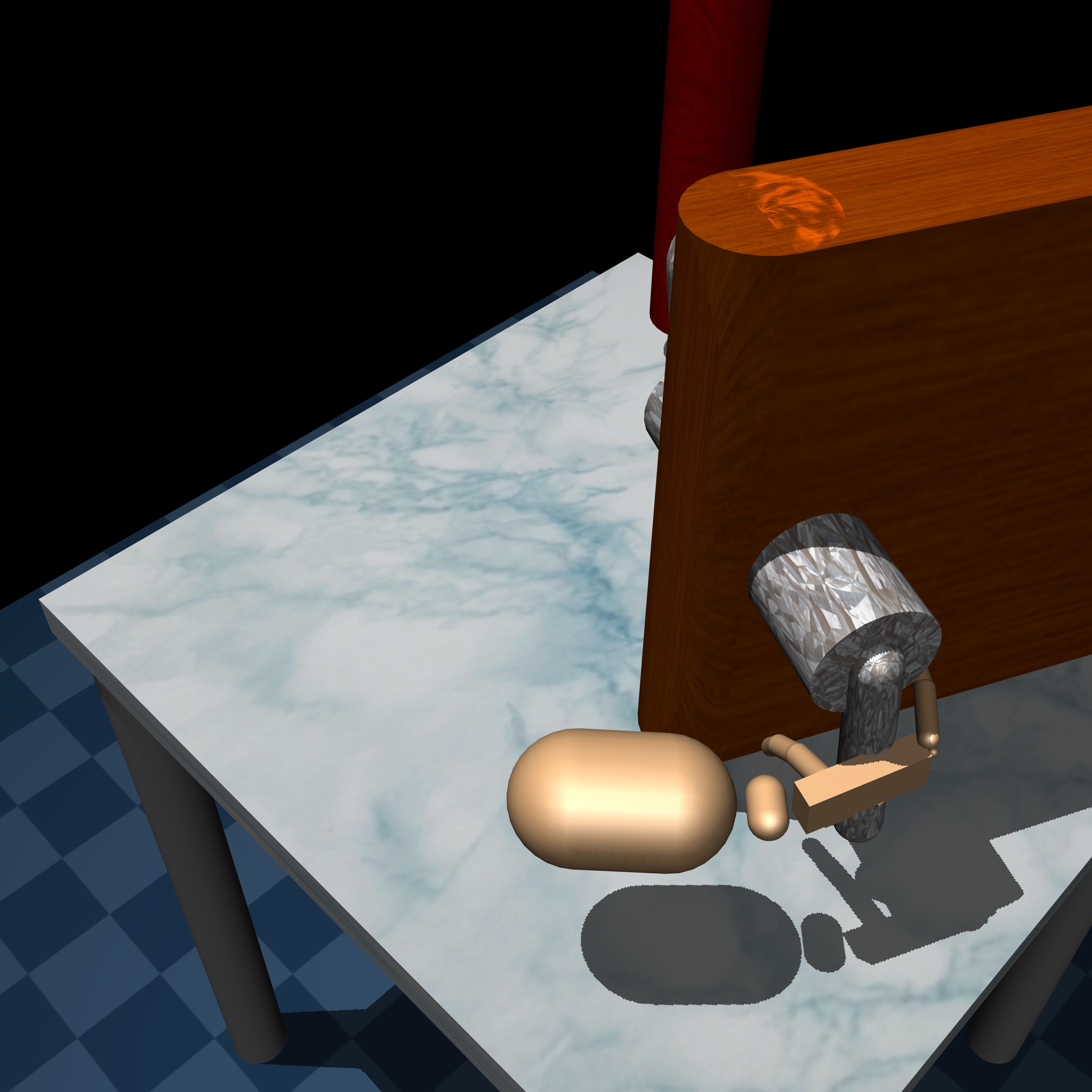}
\includegraphics[width=\suppwidth\textwidth,valign=m]{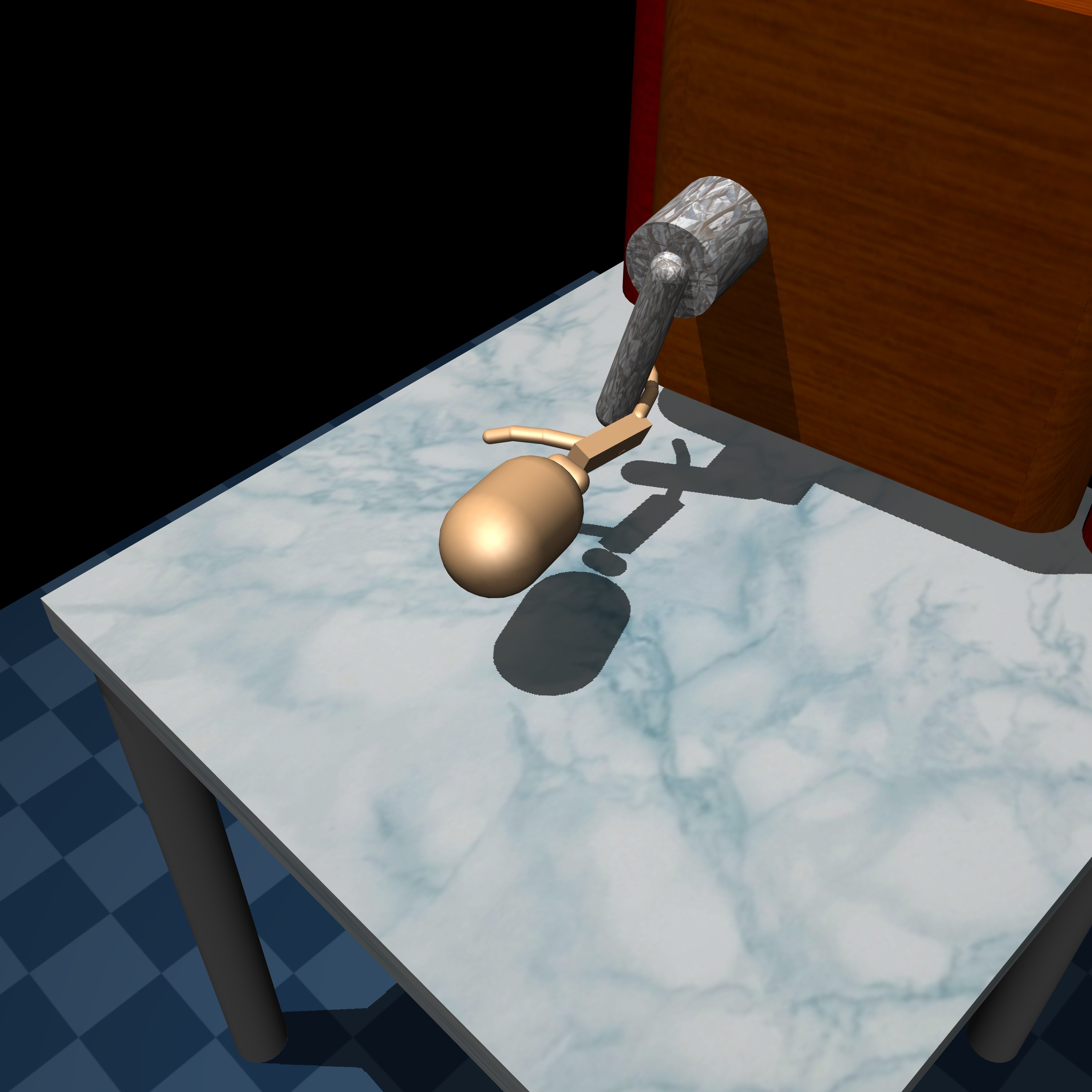}
\includegraphics[width=\suppwidth\textwidth,valign=m]{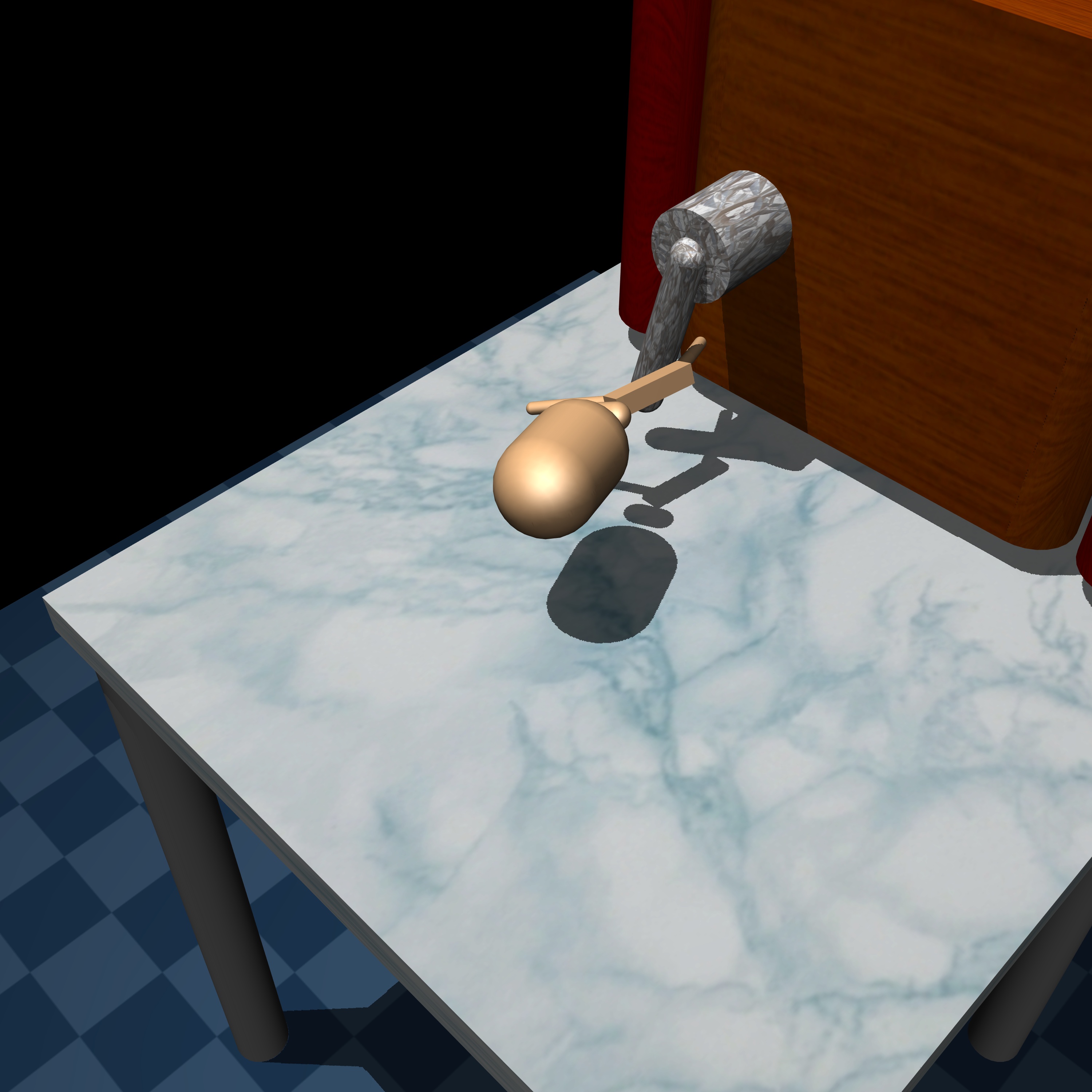} \\ \\
\includegraphics[width=\suppwidth\textwidth, valign=m]{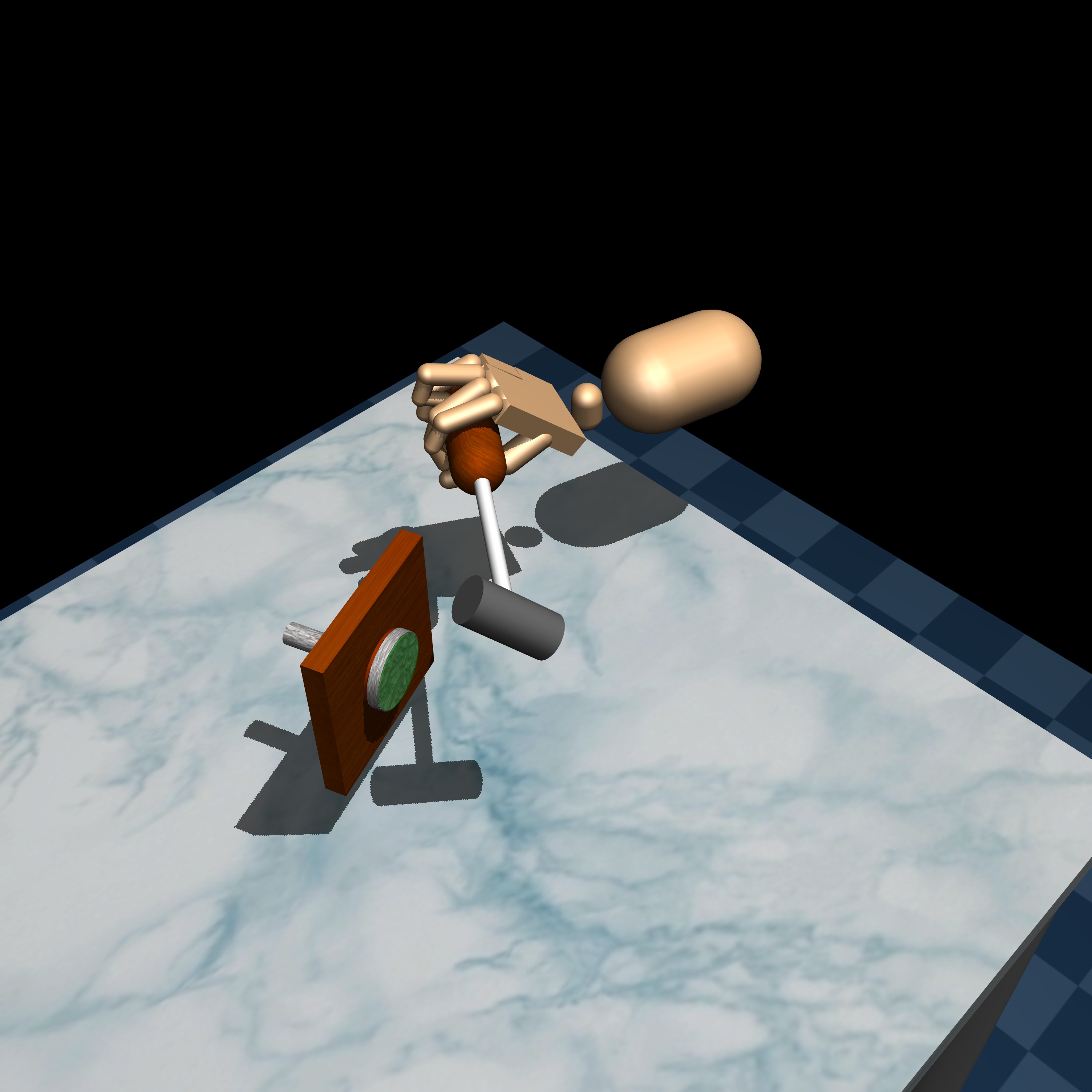}
\includegraphics[width=\suppwidth\textwidth,valign=m]{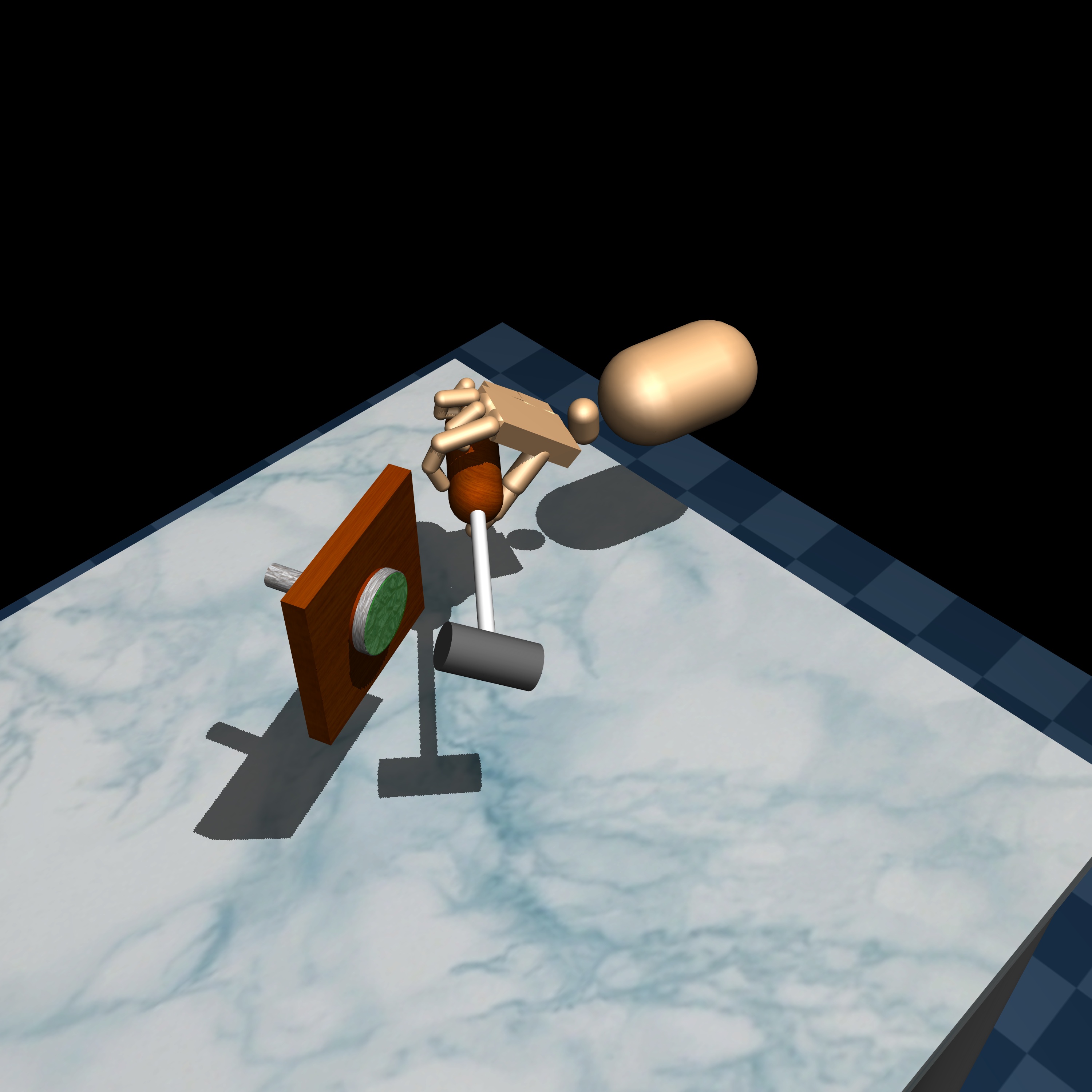}
\includegraphics[width=\suppwidth\textwidth,valign=m]{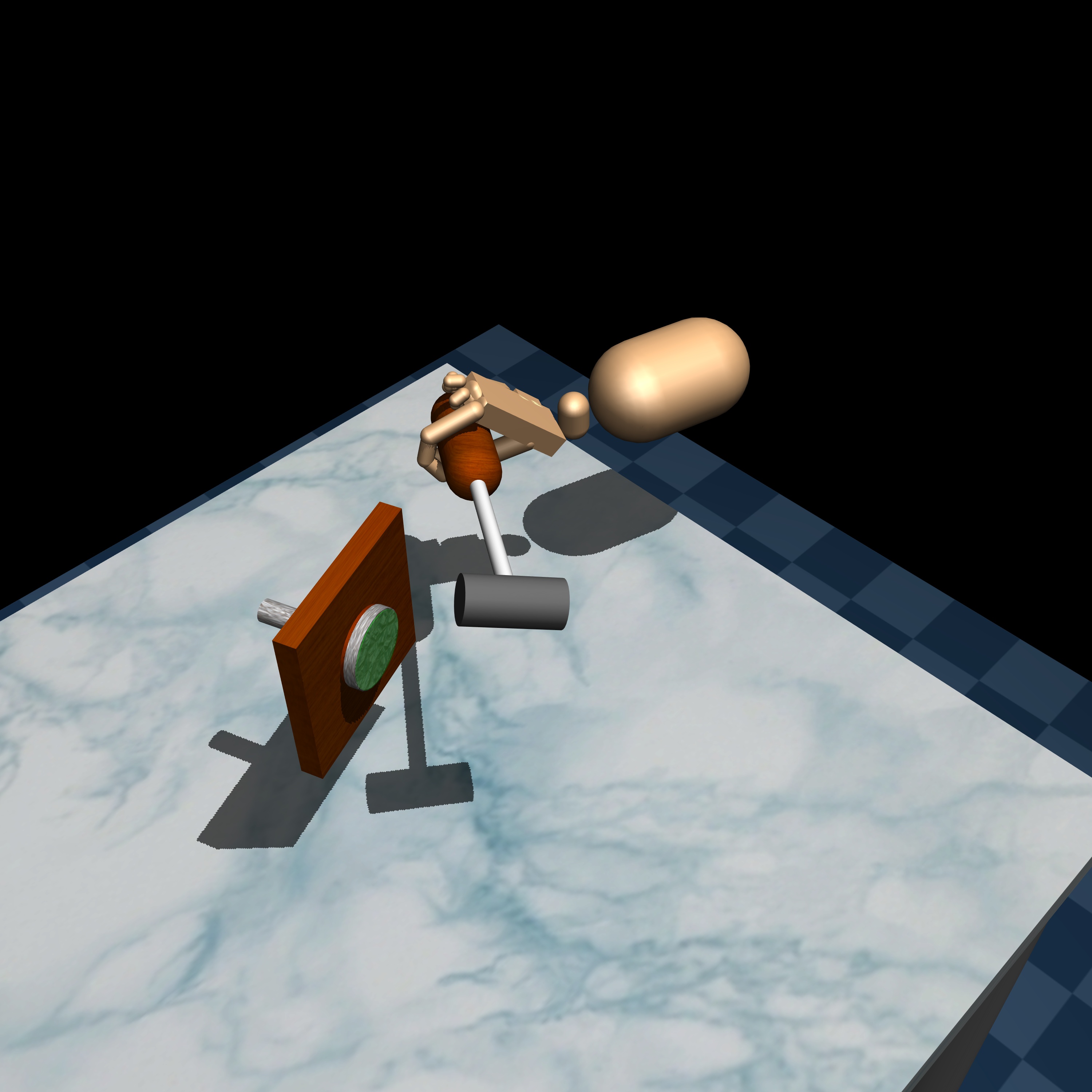}
\includegraphics[width=\suppwidth\textwidth, valign=m]{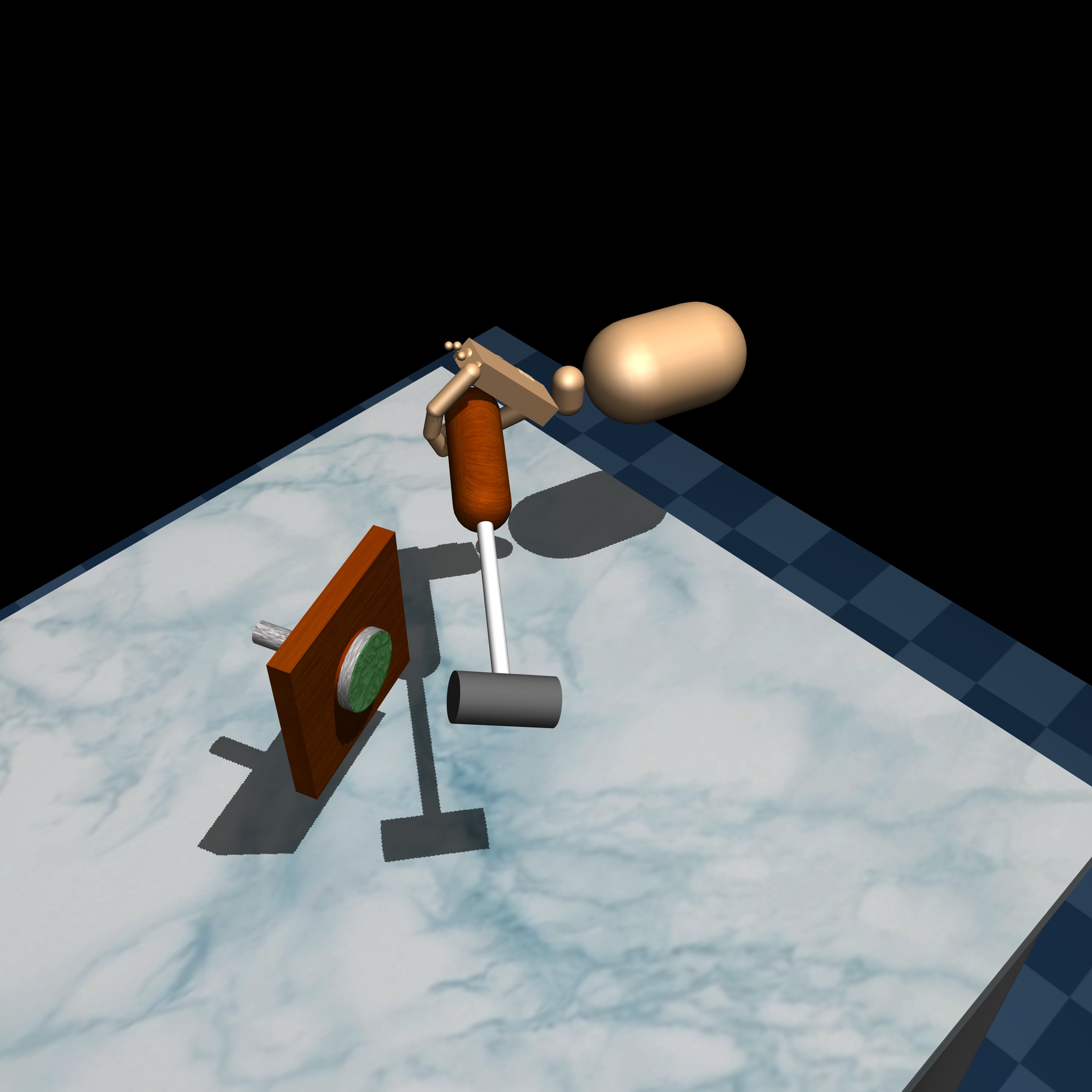}
\includegraphics[width=\suppwidth\textwidth,valign=m]{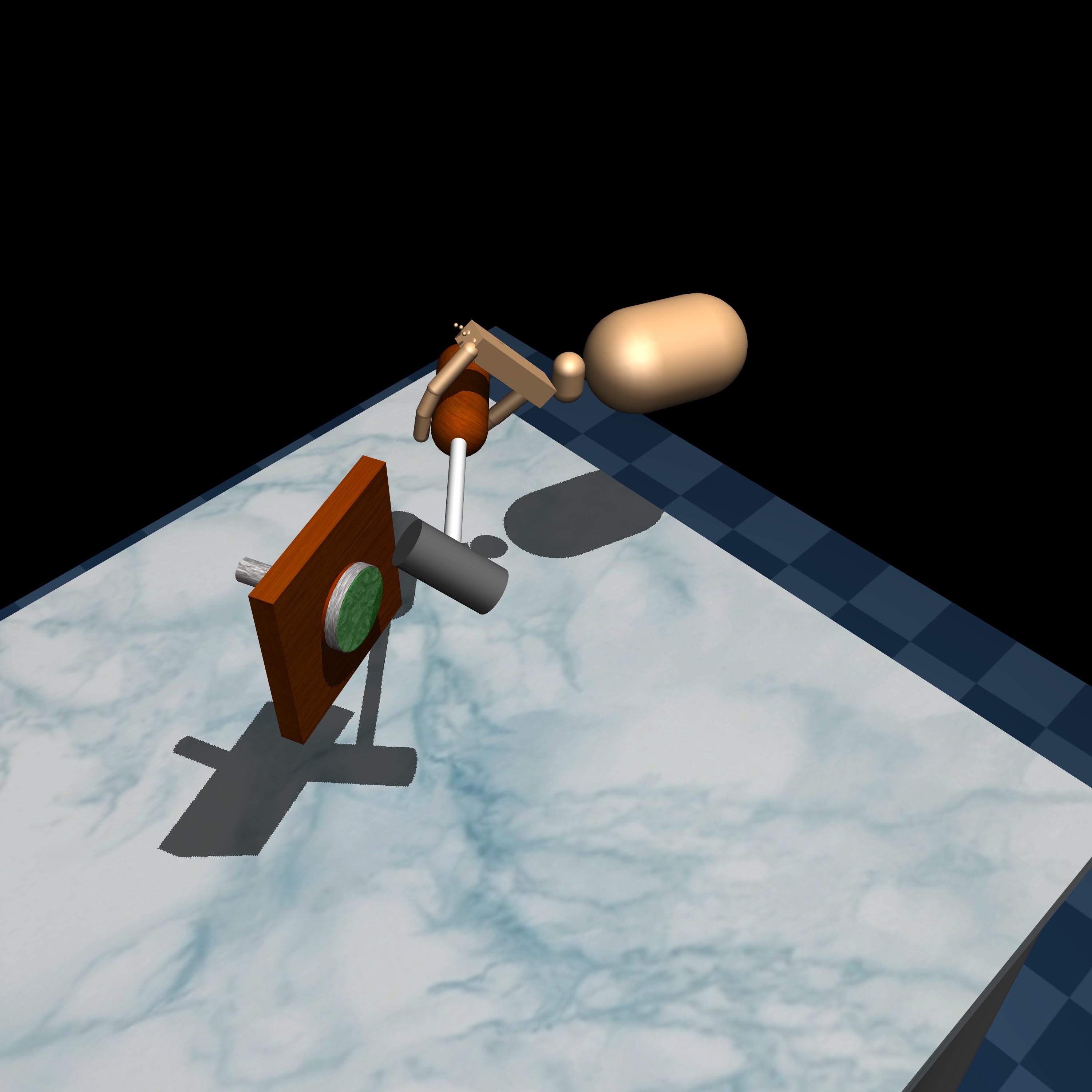}
\includegraphics[width=\suppwidth\textwidth,valign=m]{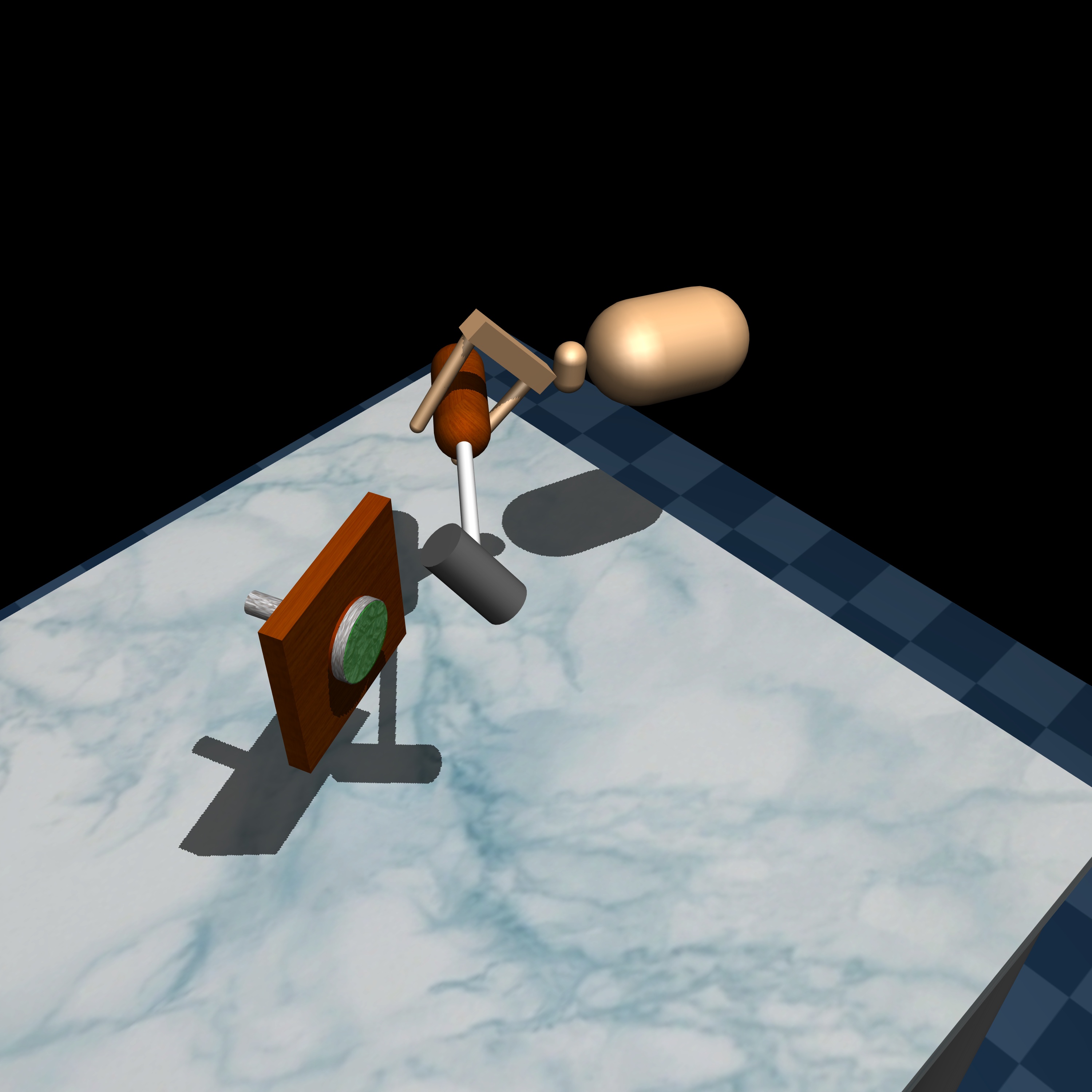} \\ \\
\includegraphics[width=\suppwidth\textwidth,valign=m]{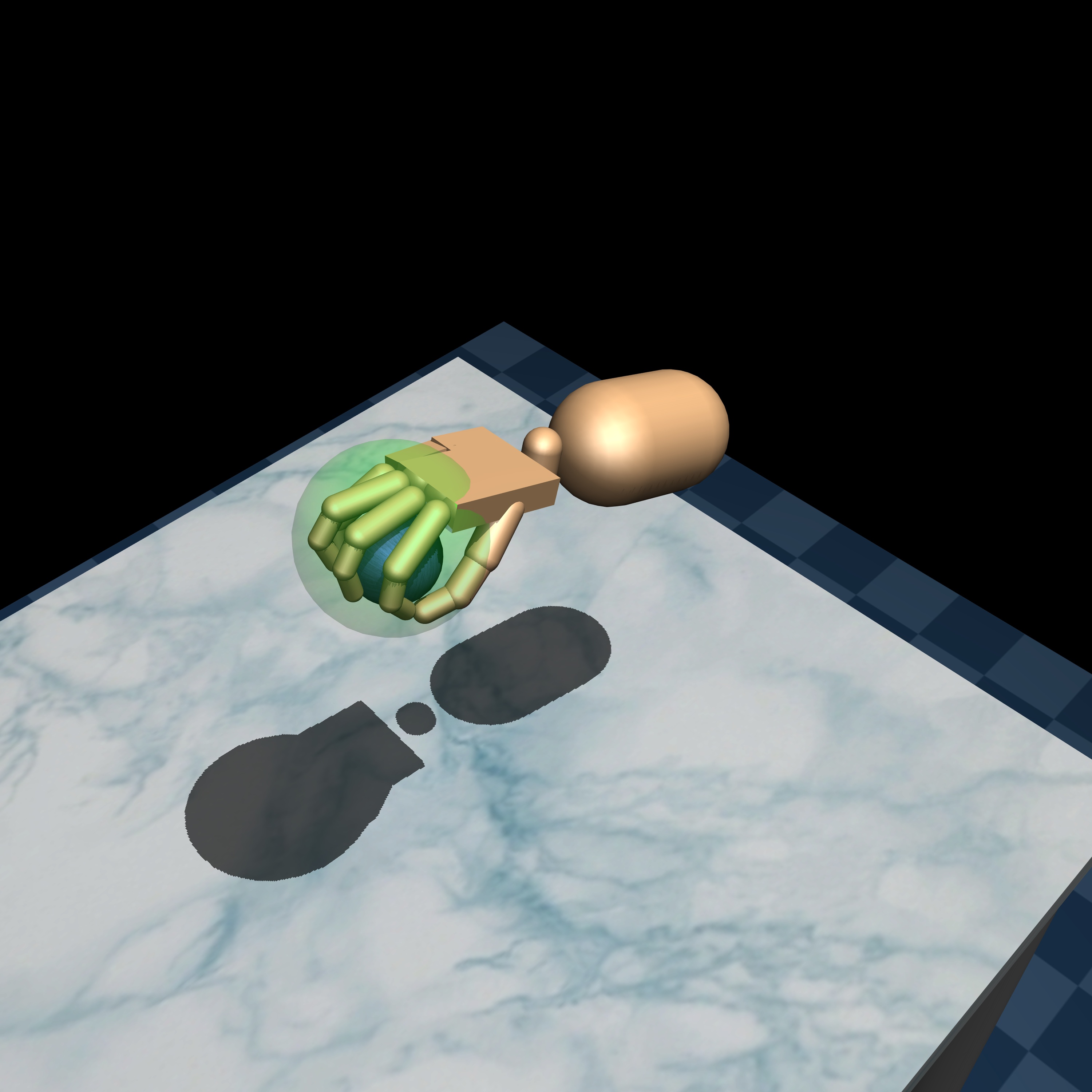}
\includegraphics[width=\suppwidth\textwidth,valign=m]{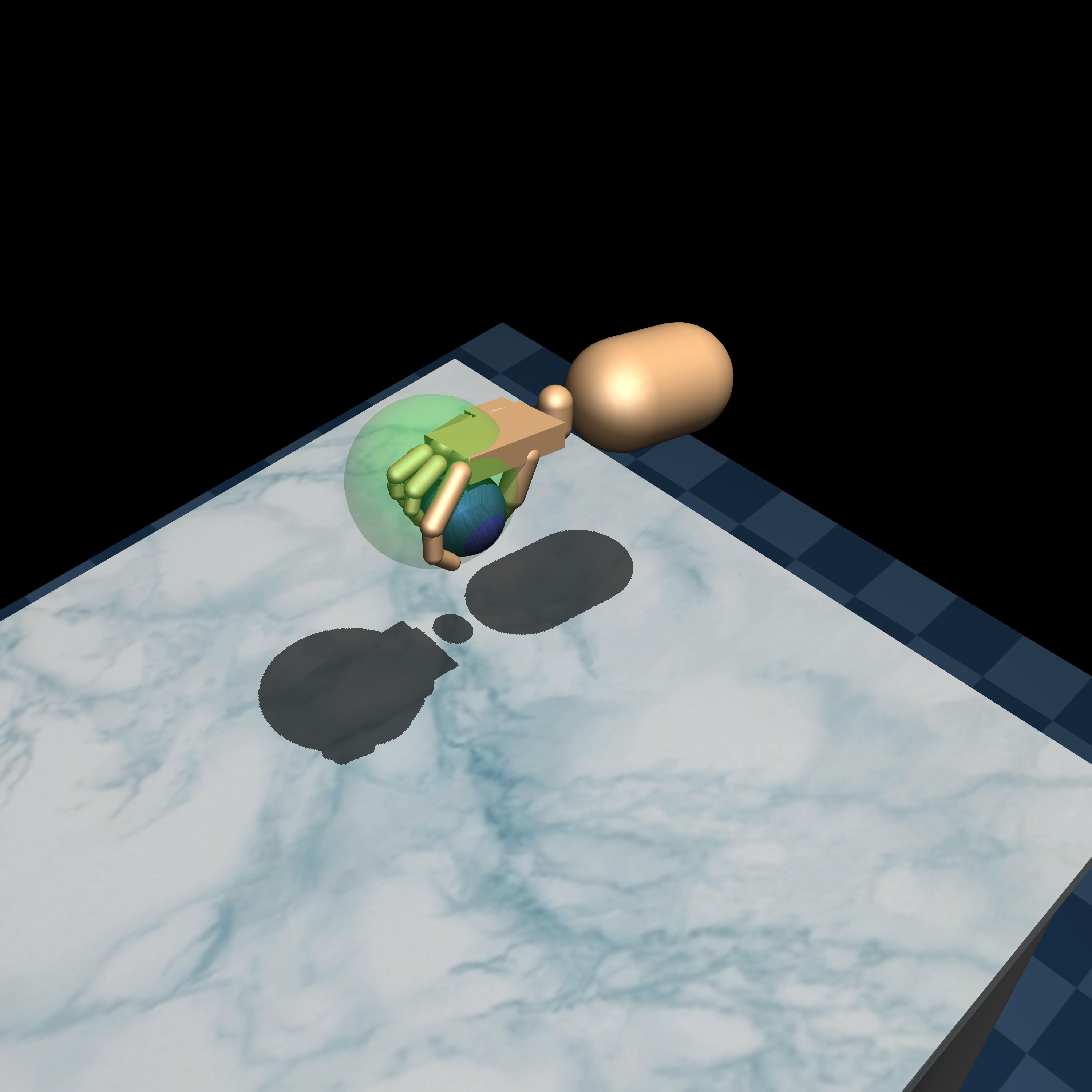}
\includegraphics[width=\suppwidth\textwidth,valign=m]{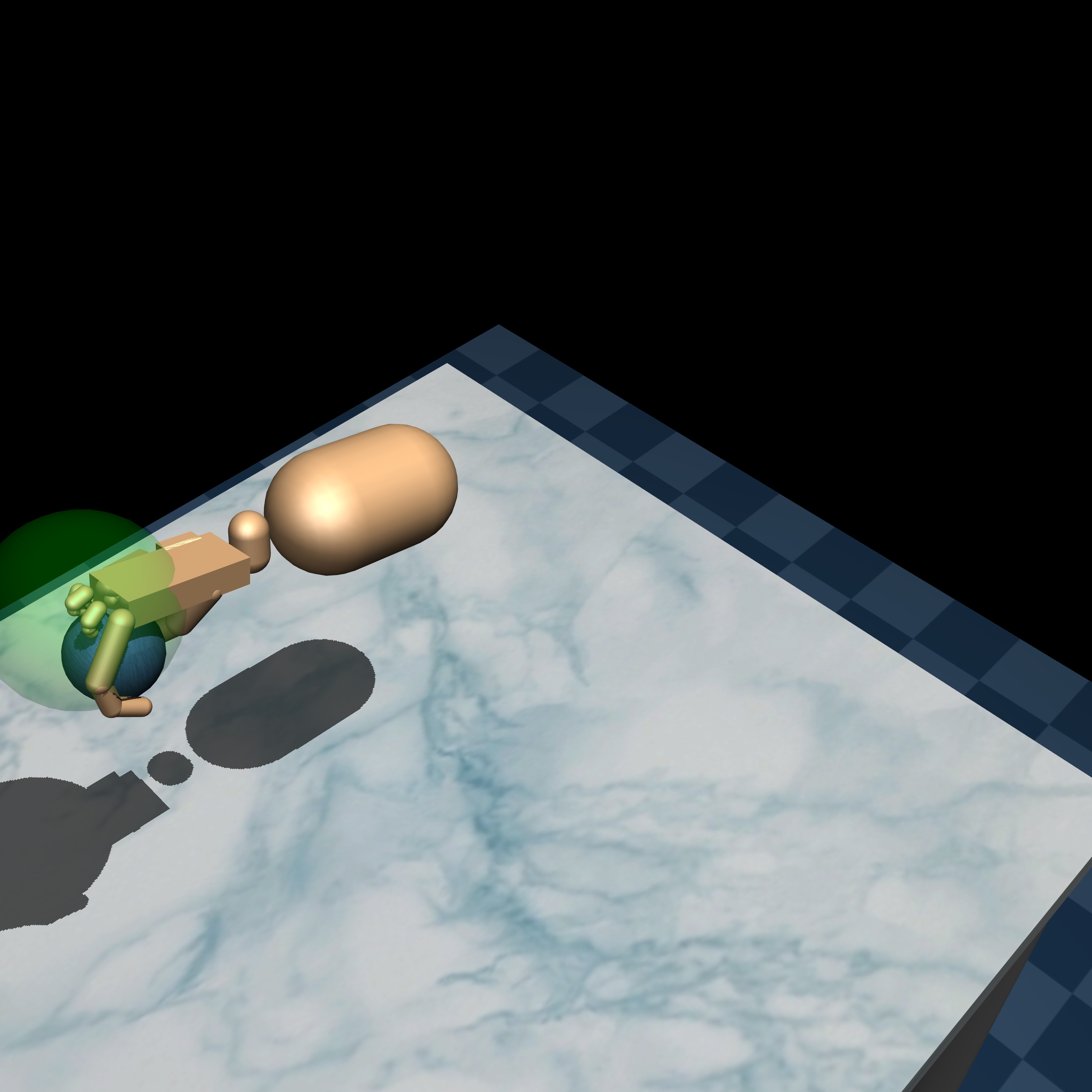}
\includegraphics[width=\suppwidth\textwidth,valign=m]{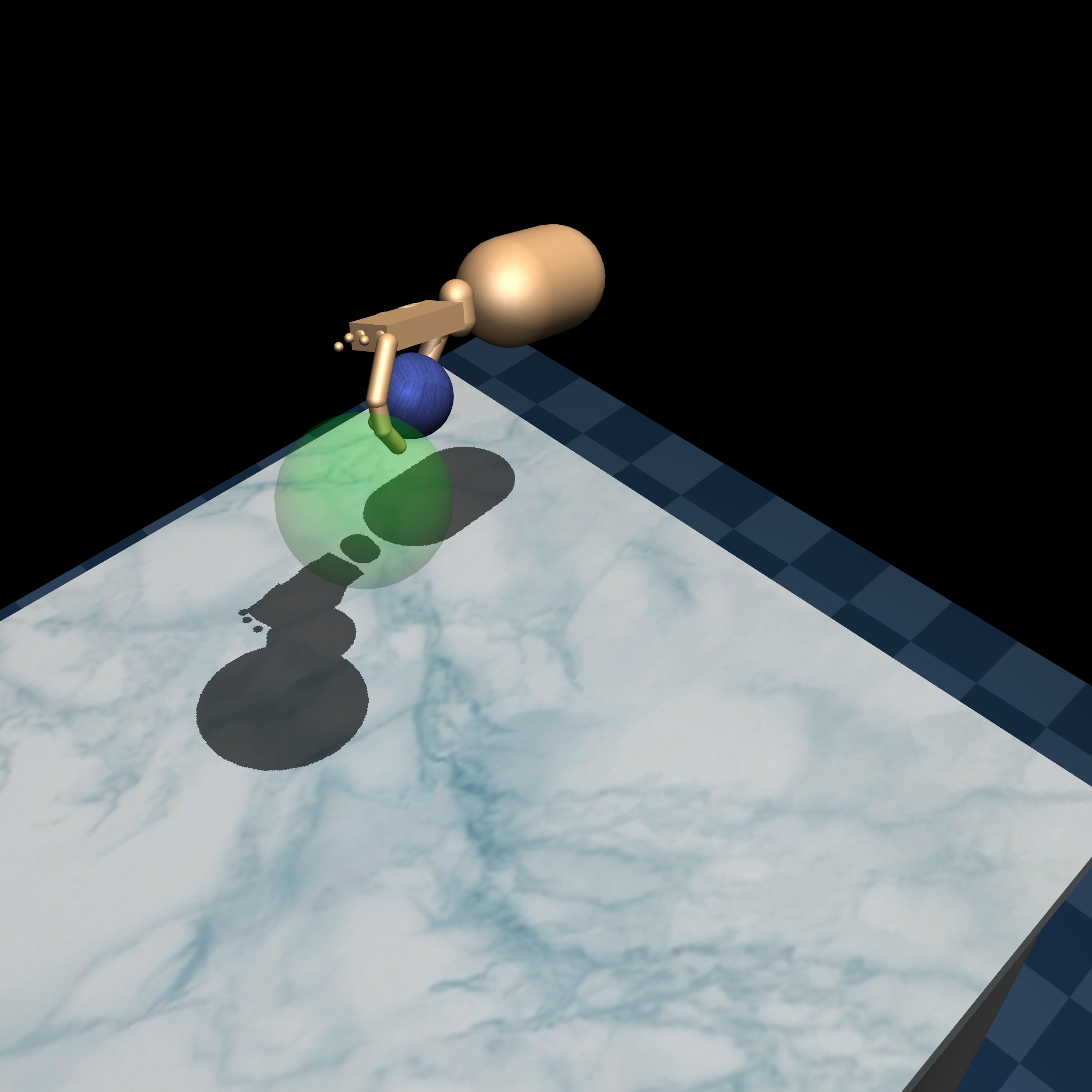}
\includegraphics[width=\suppwidth\textwidth,valign=m]{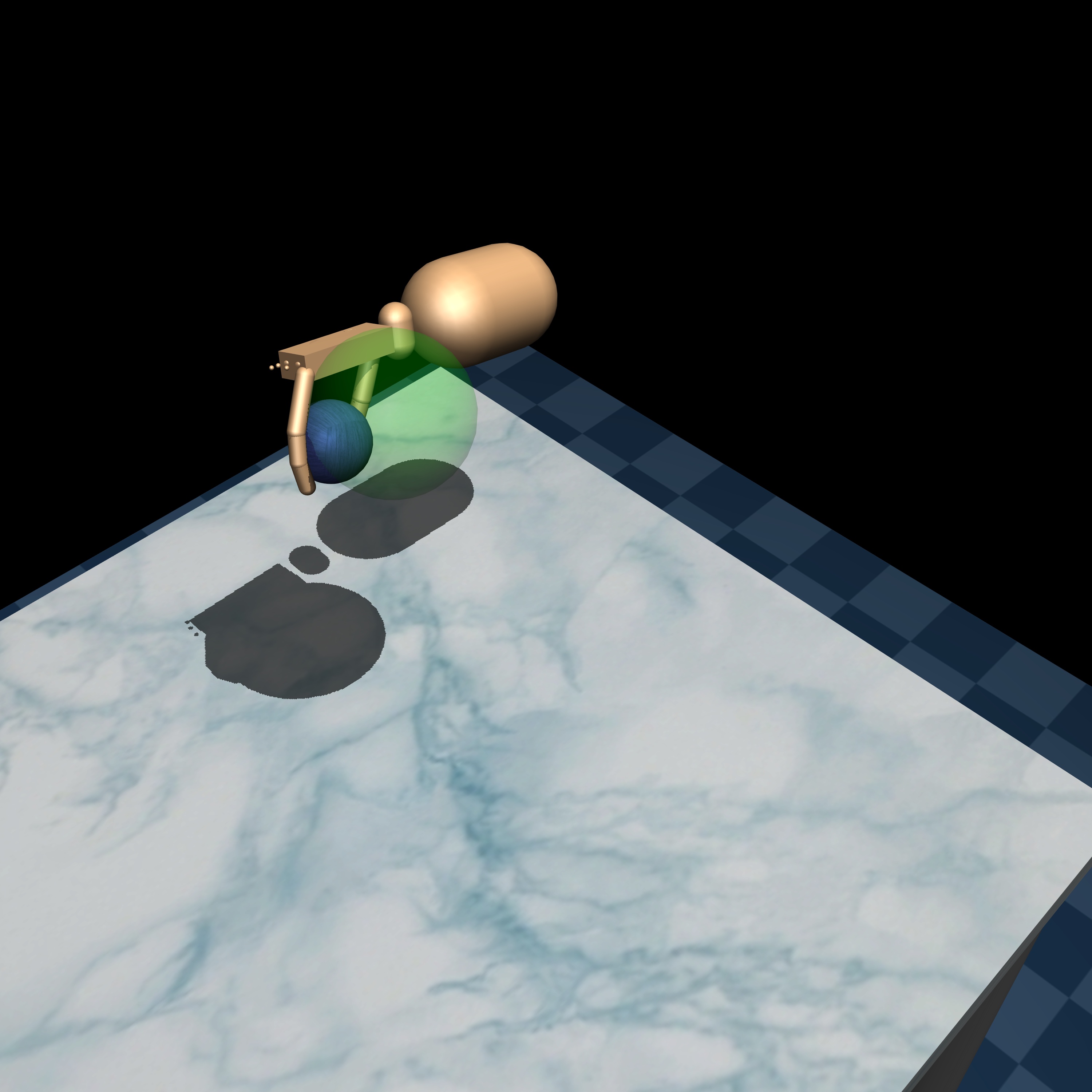}
\includegraphics[width=\suppwidth\textwidth,valign=m]{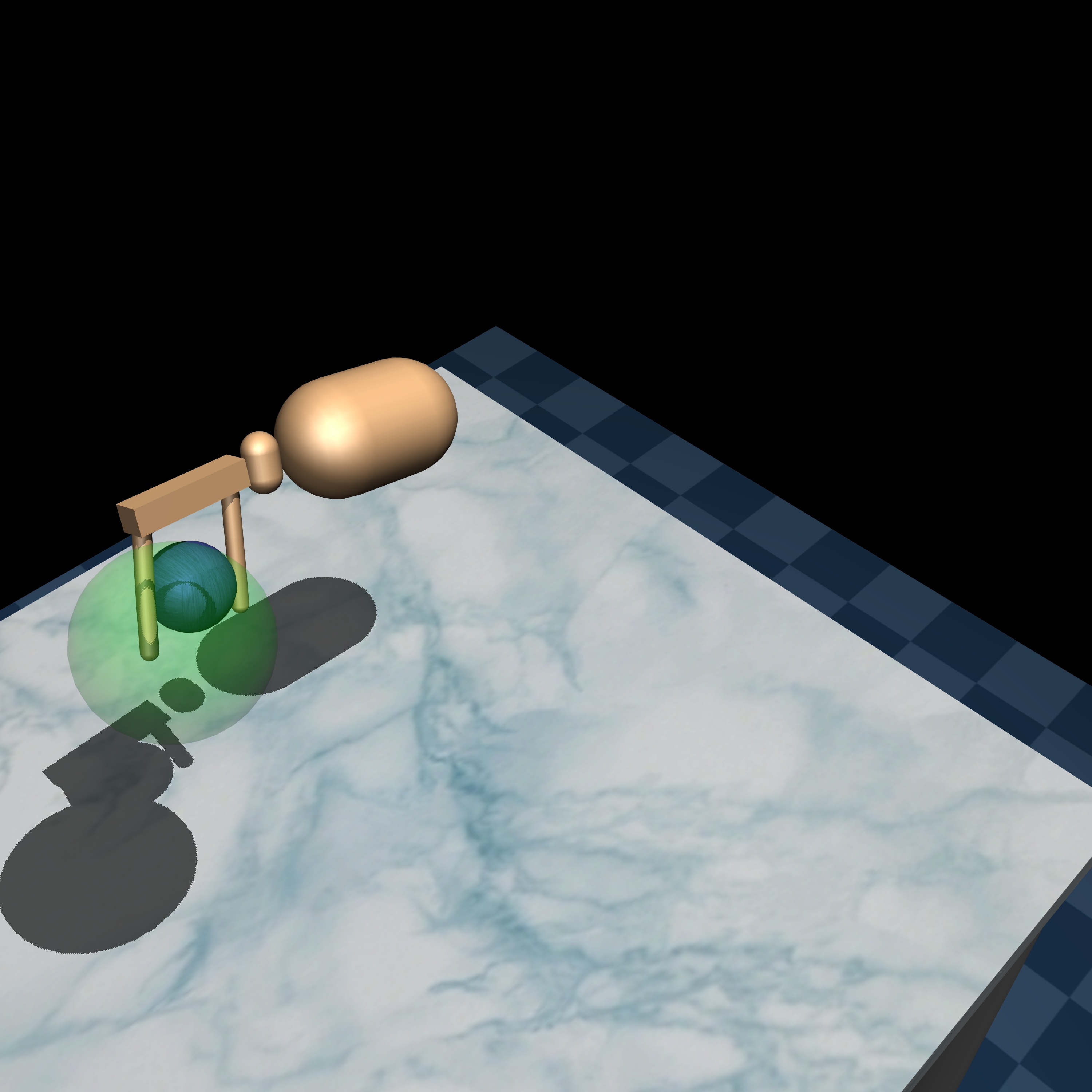}
\end{tabular}
\vspace{-1ex}
\caption{
\textbf{Visualization of policy on evolving intermediate robots on Hand Manipulation Suite tasks.}
The three rows shows  \texttt{Door}, \texttt{Hammer}, and \texttt{Relocate} tasks respectively.
From left to right in each row, we show a snapshot of robot at evolution parameters $\alpha$ at $0, 0.2, 0.4, 0.6, 0.8, 1$ respectively in the six columns.
} 
\label{fig:dapg:evolve:viz}
\end{figure}

\begin{figure}[ht]
\centering
\newcommand\suppwidth{0.15}
\begin{tabular}{cccccc}
\includegraphics[width=\suppwidth\textwidth,valign=m]{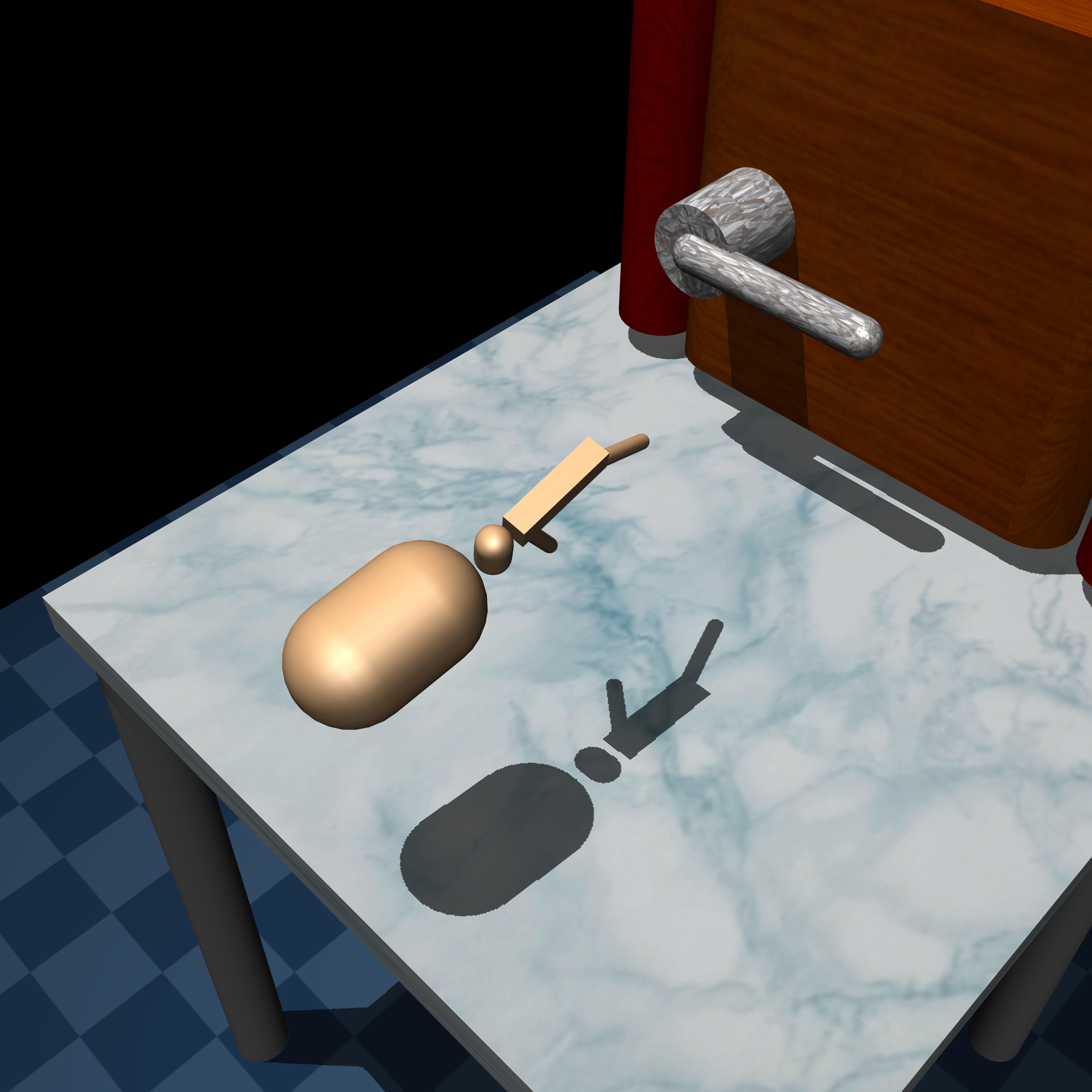}
\includegraphics[width=\suppwidth\textwidth,valign=m]{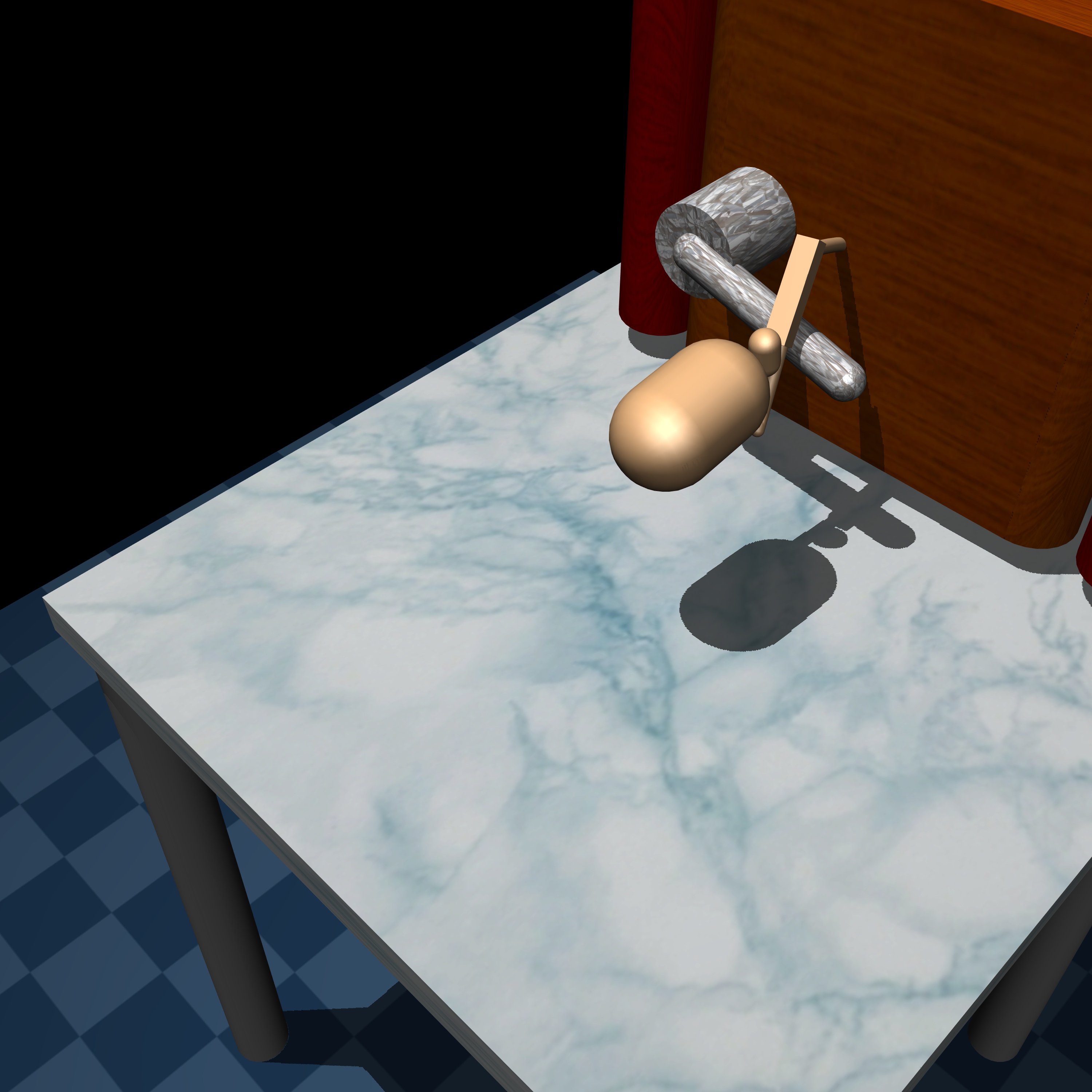}
\includegraphics[width=\suppwidth\textwidth,valign=m]{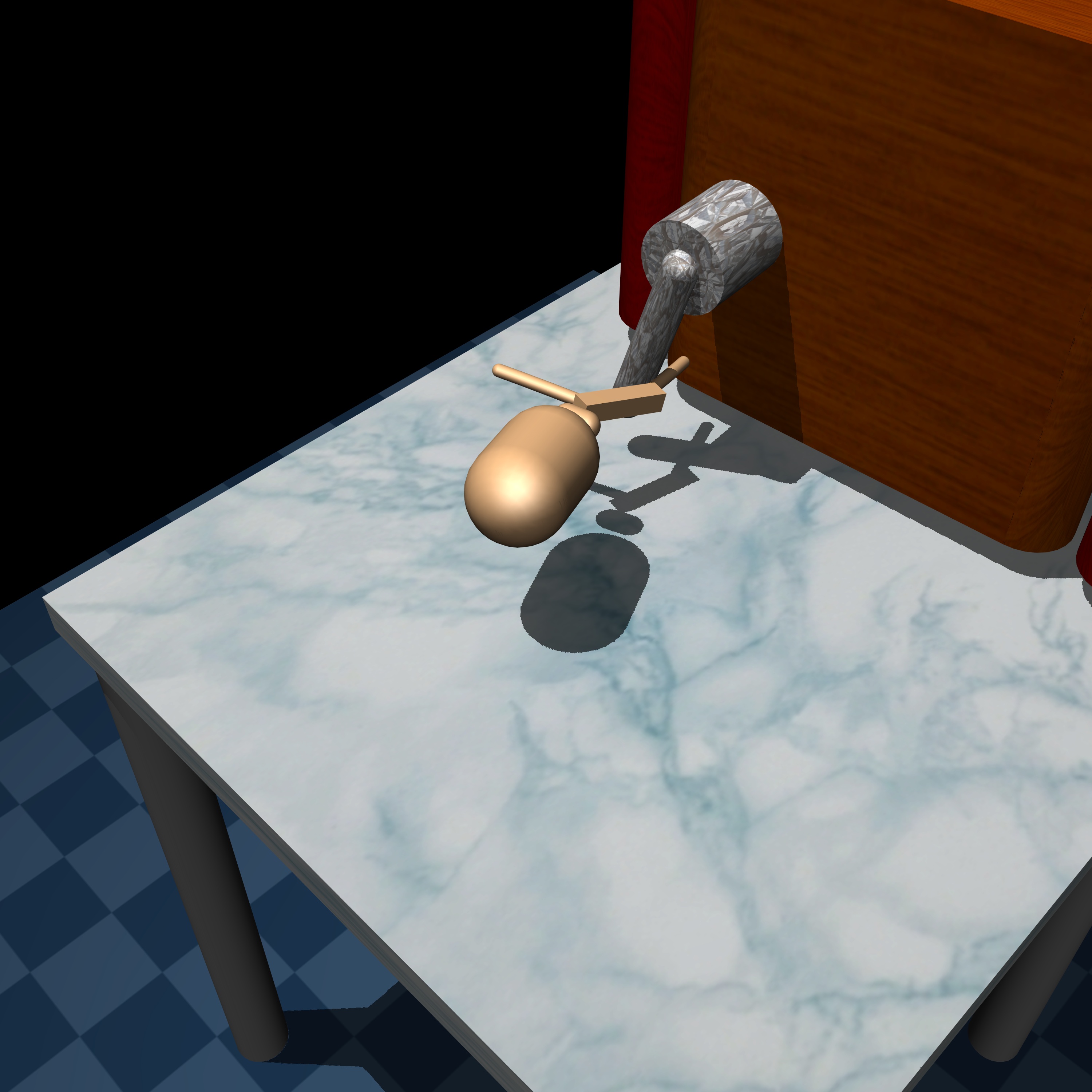}
\includegraphics[width=\suppwidth\textwidth,valign=m]{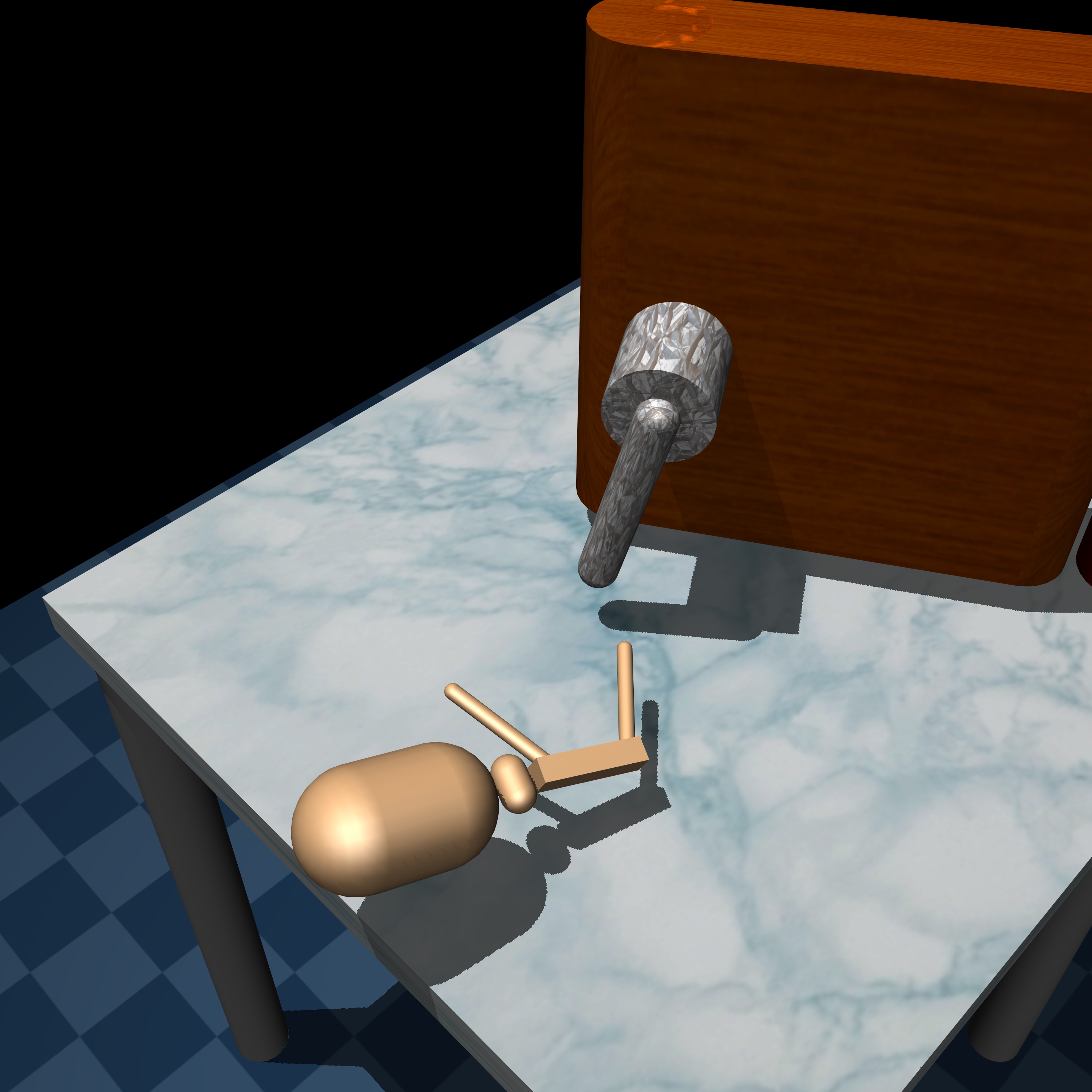}
\includegraphics[width=\suppwidth\textwidth,valign=m]{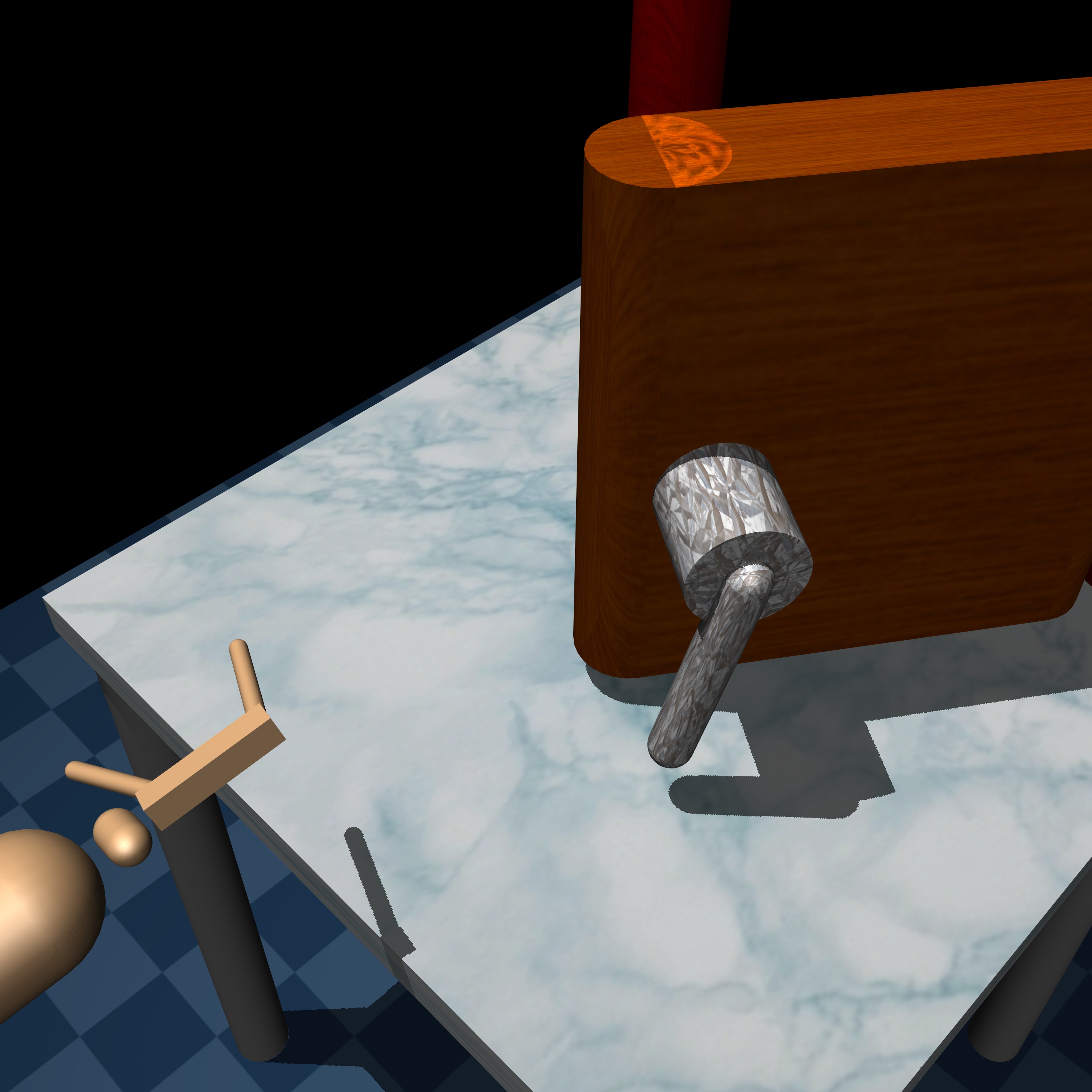}
\includegraphics[width=\suppwidth\textwidth,valign=m]{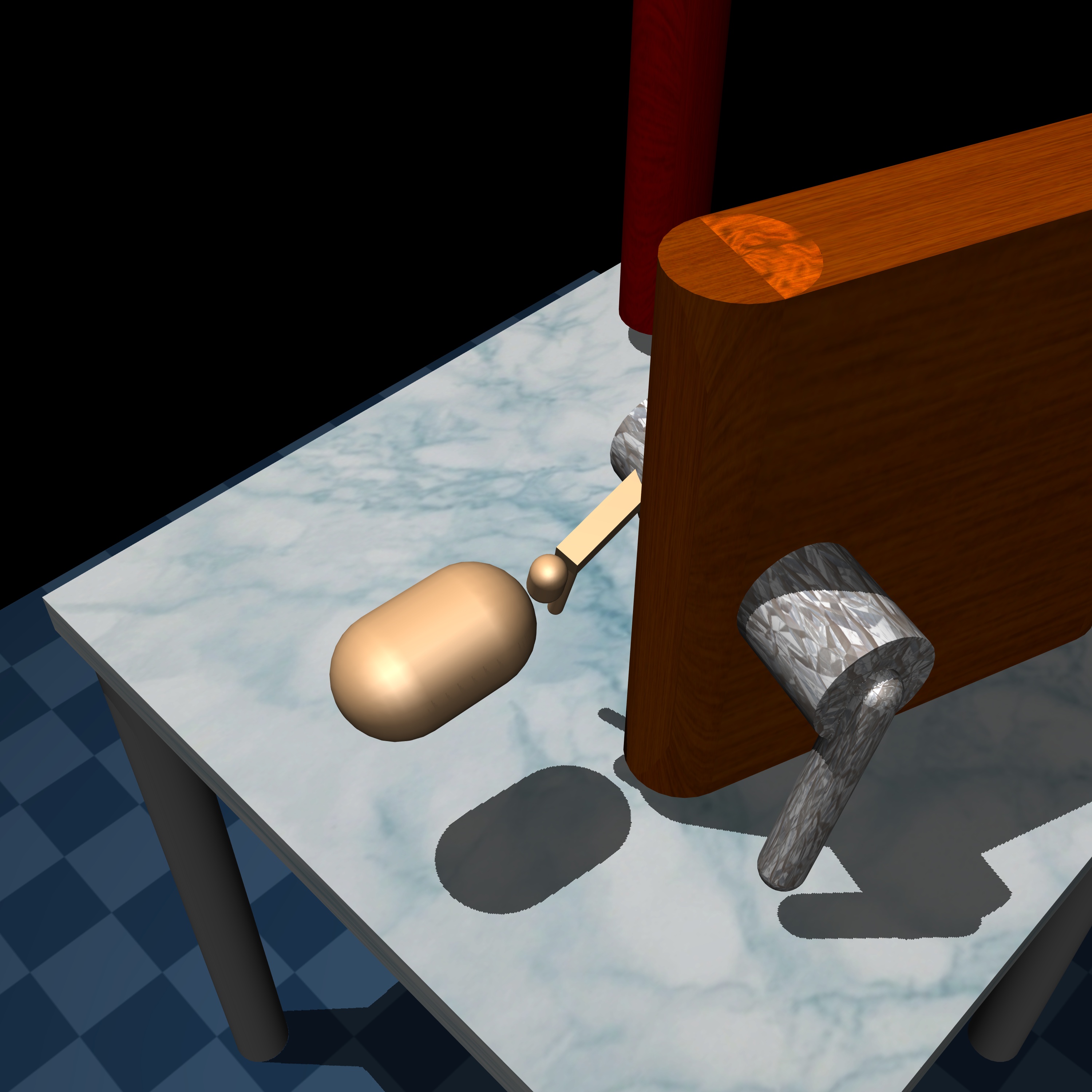} \\ \\
\includegraphics[width=\suppwidth\textwidth, valign=m]{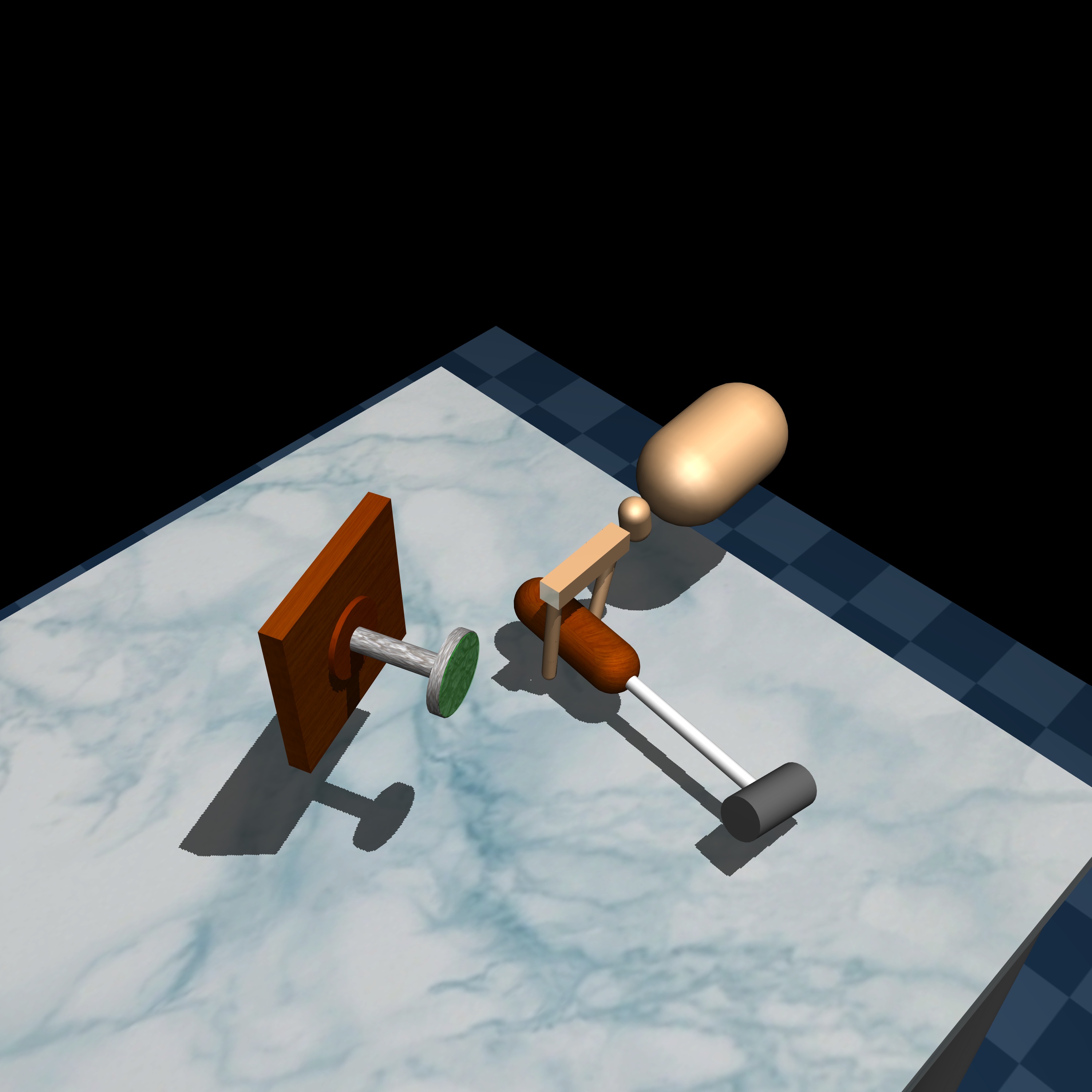}
\includegraphics[width=\suppwidth\textwidth,valign=m]{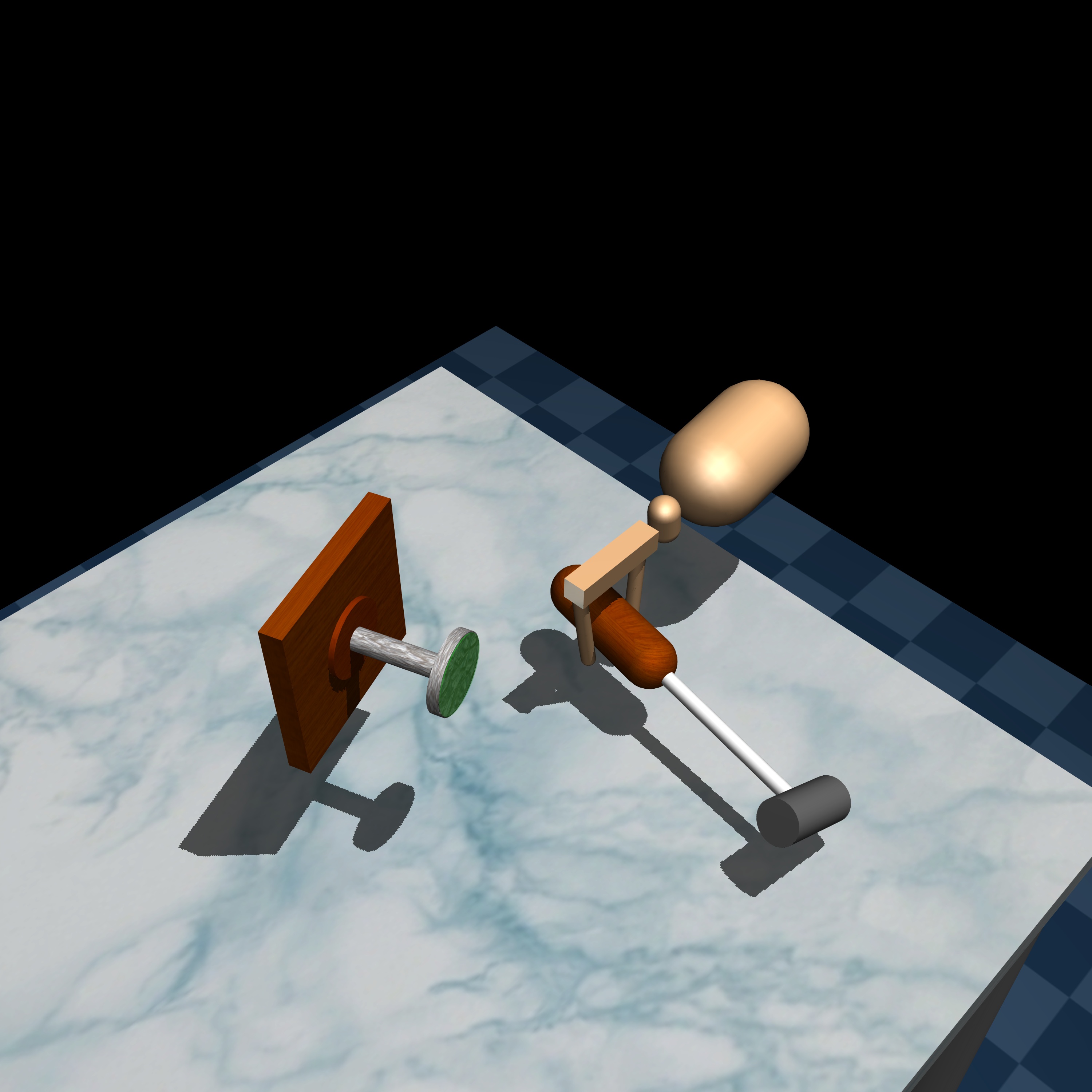}
\includegraphics[width=\suppwidth\textwidth,valign=m]{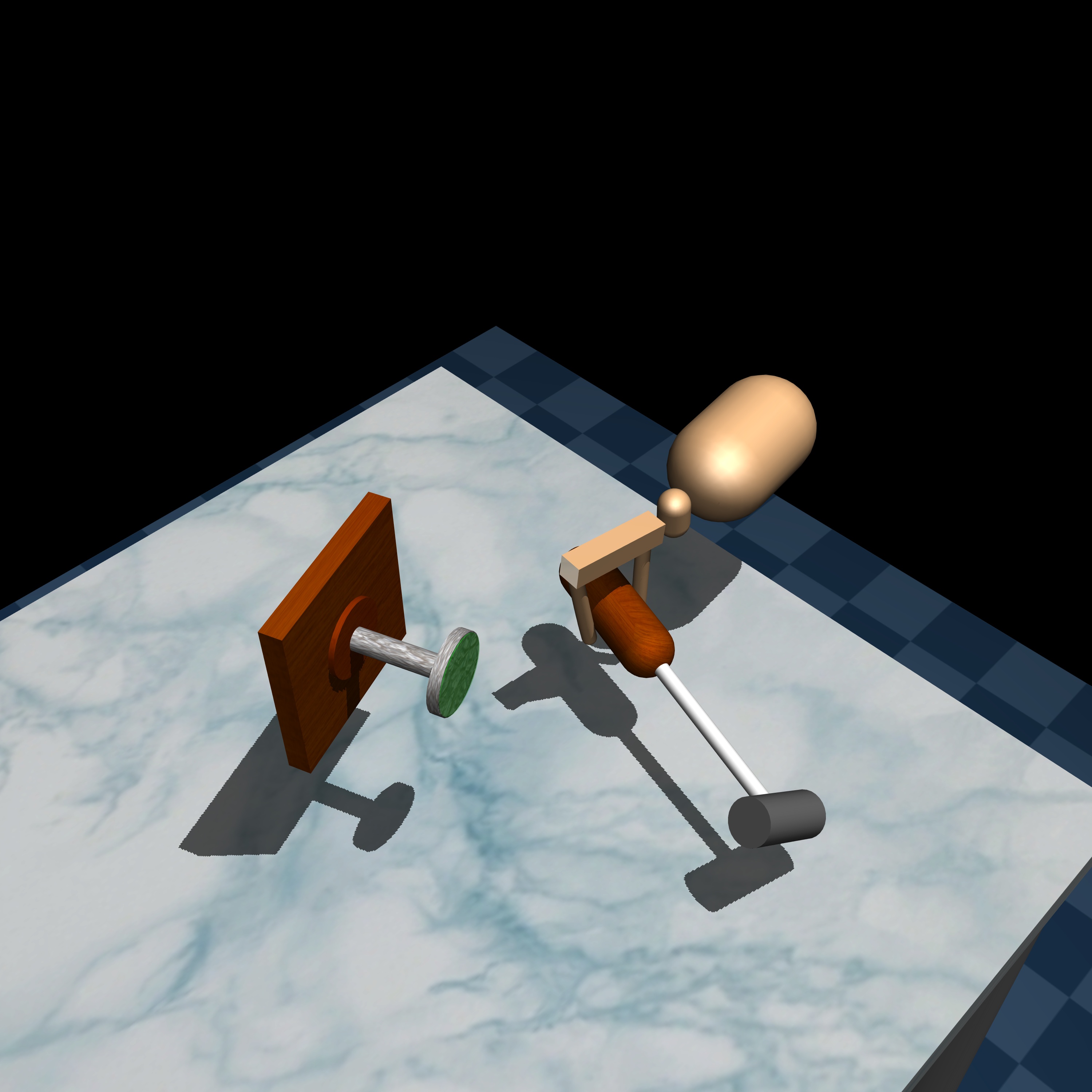}
\includegraphics[width=\suppwidth\textwidth, valign=m]{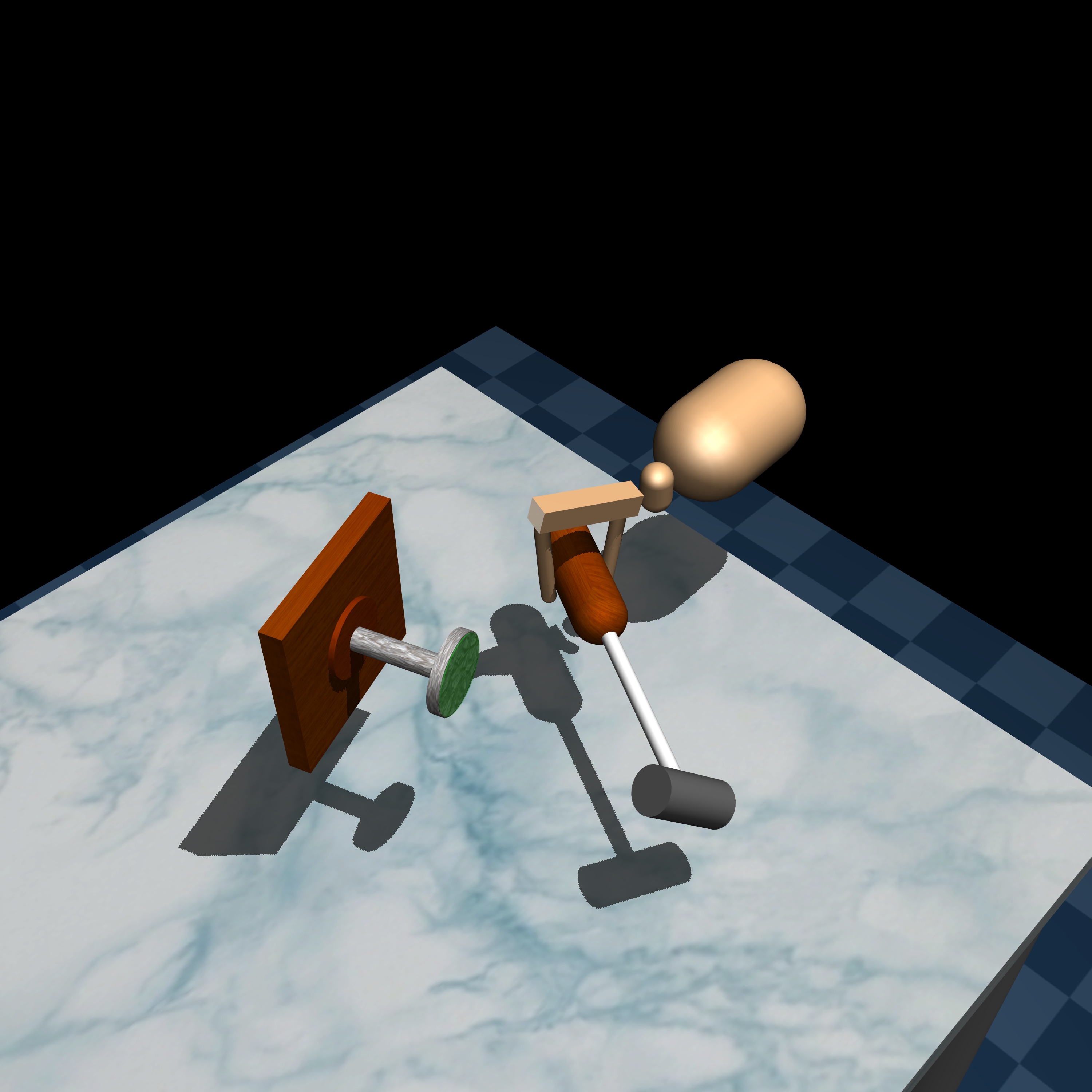}
\includegraphics[width=\suppwidth\textwidth,valign=m]{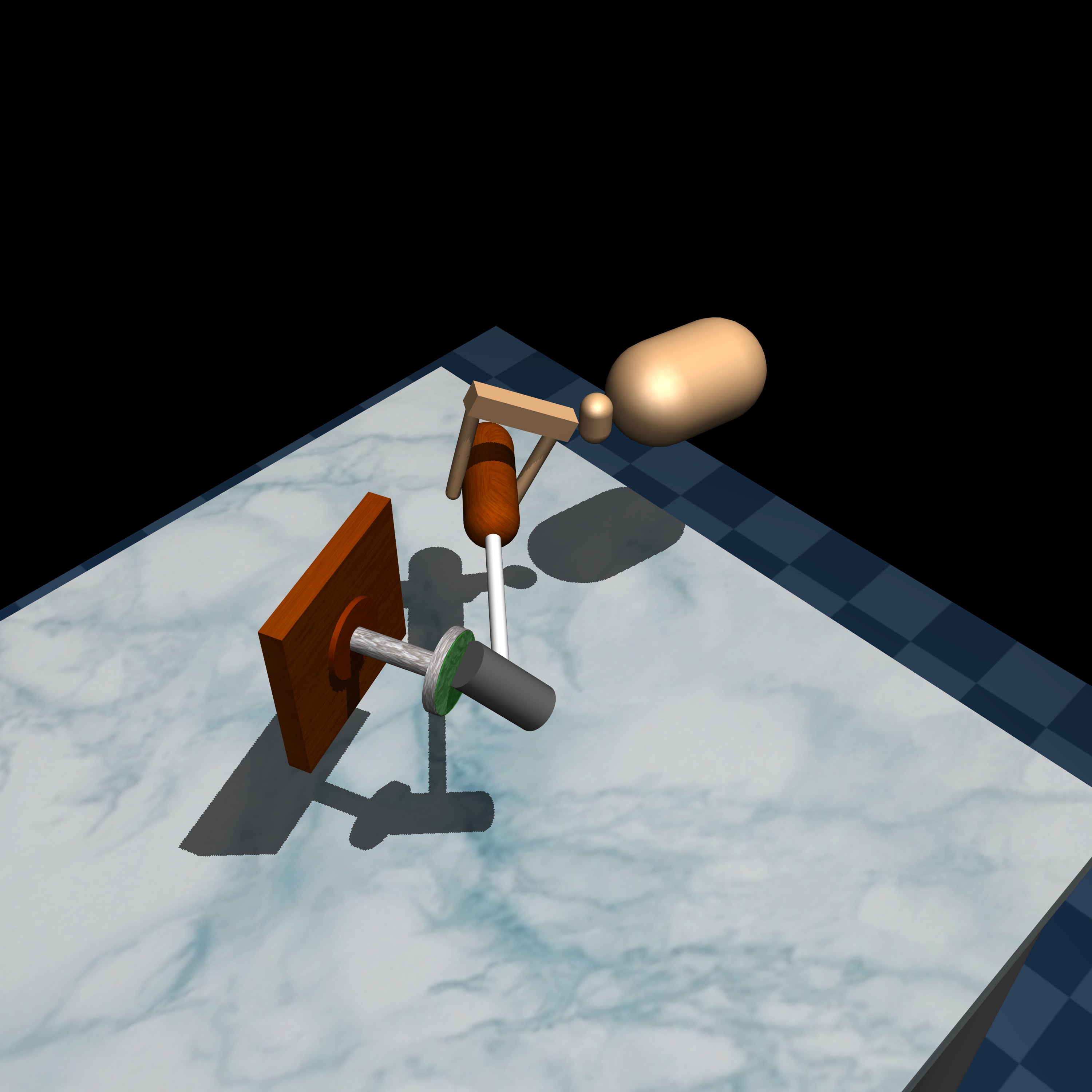}
\includegraphics[width=\suppwidth\textwidth,valign=m]{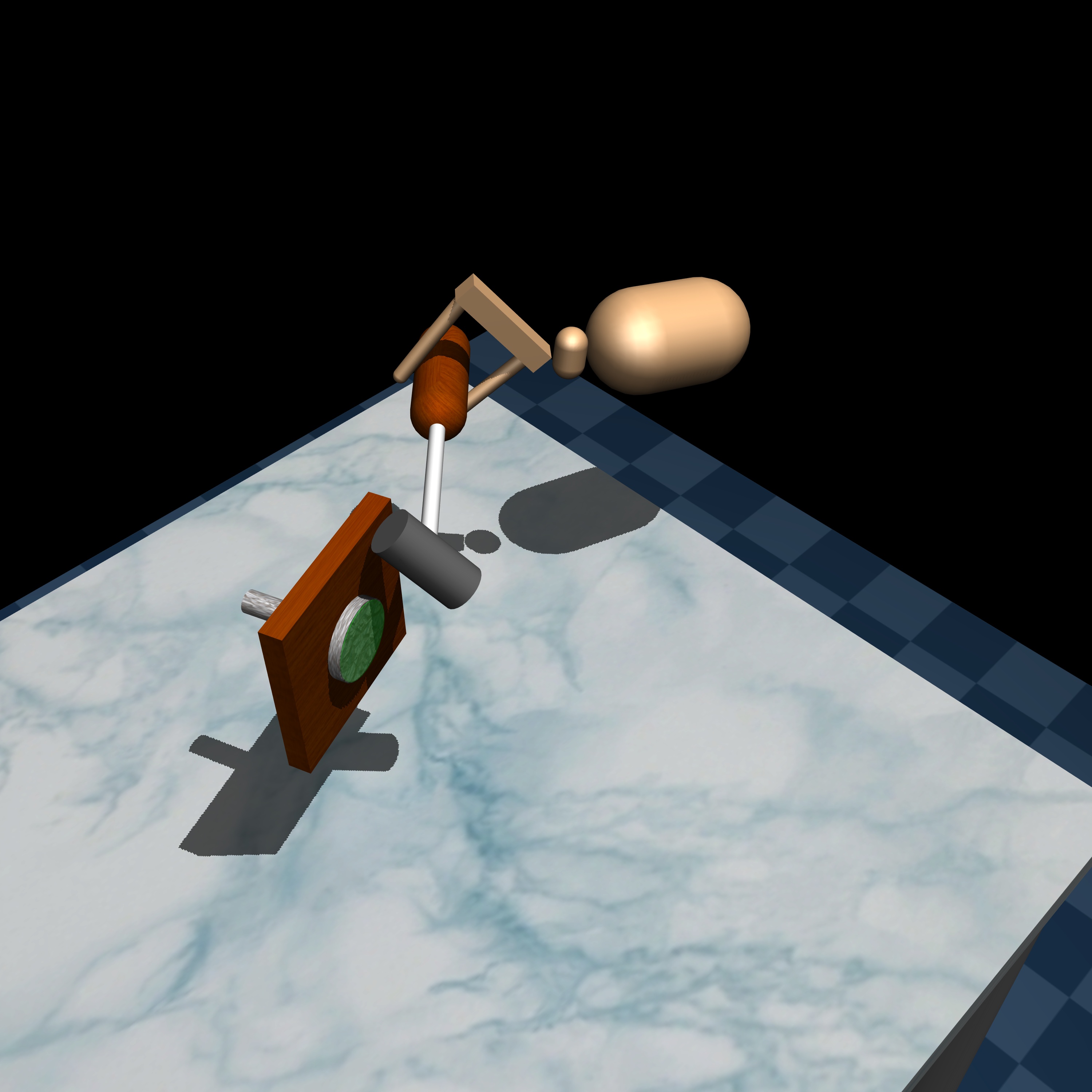} \\ \\
\includegraphics[width=\suppwidth\textwidth,valign=m]{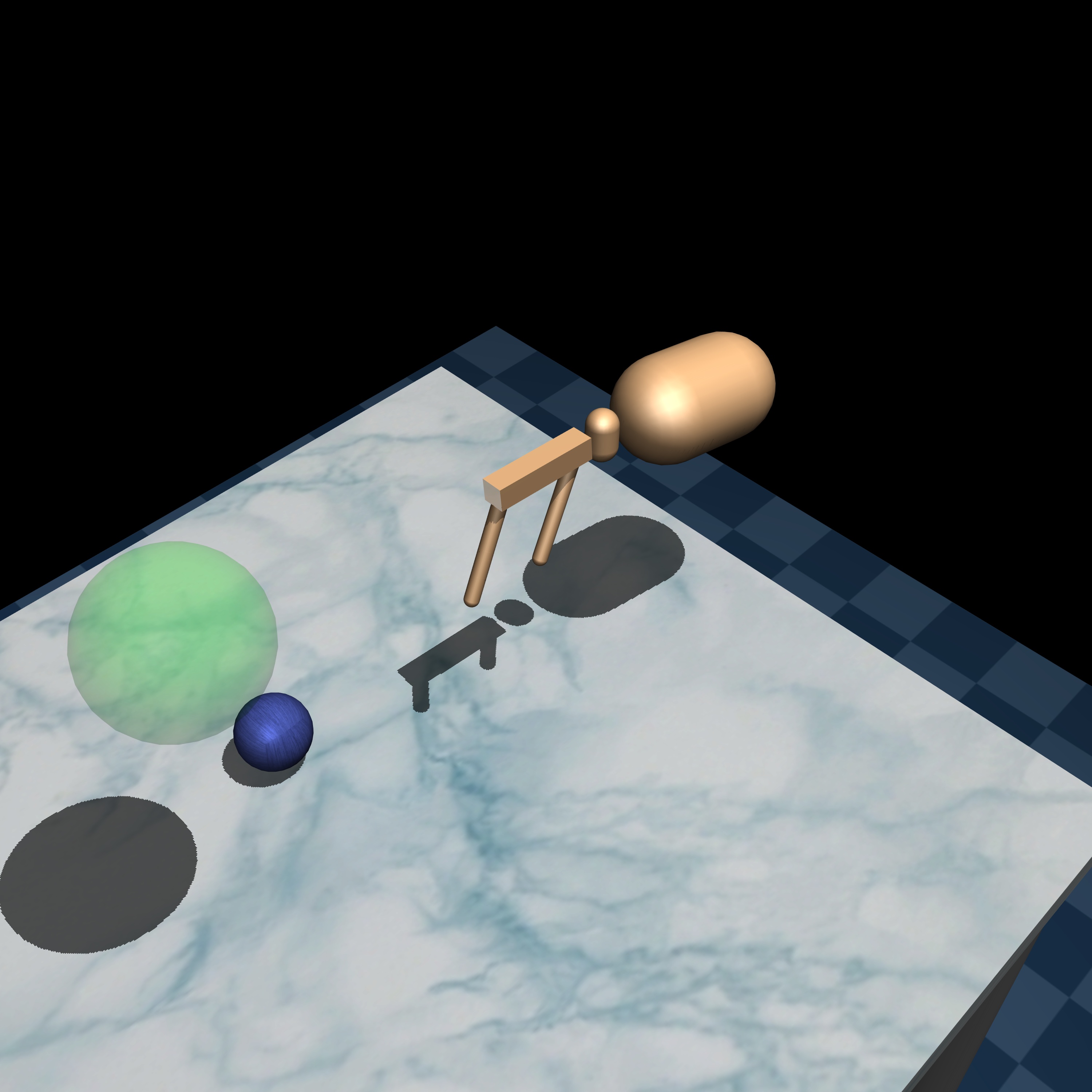}
\includegraphics[width=\suppwidth\textwidth,valign=m]{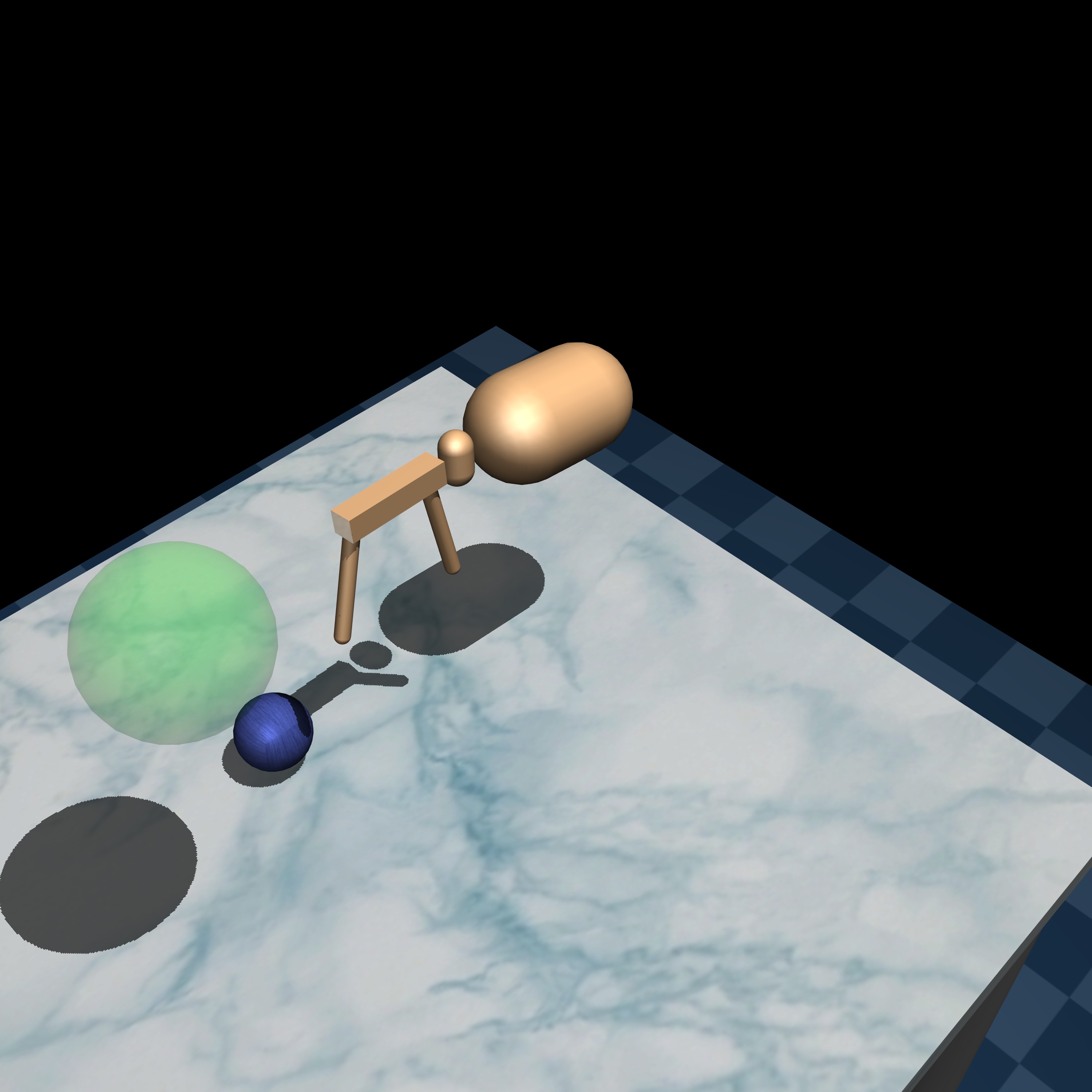}
\includegraphics[width=\suppwidth\textwidth,valign=m]{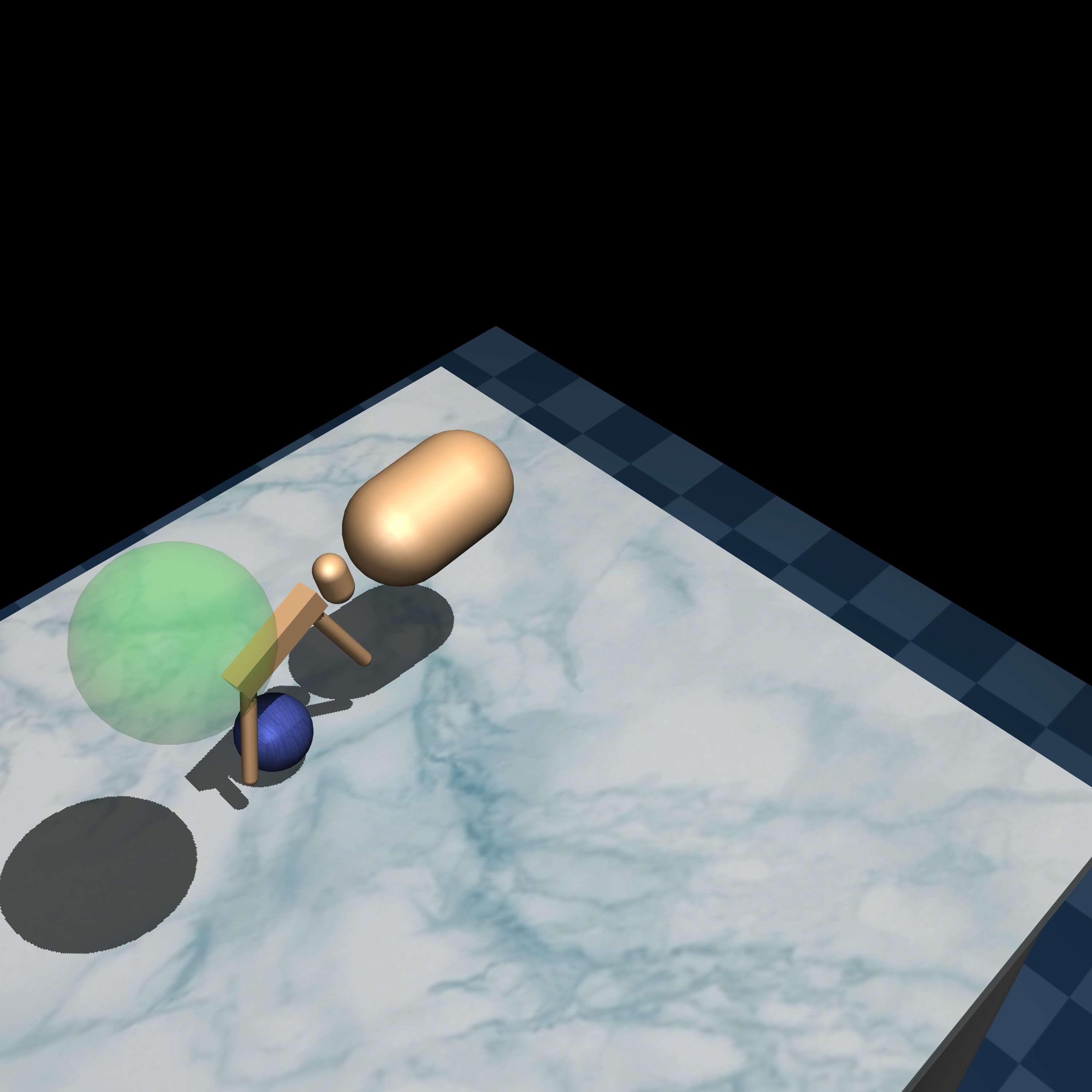}
\includegraphics[width=\suppwidth\textwidth,valign=m]{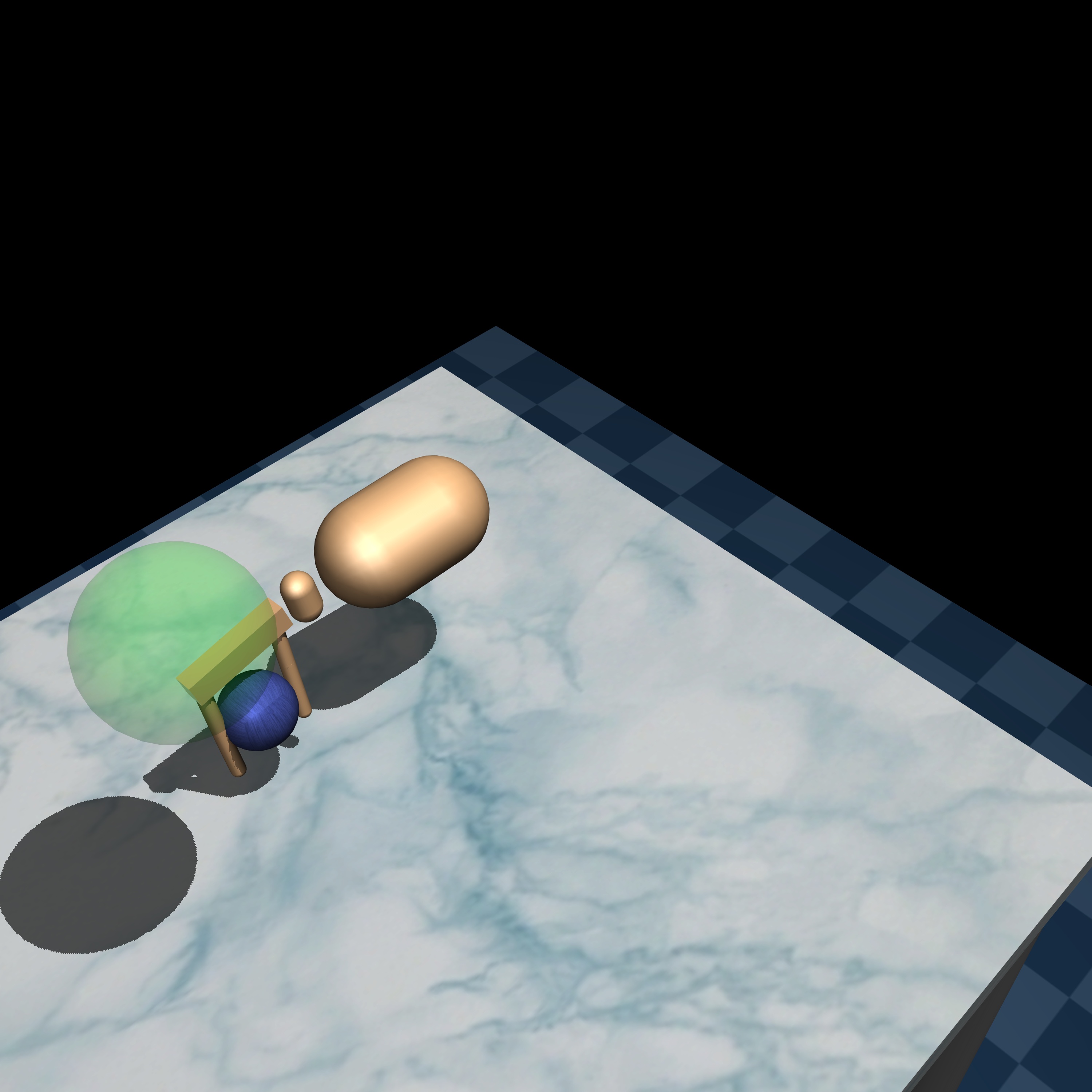}
\includegraphics[width=\suppwidth\textwidth,valign=m]{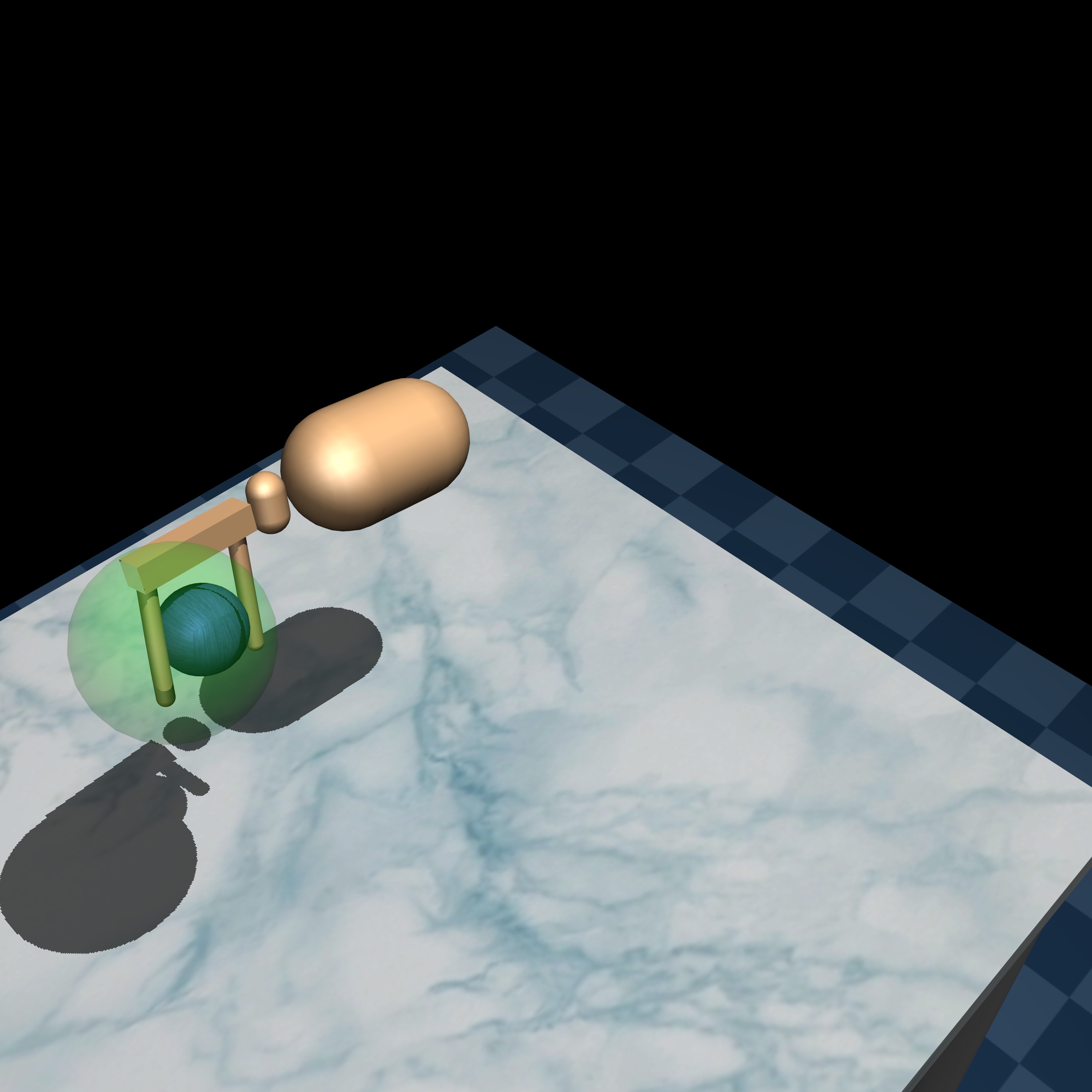}
\includegraphics[width=\suppwidth\textwidth,valign=m]{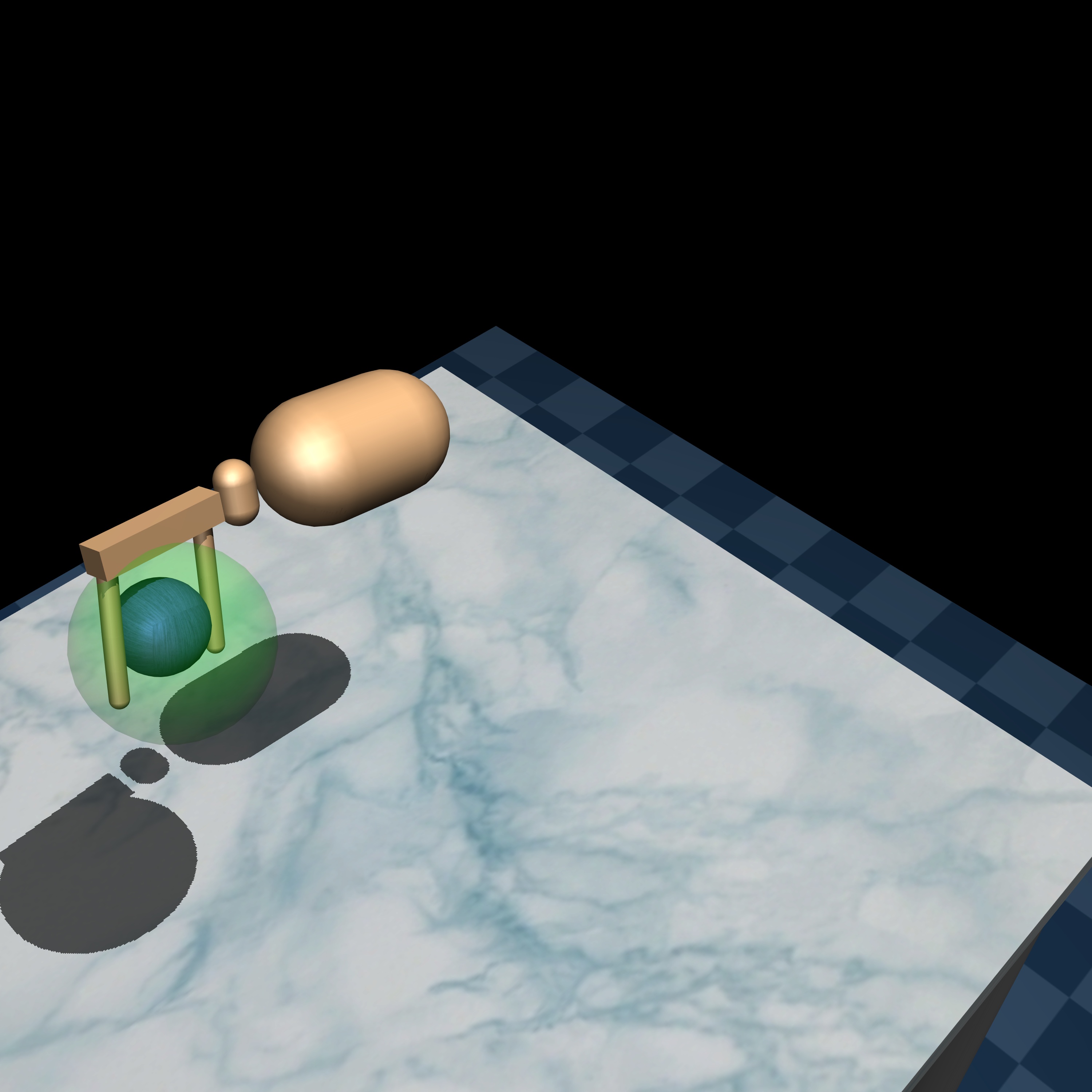}
\end{tabular}
\vspace{-1ex}
\caption{\textbf{Qualitative results on the target robot on Hand Manipulation Suite tasks.}
We show the transferred policy on target robots.
The three rows shows  \texttt{Door}, \texttt{Hammer}, and \texttt{Relocate} tasks respectively.
From left to right in each row is a policy rollout on target robot at $\alpha=1$.
} 
\label{fig:dapg:policy:viz}
\end{figure}

We show the process of transferring policy on intermediate robots on Hand Manipulation Suite tasks in Figure \ref{fig:dapg:evolve:viz}.
We show the transferred policy on the target robot of Hand Manipulation Suite tasks in Figure \ref{fig:dapg:policy:viz}. 
For more details, please refer to the supplementary video.

\end{document}